\documentclass{article} 
\usepackage{iclr2025_conference,times}

\usepackage{amsmath,amsfonts,bm}

\def\eqref#1{equation~\ref{#1}}

\def\1{\bm{1}}

\DeclareMathAlphabet{\mathsfit}{\encodingdefault}{\sfdefault}{m}{sl}
\SetMathAlphabet{\mathsfit}{bold}{\encodingdefault}{\sfdefault}{bx}{n}

\usepackage{hyperref}
\usepackage{url}

\usepackage{booktabs}       
\usepackage{xcolor}         

\usepackage{pifont}

\usepackage{multirow}

\usepackage{enumitem} 

\usepackage{amsmath}
\usepackage{amssymb}
\usepackage{mathtools}
\usepackage{amsthm}

\usepackage{makecell}

\usepackage{bm}
\newcommand{\ca}[1]{\bm{\mathcal{#1}}}
\usepackage{graphicx}
\usepackage{subfigure}
\usepackage{wrapfig}
\usepackage{adjustbox}

\theoremstyle{plain}
\newtheorem{theorem}{Theorem}[section]

\theoremstyle{definition}

\theoremstyle{remark}

\newcommand{\fatal}[1]{\textcolor{red}{ #1}}

\title{IDInit: A Universal and Stable Initialization Method for Neural Network Training}

\author{
Yu Pan$^1$ \hspace{0.5em} Chaozheng Wang$^{2}$ \hspace{0.5em} Zekai Wu$^3$ \hspace{0.5em} Qifan Wang$^4$ \hspace{0.5em} Min Zhang$^1$ \hspace{0.5em} Zenglin Xu$^{5,6}$\thanks{Corresponding author.} \\
$^1$Harbin Institute of Technology, Shenzhen \quad $^2$The Chinese University of Hong Kong \\
$^3$The Hong Kong Polytechnic University \quad $^4$MetaAI \quad
$^5$Fudan University \\
$^6$Shanghai Academy of AI for Science \\
\texttt{iperryuu@gmail.com} \quad \texttt{czwang23@cse.cuhk.edu.hk} \quad \texttt{zenglin@gmail.com}
}

\iclrfinalcopy 
\begin{document}

\maketitle

\begin{abstract}
Deep neural networks have achieved remarkable accomplishments in practice. The success of these networks hinges on effective initialization methods, which are vital for ensuring stable and rapid convergence during training. Recently, initialization methods that maintain identity transition within layers have shown good efficiency in network training. These techniques (e.g., Fixup) set specific weights to zero to achieve identity control. However, settings of remaining weight (e.g., Fixup uses random values to initialize non-zero weights) will affect the inductive bias that is achieved only by a zero weight, which may be harmful to training. Addressing this concern, we introduce fully identical initialization (IDInit), a novel method that preserves identity in both the main and sub-stem layers of residual networks. IDInit employs a padded identity-like matrix to overcome rank constraints in non-square weight matrices. Furthermore, we show the convergence problem of an identity matrix can be solved by stochastic gradient descent. Additionally, we enhance the universality of IDInit by processing higher-order weights and addressing dead neuron problems. IDInit is a straightforward yet effective initialization method, with improved convergence, stability, and performance across various settings, including large-scale datasets and deep models.
\end{abstract}

\section{Introduction}
\label{sec:intro}

Deep neural networks have attracted significant attention due to their versatility in various applications~\citep{DBLP:conf/cvpr/HeZRS16,li2021heuristic,wang2023tensor}. Behind these successes, initialization methods play a crucial role in promoting stable and fast-convergent training processes for networks~\citep{DBLP:conf/icml/SutskeverMDH13,DBLP:conf/nips/ArpitCB19,DBLP:conf/icml/0005SLW0X22,DBLP:conf/aaai/0005YYSXZSJL24}. Usually, initialization methods make effects by controlling the magnitude of signals. For example, Xavier~\citep{DBLP:journals/jmlr/GlorotB10} initialization is originally proposed to maintain signals in the non-saturated region of the sigmoid activation function by restricting signal variances, which greatly solved the difficulty of training. Then, \citet{DBLP:conf/nips/PooleLRSG16} propose to initialize network weights by constraining signals on the edge of chaos through dynamical isometry, which can further benefit the network training. Later, \citet{DBLP:conf/iclr/HardtM17} analyzed the optimization landscape of linear residual networks, and found that weights that transit identity in layers can help networks converge fast as their F-norm is close to that of the final converged weights. And identity transition also corresponds to isometry theory~\citep{DBLP:conf/iclr/ZhangDM19}, thereby, contributing to avoiding gradient explosion and diffusion.

\begin{wrapfigure}[8]{r}{0.46\textwidth}
\small
\begin{center}
    \vspace{-19pt}
    \includegraphics[width=0.34\textwidth]{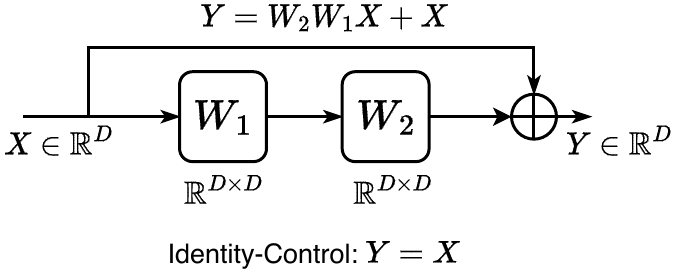}
    \vspace{-6pt}
    \caption{A case of identity-control initialization, which sets $W_2=\mathbf{0}$ to satisfy $Y=X$.
    }
    \label{fig:id-control}
\end{center}
\end{wrapfigure}

\begin{figure*}[t]
	\centering
	\subfigure[Initialization methods.]{
		\includegraphics[width=0.46\textwidth]{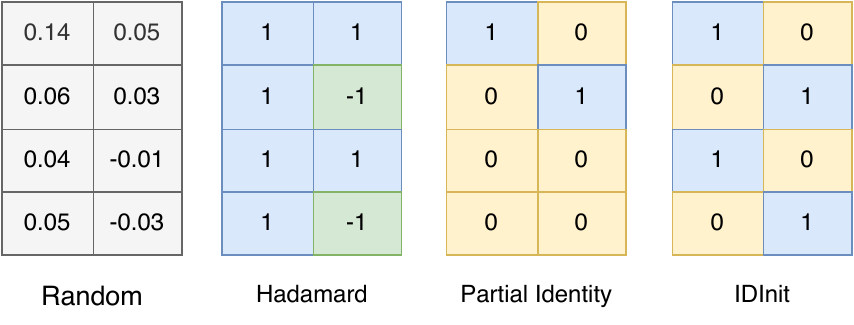}
		\label{fig:initials}
	}
	\subfigure[Square Loss.]{
		\includegraphics[width=0.22\textwidth]{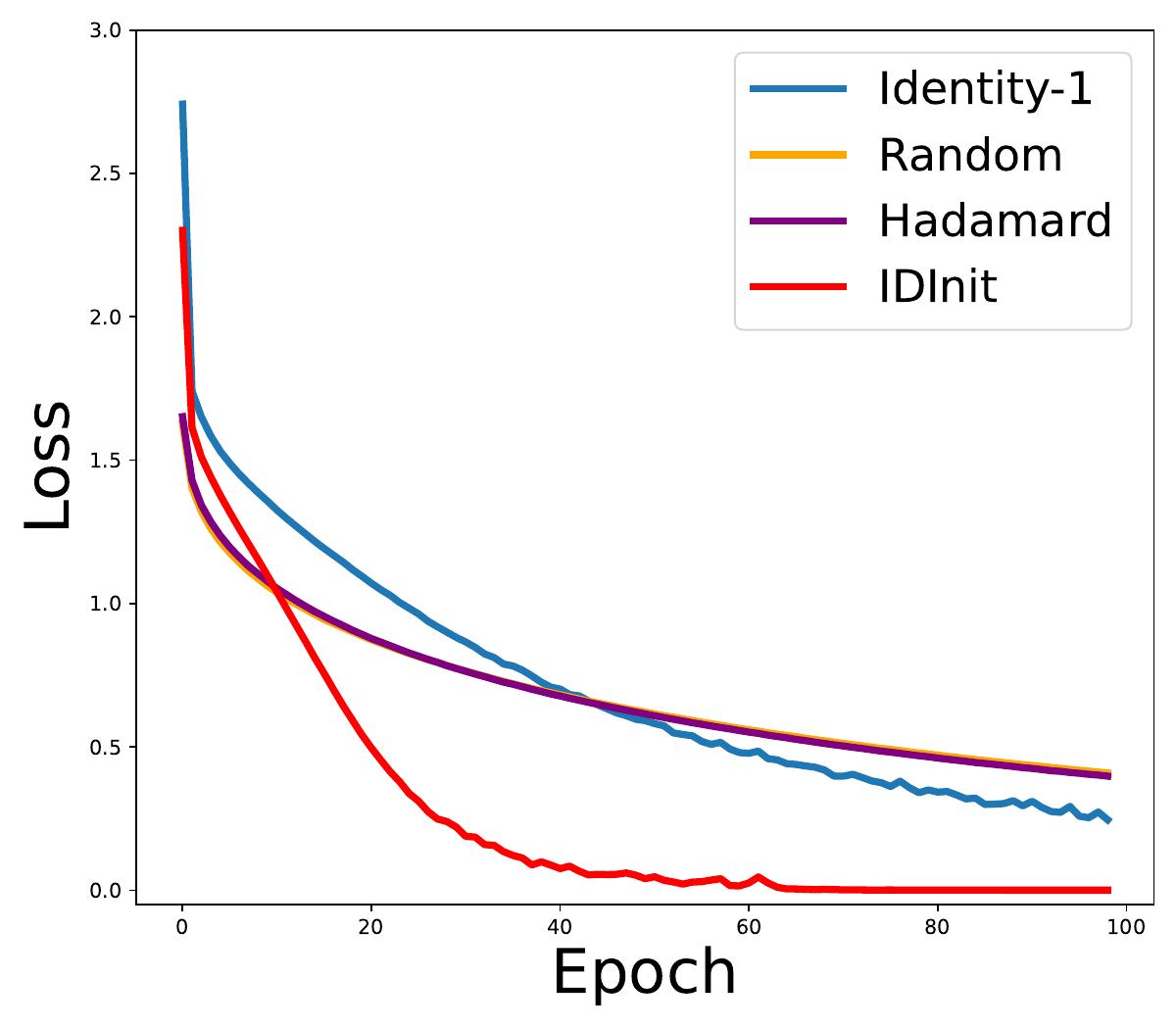}
		\label{fig:square-res}
	}
	\subfigure[Rectangle Loss.]{
		\includegraphics[width=0.22\textwidth]{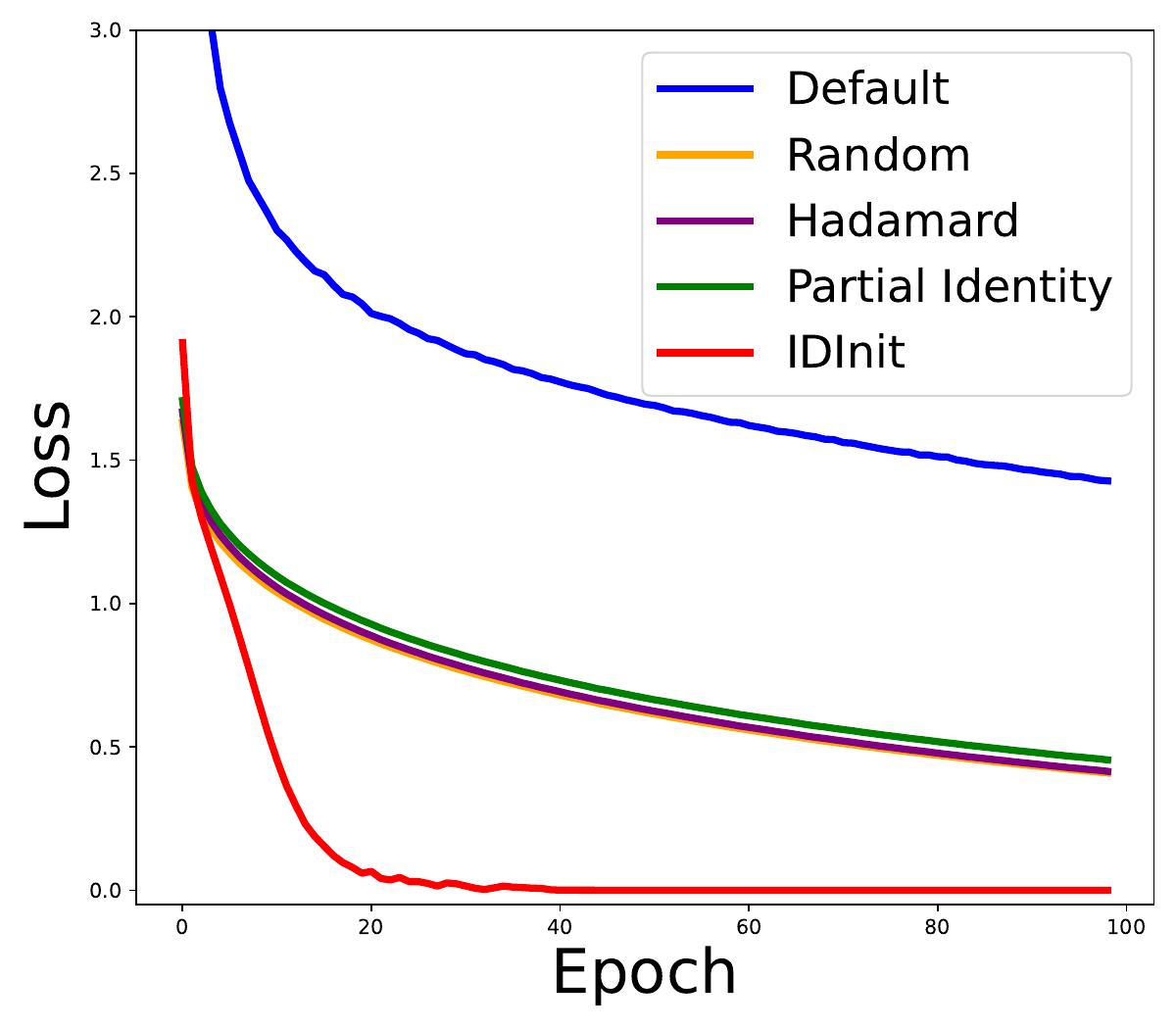}
		\label{fig:rec-res}
	}
 \vspace{-2mm}
	\caption{Analyzing effect of initializing $W_1$ while $W_2=\mathbf{0}$. The experiment uses Cifar10 and blocks in Figure~\ref{fig:id-control}, and more details are in Appendix~\ref{sec:jacana}. \subref{fig:initials} The initialization methods for $W_1$ in a rectangular format. Fixup: ``Random''; ZerO: ``Hadamard''. And ``Partial Identity'' and ``IDInit'' denote padding $\mathbf{0}$ and $I$ to an identity matrix, respectively. \subref{fig:square-res} Set $W_1\in \mathbb{R}^{240\times 240}$ and $W_2\in \mathbb{R}^{240\times 240}$ as square matrices. ``Identity-1'' represents a configuration where only one weight is initialized as $\mathbf{0}$.  Interestingly, while ``Random'' and ``Hadamard'' methods may outperform ``Identity-1'' in initial training epochs due to more network weights, they are hard to capture the inductive bias of ``Identity-1'', resulting in convergence difficulties. In contrast, IDInit can effectively leverage the training dynamics associated with ``Identity-1''. \subref{fig:rec-res} Set $W_1\in \mathbb{R}^{280\times 240}$ and $W_2\in \mathbb{R}^{240\times 280}$ as rectangle matrices. ``Default'' means $W_1$ and $W_2$ are initialized with Xavier. However, ``Default'' proves ineffective for training, as it conflicts with dynamical isometry. Furthermore, even though ``Partial Identity'' exhibits the capability to transmit partial signals, it performs poorly due to rank constraint issues. Finally, IDInit maintains well-training conditions by padding the identity matrix.
}
	\label{fig:idi-motivation}
 \vspace{-4mm}
\end{figure*}

An instance of preserving identity across neural network layers, known as "identity-control," is depicted in Figure~\ref{fig:id-control} and formally expressed as $Y=X$. This type of initialization can be implemented by setting specific weights (e.g., $W_2$) to $\mathbf{0}$, thereby ensuring zero output in the sub-stem, as elucidated by \citet{DBLP:conf/iclr/HardtM17}. This approach, however, poses challenges in configuring the remaining weight $W_1$. Previous work such as Fixup~\citep{DBLP:conf/iclr/ZhangDM19} and ZerO~\citep{zhao2022zero} initialize $W_1$ using the Xavier and Hadamard methods, respectively. These initializations can adversely affect the inductive bias already established by setting $W_2=\mathbf{0}$, a setting beneficial for training. As evidenced in Figure~\ref{fig:idi-motivation}, both Xavier and Hadamard methods cause difficulties in achieving convergence. Observing this, we propose initializing $W_1$ with an identity matrix  $I$, which retains the inductive bias as $IW_2\equiv W_2$. Moreover, $I$ also achieves dynamical isometry in the sub-stem layer as discussed by \citet{zhao2022zero}. Figure~\ref{fig:idi-motivation} demonstrates that using an identity matrix significantly aids in training convergence. Nonetheless, the practical application of an identity matrix faces two primary obstacles.  First, an identity matrix requires square-shaped weights, a condition seldom met in practical networks. While a partial identity matrix (by padding $\mathbf{0}$ to an identity matrix) offers a workaround, it leads to rank constraints issues~\citep{zhao2022zero} when the output dimension exceeds the input dimension, impairing network generalization. The second obstacle concerns the convergence capability. As \citet{DBLP:journals/neco/BartlettHL19} pointed out, weights initialized with an identity matrix are difficult to converge to the ground truth, of which eigenvalues contain negative values. This convergence problem is important as it indicates a limited universality of applying an identity matrix as an initialization method.

\textbf{IDInit.} In light of the preceding discussion, we aim to address these two major obstacles. To handle a non-square matrix, we pad a new identity matrix in adjacency to an identity matrix. We theoretically demonstrate this operation can resolve the rank constraint problem. Then, to alleviate the replica problem induced by this padding scheme, we impose a loosening condition on the padded identity-like matrix. Turning to the matter of convergence, we conduct an experiment to analyze it. Interestingly, we find that the convergence problem can be solved by adding a moment in an optimizer (e.g., the stochastic gradient descent optimizer), which is the most general setting for training neural networks. By introducing the identity-like matrix into the identity-control framework, we implement a fully identical initialization (IDInit), which ensures identity transition across both main and sub-stem layers.
Moreover, we explore two additional techniques for improving the universality of IDInit and the identity-control framework:

(1) Higher-order Weights: An identity matrix is a 2-D array and it is necessary to consider an efficient method to transfer the identity matrix to a higher-order weight (e.g., a 4-D convolution). A previous strategy is to keep identity along the channel (see Sec.~\ref{sec:idiconv}). However, this causes diversity loss in channels, which is harmful to performance. To remedy this shortage, we keep identity in patches alternatively for more diversity in channels to achieve improvement.

(2) Dead Neurons: As an identity-control method, IDInit sets the last layer of the sub-stem to 0 for transiting identity in the main branch. However, a dead neuron problem is possibly caused by this setting, especially for residual convolutional networks~\citep{DBLP:conf/iclr/ZhangDM19,zhao2022zero}. Addressing this, we select some elements to a small numerical value $\varepsilon$ to increase trainable neurons as in Figure~\ref{fig:idinit-overivew}.

To our knowledge, IDInit is the first successful trial to maintain identity in both main- and sub-stems by breaking the rank constraints, which promise the expressive power of IDInit. Then, we address the replica problem by adding small noise while maintaining the dynamical isometry. By further proposing modifications to CNNs and solutions to dead neuron problems, we have significantly improved accuracy of classifying Cifar10 on ResNet-20 by 3.42\% and 5.89\%, respectively. (see Section~\ref{sec:ablation}). Note that, although the identity matrix is used as initialization in prior work, it was only used for square matrix, e.g., \citet{DBLP:journals/corr/LeJH15} set a hidden-to-hidden layer in a recurrent neural network with an identity matrix for better performance. IDInit is novel for the consideration of non-standard situations, e.g., non-square matrix. On ImageNet, compared to the default random initialization, IDInit demonstrates superior performance, achieving an average improvement of 0.55\%, and facilitates faster convergence across various settings, reducing the required training time by an average of 7.4 epochs. IDInit can accelerate the training procedure of BERT-Base, manifesting an 11.3\% reduction in computational cost. Therefore, our approach yields consistently significant advantages in the training of neural networks.

\section{Related Work}
\label{sec:back}

Consider an $L$-layer residual network, each residual block of which consists of a residual connection and a residual stem that refers to the component excluding the residual connection. Assuming each residual stem contains two parameters, and the network's input signal is denoted as $x^{(0)}$, the $i$-th layer can be formulated as
\begin{align}
\label{eq:residual}
    x^{(i+1)} = a(I + \theta^{(i, 0)}\theta^{(i, 1)}) x^{(i)},
\end{align}
where $a(\cdot)$ denotes the activation function, $x^{(i)}$ means an input of $i$-th residual block in a network, $I$ is an identity matrix denoting residual connection, and $\theta^{(i, 0)}$ and $\theta^{(i, 1)}$ are weights in the $i$-th residual stem of a residual block.

\paragraph{Dynamical Isometry.} 
Assuming the signal magnitude (e.g., $\sigma^2(x^{(i)})$) of each layer changing in a scale $\alpha$, the last signal magnitude can reach $\alpha^L$ (e.g., $\sigma^2(x^{(L)}) = \alpha^L \sigma^2(x^{(0)})$), making it easy to cause signal explosion and diffusion, especially for large $L$. To mitigate this issue, dynamic isometry provides an effective solution.
Considering the input-output Jacobian which is defined as
\begin{align}
    J_{io} = \frac{\partial x^{(L)}}{\partial x^{(0)}},
\end{align}
the dynamical isometry is achieved when all the singular values of $J_{io}$ are close to 1. Moreover, with the mean squared singular value of $J_{io}$ noted as $\chi$, \citet{DBLP:conf/nips/PenningtonSG17} and \citet{DBLP:conf/uai/BachlechnerMMCM21} show that $\chi > 1$ indicates that the model is in a chaotic phase, and back-propagated gradients will explode exponentially. By contrast, $\chi < 1$ means a model in an ordered manner that back-propagated gradients vanish exponentially. $\chi = 1$ is a critical line of initialization, avoiding gradient vanishing or exploding. The isometry can provide sufficient robustness for the network training~\citep{DBLP:journals/corr/abs-1901-08987,DBLP:conf/nips/PooleLRSG16,DBLP:conf/nips/YangS17}.

\paragraph{Network Initialization.} 
Common initialization methods are Xavier~\citep{DBLP:journals/jmlr/GlorotB10} and Kaiming initialization~\citep{DBLP:conf/iccv/HeZRS15}. Especially for residual networks efficiency, \citet{DBLP:conf/iclr/HardtM17} theoretically demonstrates that network training benefits from keeping identity. \citet{DBLP:journals/corr/LeJH15} set a hidden-to-hidden layer in a recurrent neural network with an identity matrix for better performance.
Fixup~\citep{DBLP:conf/iclr/ZhangDM19} and ZerO~\citep{zhao2022zero} successfully initialize ResNets by setting residual stem to 0 (not residual connections) to guarantee the identity of signals. 
SkipInit~\citep{DBLP:conf/nips/DeS20} replaces Batch Normalization with a multiplier whose value is 0. ReZero~\citep{DBLP:conf/uai/BachlechnerMMCM21} directly adds extra parameters of value 0 to keep identity, leading to fast convergence.

\paragraph{Identity-Control Training Framework.} 
Net2Net~\citep{DBLP:journals/corr/ChenGS15} proposes to expand network depth by maintaining identity. DiracNet~\citep{DBLP:journals/corr/ZagoruykoK17} maintains an identity for propagating information deeper into the network. However, it suffers from reducing residual connection, causing performance loss. ISONet~\citep{DBLP:conf/icml/QiYWMM20} is an isometric learning framework that contains an identical initialization (i.e., the Dirac function that is also used in ZerO~\citep{zhao2022zero} by padding 0 in a non-square matrix case), and isometric regulation in training. ISONet multiplies 0 to the residual stem like Fixup~\citep{DBLP:conf/iclr/ZhangDM19}. ISONet lacks the flexibility for various convolutions as it specifies the net without normalization, and requires SReLU.

\section{Fully Identical Initialization }
\label{sec:idinit}

\begin{figure}[t]
	\centering
	\includegraphics[width=0.99\textwidth]{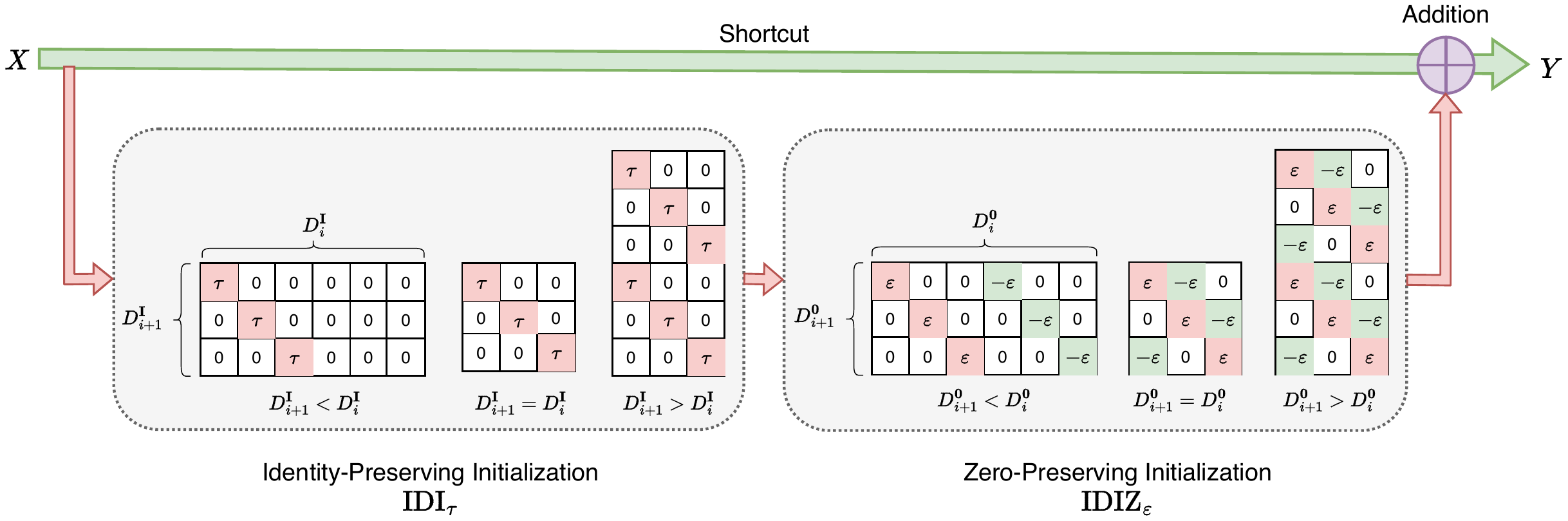}
	\caption{An overview of IDInit, which consists of identity-preserving initialization $\operatorname{IDI}_{\tau}$ and zero-preserving initialization $\operatorname{IDIZ}_{\varepsilon}$, of which dimensions are denoted as $D^{\mathbf{I}}$ and $D^{\mathbf{0}}$. $\tau$ and $\epsilon$ are usually set to 1 and 1e-6 to maintain identity and transit zero. $i$ and $i+1$ mean two adjacent layer indices.}
	\label{fig:idinit-overivew}
\end{figure}

The identity-control scheme serves as a practical initialization framework, with prior studies such as Fixup and ZerO demonstrating success within this paradigm.
As depicted in Figure~\ref{fig:idinit-overivew}, IDInit achieves this scheme with two components: identity-preserving initialization and zero-preserving initialization, aimed at transferring identity and zero, respectively. We elaborate on the identity-preserving initialization, which involves padding identity matrices, in Section~\ref{sec:identity}, and discuss the zero-preserving initialization, which addresses dead neurons, in Section~\ref{sec:zero-preserving}.

\subsection{Preserving Identity by Padding Identity}
\label{sec:identity}

A standard identity matrix can naturally satisfy identity transition. However, in a non-square situation, this natural advantage is lost. To address this problem, we pad the identity matrix on an identity matrix to fit a non-square matrix. Specifically, for a fully-connected layer transformed from Eq.~(\ref{eq:residual}) as $x^{(i+1)} = \theta^{(i)} x^{(i)}$, we set the weight $\theta^{(i)} \in \mathbb{R}^{D^{\mathbf{I}}_{i+1} \times D^{\mathbf{I}}_{i}}$ to
\begin{align}
\label{eq:idi-tau}
\theta^{(i)}_{m,j} = 
    \begin{cases}
      \tau, &\text{if $m \equiv j \pmod{D^{\mathbf{I}}_{i}}$}, \\
      0, &\text{otherwise}.
    \end{cases}
\end{align}
The initialization formulated as Eq.~(\ref{eq:idi-tau}) is termed as $\operatorname{IDI}_{\tau}$, where $\operatorname{IDI}$ means the identical initialization function, and 
$\tau$ is calculated by considering the activation function, e.g., $\tau_{ReLU}=\sqrt{2}$ for the ReLU function.
As shown in Figure~\ref{fig:idinit-overivew}, setting $\tau = 1$ can form $\operatorname{IDI}_1$ initialization.

\subsubsection{Analysis on Convergence Ability of the Identity Matrix}
\label{sec:convergence}

As proposed by \cite{DBLP:journals/neco/BartlettHL19}, weights initialized with an identity matrix face difficulty in converging towards the target when its eigenvalues include negative values. This implies a potential constraint on the convergence efficacy of the IDInit method. Consequently, we will delve deeper into this issue in the following discussion. According to their study, when layers in a neural network are initialized using the identity matrix, all the weight matrices of layers will be symmetric at each step of the training process. This persistent symmetry leads to the weights of layers always being the same at any training step, causing the aforementioned convergence difficulty. Interestingly, we find that this problem is mainly caused by the gradient descent (GD) which uses all the data in one batch, and employing a stochastic gradient descent (SGD) of which data in different batches can be different, can effectively break the symmetry in gradients which facilitates convergence, and incorporating momentum can further accelerate the convergence process.

To elaborate on this problem, we present a training case for a single-layer network expressed as $y=\theta x$, where $x\in \mathbb{R}^d$ represents the input, $y\in \mathbb{R}^d$ denotes the output, and $\theta\in \mathbb{R}^{d\times d}$ is the weight matrix. The weight matrix $\theta$ is initialized to the identity matrix $I$, denoted as $\theta^{(0)}=I$. For our loss function, we employ the Mean Squared Error (MSE) and a learning rate denoted by $\eta$. Consider two training pairs $\{x_1, y_1\}$ and $\{x_2, y_2\}$ sampled from the same dataset $\mathcal{D}$. The network is initially trained with $\{x_1, y_1\}$, and trained with $\{x_2, y_2\}$ in the next step. 

Being updated after two steps, the final 
gradient $\Delta \theta^{(1)}$ can be calculated as
\begin{align}
\label{eq:step2-grad}
\centering
    x_2x_2^T - \eta x_1x_1^Tx_2x_2^T + \eta y_1x_1^Tx_2x_2^T - y_2x_2^T.
\end{align}
While $x_2x_2^T$ is symmetric, $x_1x_1^Tx_2x_2^T$, $y_1x_1^Tx_2x_2^T$, and $y_2x_2^T$ can be asymmetric. To quantify the magnitude of the asymmetry in $\Delta \theta^{(1)}$, let $\Omega = -\eta x_1x_1^Tx_2x_2^T + \eta y_1x_1^Tx_2x_2^T - y_2x_2^T$ denote the asymmetric component. The magnitude of the asymmetry can be calculated as $\mathbb{E}(||\Omega - \Omega^T||_F^2)$. Assuming $x_1, x_2, y_1, y_2 \in \mathbb{R}^d$ are random vectors with entries that are i.i.d. Gaussian random variables distributed as $N(0, \sigma^2)$, the magnitude of the asymmetry is bounded as
\begin{align}
   4\eta^2d^3\sigma^8-4\eta^2d^2\sigma^8+2d^2\sigma^4 \leq \mathbb{E}(||\Omega - \Omega^T||_F^2) \leq 6\eta^2d^3\sigma^8 + 3d^2\sigma^4.
\end{align}
As $\eta$ is usually $1e-1$, and both training pairs $\{x_1, y_1\}$ and $\{x_2, y_2\}$ can be generally normalized to $\mathcal{N}\sim(0, 1)$, thereby, the symmetry of the weight can be sufficiently influenced as 
\begin{align}
    \theta^{(2)} &= \theta^{(1)} - \eta \Delta \theta^{(1)}.
\end{align}
When introducing a momentum $m^{(0)}$ initialized to $\Delta \theta^{(0)}$, $\theta^{(2)}$ will be updated as
\begin{align}
\centering
    m^{(1)} &= \gamma m^{(0)} + \eta \Delta \theta^{(1)}, \notag \\ 
    \theta^{(2)} &= \theta^{(1)} - m^{(1)} = \theta^{(1)} - \gamma m^{(0)} - \eta \Delta \theta^{(1)},
\end{align}
where $\gamma$ is the coefficient of $m$. Therefore, momentum can promote the weight to become asymmetric by accumulating the asymmetry of gradients in steps and impact more when samples are increased. We show that SGD with momentum can effectively resolve the issue of layers being the same in networks initialized with the identity matrix during training, which facilitates the convergence process. The completed derivation is provided in Sec.~\ref{sec:converge-analysis} of the appendix.

As for networks of multiple layers, when their layers are asymmetric, each layer can be updated differently which breaks the convergence problem caused by the same gradients in each step (which is stated in Lemma 5 of \citet{DBLP:journals/neco/BartlettHL19}). As illustrated in Figure~\ref{fig:differlayers} of the appendix, it is evident that layers trained using SGD are different from each other, with the momentum component amplifying the degree of this difference.

\subsubsection{On Rank Constraint Problem}

\begin{wrapfigure}[13]{r}{0.5\textwidth}
\small
\begin{center}
    \vspace{-59pt}
     \subfigure[Padding schemes.]{
		\includegraphics[width=0.2\textwidth]{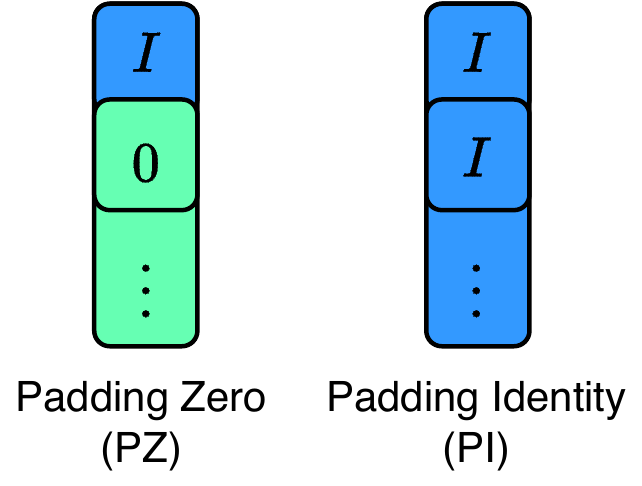}
		\label{fig:paddings}
	}
 \subfigure[Rank plot.]{
		\includegraphics[width=0.24\textwidth]{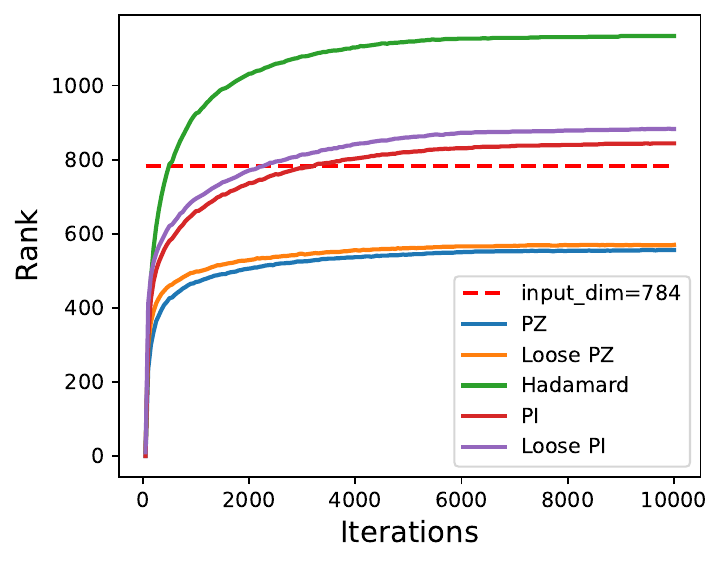}
		\label{fig:matrix-rank}
	}
    \vspace{-10pt}
    \caption{Two padding schemes and their influence on ranks of a layer. We trained a 3-layer network on MNIST, and set $D_0=768$ and $D_h=2048$. We plot $\text{rank}(\Delta\theta^{(1)})\in \mathbb{R}^{D_h \times D_h}$ in \subref{fig:matrix-rank}. As shown in \subref{fig:matrix-rank}, padding identity can achieve more than a rank of 768 like Hadamard, while padding zero is limited under 768. The loose condition can lead to better rank performance, however, cannot solve the rank constraint problem of padding zero.}
    \label{fig:padding-rank}
\end{center}
\end{wrapfigure}

ZerO~\citep{zhao2022zero} identifies that a dimension-increasing matrix may face a rank constraint problem if padding zero values. In this analysis, we investigate whether padding with an identity matrix results in this constraint.

\textbf{Rank Constraint Problem.} Consider a 3-layer network with weights $\{\theta^{(i)}\}_{i=0}^{2}$, where $\theta^{(0)}\in \mathbb{R}^{D_h \times D_0}$,  $\theta^{(1)}\in \mathbb{R}^{D_h \times D_h}$, $\theta^{(2)}\in \mathbb{R}^{D_L \times D_h}$ where  $D_h > D_0, D_L$. Given an input batch  $x^{(0)}\in \mathbb{R}^{D_0\times N}$ with a size $N$, the formulation of the $i$-th layer is $x^{(i+1)} = \theta^{(i)} x^{(i)}$,
where $i\in [2]$. Define residual component $\Delta{\theta}^{(1)} = {\theta}^{(1)} - I$. When initializing the dimension-increasing weight $\theta^{(0)}$ by padding zeros (PZ) values, the rank constraint problem refers to
\begin{align}
    \text{rank}(\Delta{\theta}^{(1)}) \leq D_0.
\end{align}

This rank constraint issue signifies a performance limitation associated with the initialization method. Intriguingly, our findings indicate that the initialization method $\operatorname{IDI}_{\tau}$ successfully avoids this rank constraint, as detailed in Theorem~\ref{thm:dimup}. The proof is deferred to Appendix~\ref{sec:thmdimpuproof}.
\begin{theorem}
 \label{thm:dimup}
If initializing all weights $\{\theta^{(i)}\}_{i=0}^{2}$ by $\operatorname{IDI}_{1}$, the rank of $\Delta{\theta}^{(1)}$ can attain
 \begin{align}
    \text{rank}(\Delta{\theta}^{(1)}) \geq D_0,
\end{align}
which breaks the rank constraint.
\end{theorem}
Notably, Theorem~\ref{thm:dimup} suggests that an IDInit initialized network can break this constraint through SGD without the help of non-linearity like ReLU which is mentioned as necessary in the prior study~\citep{zhao2022zero}. Specifically, when non-linearity like ReLU is not applied, the rank of the middle weight being limited to $D_0$ only happens at the beginning. After training for several steps, an IDInit-initialized network can break this constraint.

\textbf{Replica Problem.} When recurrently padding the identity matrix, the output features are still replicated. According to \citet{DBLP:conf/icml/BlumenfeldGS20}, such a replica problem can be solved by adding noise to weights.  Inspired by that, we loosen the identity condition to generate $\tau \sim N(\tau, \epsilon_{\tau})$, while keeping most identity. $\epsilon_\tau$ is a small value and set to 1e-6 in this paper.  With this loose condition, IDInit can give additional noise to output features and bring more feature diversity. Profiting from the feature diversity, IDInit therefore can increase the rank values as shown in Figure~\ref{fig:matrix-rank}.

\subsubsection{Patch-Maintain Convolution}
\label{sec:idiconv}

Convolution layers are important structures in deep neural networks.
Here, we will explore an initialization pattern for convolution with the identity transition. A convolution kernel is usually defined as $\ca{C}\in\mathbb{R}^{k\times k \times c_{in} \times c_{out}}$, where $c_{in}$ and $c_{out}$ denote the number of channels of input and output, respectively, and $k$ denotes convolutional kernel size.
Similar to an identity matrix, \citet{zhao2022zero} propose a channel-maintain convolution layer that transits identity by setting 0-filled $\ca{C}$ through $\operatorname{IDI}_{\tau}(\ca{C}_{n, n, :, :})$, where $n \in \mathbb{N}^{+}$ and $k=2n+1$. As a convolutional kernel window size, $k$ is usually an odd number. 
When $c_{in}=c_{out}$, the convolution maintains the identity.  When $c_{in}>c_{out}$ or $c_{in}<c_{out}$, $\ca{C}$ will under-sample and over-sample on an input feature along channel respectively. Keeping identity is usually considered as an efficient way to improve model performance, however, we find that this setting can lead to a fatal performance degeneration (see Sec.~\ref{sec:ablation}).

\paragraph{Patch-Maintain Convolution.} Inspired by \citet{DBLP:conf/cvpr/HanW0GXX20} that enhance model performance by increasing channel diversity, we propose to fuse spatial information by simply reshaping a matrix initialized with $\operatorname{IDI}_{\tau}$.
Specifically, we reshape the convolutional kernel $\ca{C}$ into a matrix $C\in\mathbb{R}^{c_{out}\times kkc_{in}}$.
We initialize $C$ as
\begin{align}
    \operatorname{IDI}_{\tau}(C).
\end{align}
Then by reshaping $C$ into $\ca{C}\in\mathbb{R}^{k\times k \times c_{in} \times c_{out}}$, our initialization for a convolution is completed. This reshaping strategy can shift spatial features, thereby increasing feature diversity. We utilize $\operatorname{IDIC}_{\tau}$ to denote such a reshaping process. A detailed description is in Figure~\ref{fig:idi-conv} in the Appendix.

\subsection{Preserving Zero by Tackling Dead Neurons}
\label{sec:zero-preserving}

Given a residual network formulated by Eq.~(\ref{eq:residual}), prior identity-control initialization~\citep{DBLP:conf/iclr/ZhangDM19,zhao2022zero} set the last transformation in the residual stem to 0, i.e., $\theta^{(i, 0)} = 0$, thereby maintaining an identity as
\begin{align}
    x^{(i+1)} = (I + 0) x^{(i)} = x^{(i)}.
\end{align}
However, the setting can possibly cause dead neurons.

\paragraph{Dead Neuron Problem.} The dead neuron problem occurs when a neuron's weight becomes zero and receives zero gradients, rendering it incapable of updating. This issue is harmful to the training performance of models. Fixup~\citep{DBLP:conf/iclr/ZhangDM19} only uses a multiplier of 1 after $\theta^{(i, 0)}=0$, thereby obtaining non-zero gradients. However, in a realistic implementation of neural networks, the multiplier of Batch Normalization can be set to 0~\citep{DBLP:journals/corr/GoyalDGNWKTJH17}, and down-sampling operation can also cause 0 filled features\footnote{\url{https://github.com/hongyi-zhang/Fixup/blob/master/cifar/models/resnet\_cifar.py}}\footnote{ \url{https://github.com/akamaster/pytorch\_resnet\_cifar10/edit/master/resnet.py}}. Under the implementations, $\theta^{(i, 0)}$ always acquires gradients with 0 values, known as the dead neuron problem, which causes failed weight updating.
\begin{figure}[h]
	\centering
	\subfigure[Weight initialized with numerical value 0.]{
		\includegraphics[width=0.47\textwidth]{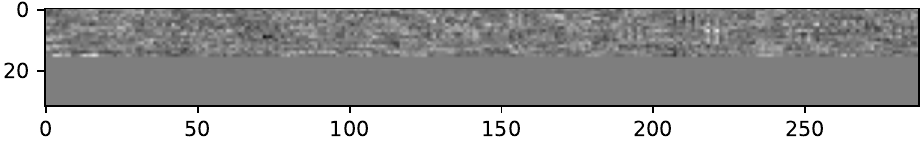}
		\label{fig:conv0}
	}
	\subfigure[Weight initialized with $\operatorname{IDIZ}_{1e-6}$.]{
		\includegraphics[width=0.47\textwidth]{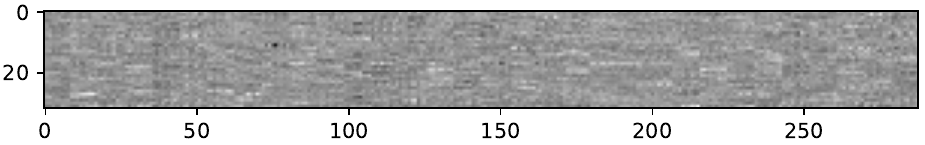}
		\label{fig:convzero}
	}
 \vspace{-2mm}
	\caption{The last weight in a residual block of a trained ResNet. More than half of elements in \subref{fig:conv0} are not trained, which is known as the dead neuron. By contrast, $\operatorname{IDIZ}_{1e-6}$ successfully solves the dead neuron problem and makes all the elements in \subref{fig:convzero} trainable.}
	\label{fig:lastconv}
 \vspace{-4mm}
\end{figure}

Tackling this problem, we generate small values on $\theta^{(i, 0)}$ to assist in training. Recall the goal of identity-control initialization that outputs 0. Therefore, we build a calculation to get the expectation and variance of outputs approaching 0. Considering two i.i.d variables, $v_1$ and $v_2$, whose variances are $\sigma^2(v_1)=\sigma^2(v_2)=\varphi$ and means are $\mu(v_1)=\mu(v_2)=\gamma$, the variable $v=\varepsilon(v1-v2)$ have
\begin{align}
  \begin{cases}
    \mu(v) = 0,\\
	\sigma^2(v) = 2\varphi{\varepsilon}^2,
  \end{cases}
\end{align}
where $\varepsilon$ is a coefficient, and $\sigma^2(v)$ will be limited to 0 when $\varepsilon$ is sufficiently small. Assuming elements of $x^{(i)}$ are i.i.d to each other, by applying subtraction on any two elements, the result has a mean of 0, and a variance related to $\varepsilon$.  We also take $\theta^{(i, 0)} \in \mathbb{R}^{D^{\mathbf{0}}_{i+1} \times D^{\mathbf{0}}_{i}}$ as an instance. At first, we initialize $\theta^{(i, 0)}$ with $\operatorname{IDI}_{\varepsilon}$. Then consider two cases: ({\romannumeral 1}) if $D^{\mathbf{0}}_{i+1} < D^{\mathbf{0}}_{i}$, setting $\theta^{(i)}_{:, D^{\mathbf{0}}_{i+1}+1:D^{\mathbf{0}}_{i}}$ with $\operatorname{IDI}_{-\varepsilon}$; ({\romannumeral 2}) if $D^{\mathbf{0}}_{i+1} \geq D^{\mathbf{0}}_{i}$, set $\theta^{(i)}_{m, j} = -\varepsilon$, when $m\%D^{\mathbf{0}}_{i}=j-1$. Therefore, we can obtain a variance of 0 by setting $\varepsilon$ to a small value. This method is termed as $\operatorname{IDIZ}_{\varepsilon}$, and we illustrate some cases in Figure~\ref{fig:idinit-overivew}. In this paper, we set $\varepsilon=1e-6$ everywhere. As shown in Figure~\ref{fig:lastconv}, $\operatorname{IDIZ}_{1e-6}$ successfully initializes the last weight in a residual block. In addition, we also transform $\operatorname{IDIZ}_{\varepsilon}$ to a convolution form $\operatorname{IDIZC}_{\varepsilon}$ through the patch-maintain scheme.

The IDInit framework is characterized as follows: (1) \textbf{For Non-Residual Networks:} It involves directly applying $\operatorname{IDI}{\tau}$ to fully-connected layers and $\operatorname{IDIC}{\tau}$ to convolutional layers. (2) \textbf{For Residual Networks:} This includes two steps: (i) Implementing $\operatorname{IDI}{\tau}$ and $\operatorname{IDIC}{\tau}$ across all fully-connected and convolutional layers, respectively; (ii) Utilizing $\operatorname{IDIZ}{\varepsilon}$ and $\operatorname{IDIZC}{\varepsilon}$ for fully-connected and convolutional layers positioned at the end of residual blocks, and for the final classification layer.

\section{Experiments}

In this section, we first analyze hyperparameters in Sec.~\ref{sec:hyper}. Then, we implement an ablation experiment in Sec.~\ref{sec:ablation} to show the effect of the proposed two modifications in Sec.~\ref{sec:idinit}. We conduct experiments on residual convolution in Sec.~\ref{sec:residual}. And we conduct image classification on ImageNet in Sec.~\ref{sec:imagenet}. Later we conduct a text classification experiment in Sec.~\ref{sec:determinacy}. At last, we employ a pre-training experiment on the large-scale dataset in Sec.~\ref{sec:wudao} separately. We conduct experiments on non-residual convolution in Sec.~\ref{sec:nonresidual}. We also analyze the variance amplification in Sec.~\ref{sec:varana}, weight distribution in Sec.~\ref{sec:weightdis}, and dynamical isometry in Sec.~\ref{sec:jacana}.

\subsection{Experiment for Hyperparameters}
\label{sec:hyper}

In this experiment, we compare IDInit with Kaiming~\citep{DBLP:conf/iccv/HeZRS15} by analyzing the training hyperparameters, i.e., the weight decay and the learning rate. We use Cifar10. The backbone is ResNet-32, we use SGD with a momentum of 0.9. The batch size is 1024. We train models for 200 epochs. The learning rate is reduced with a cosine function. Each setting is trained 3 times to calculate the standard deviation. More details and results are in Sec.~\ref{sec:hyper-app} of the appendix.

\begin{wrapfigure}[10]{r}{0.5\textwidth}
\small
\begin{center}
    \vspace{-30pt}
    \subfigure[IDInit]{
		\includegraphics[width=0.22\textwidth]{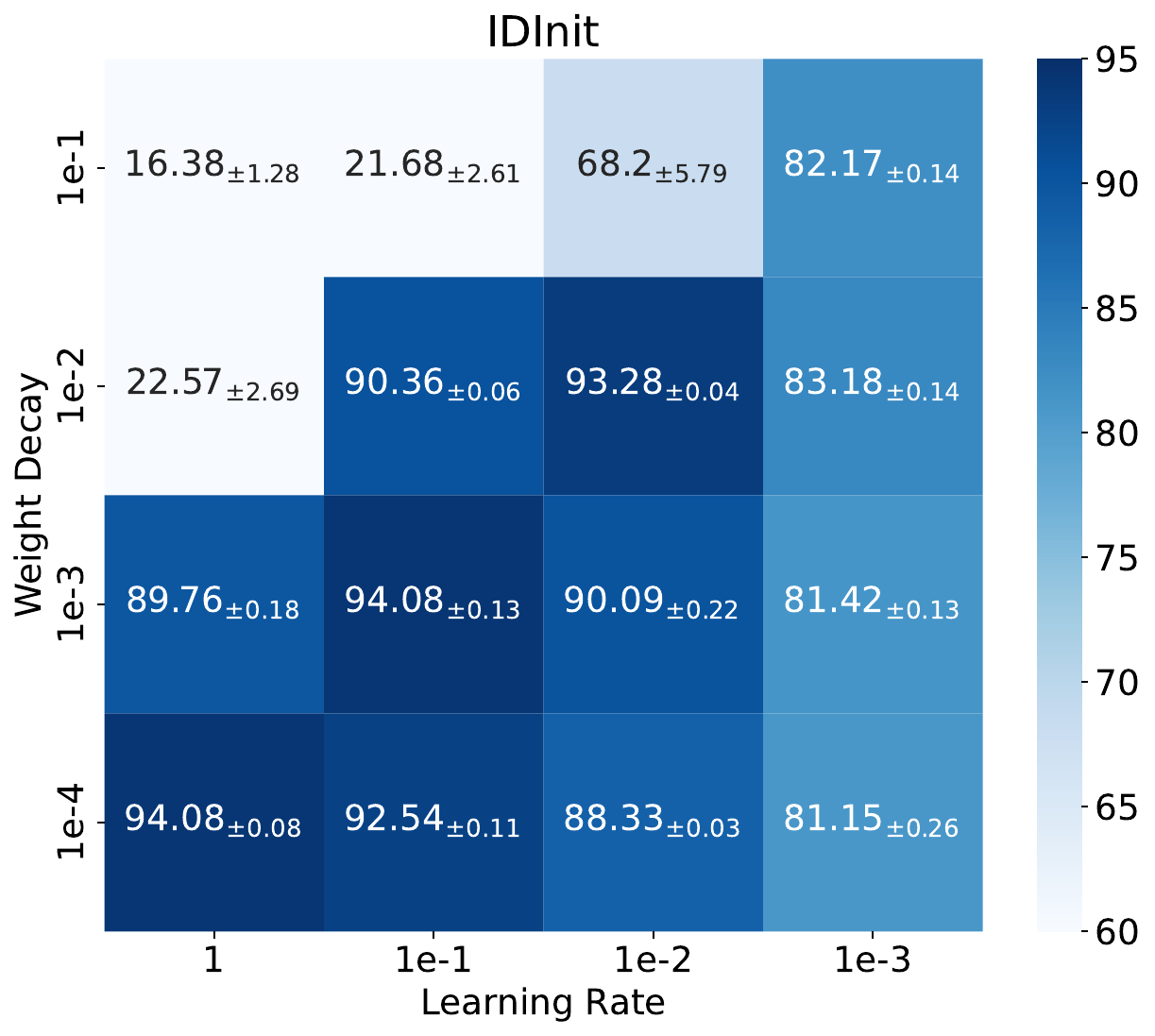}
	}
 \subfigure[Kaiming]{
		\includegraphics[width=0.22\textwidth]{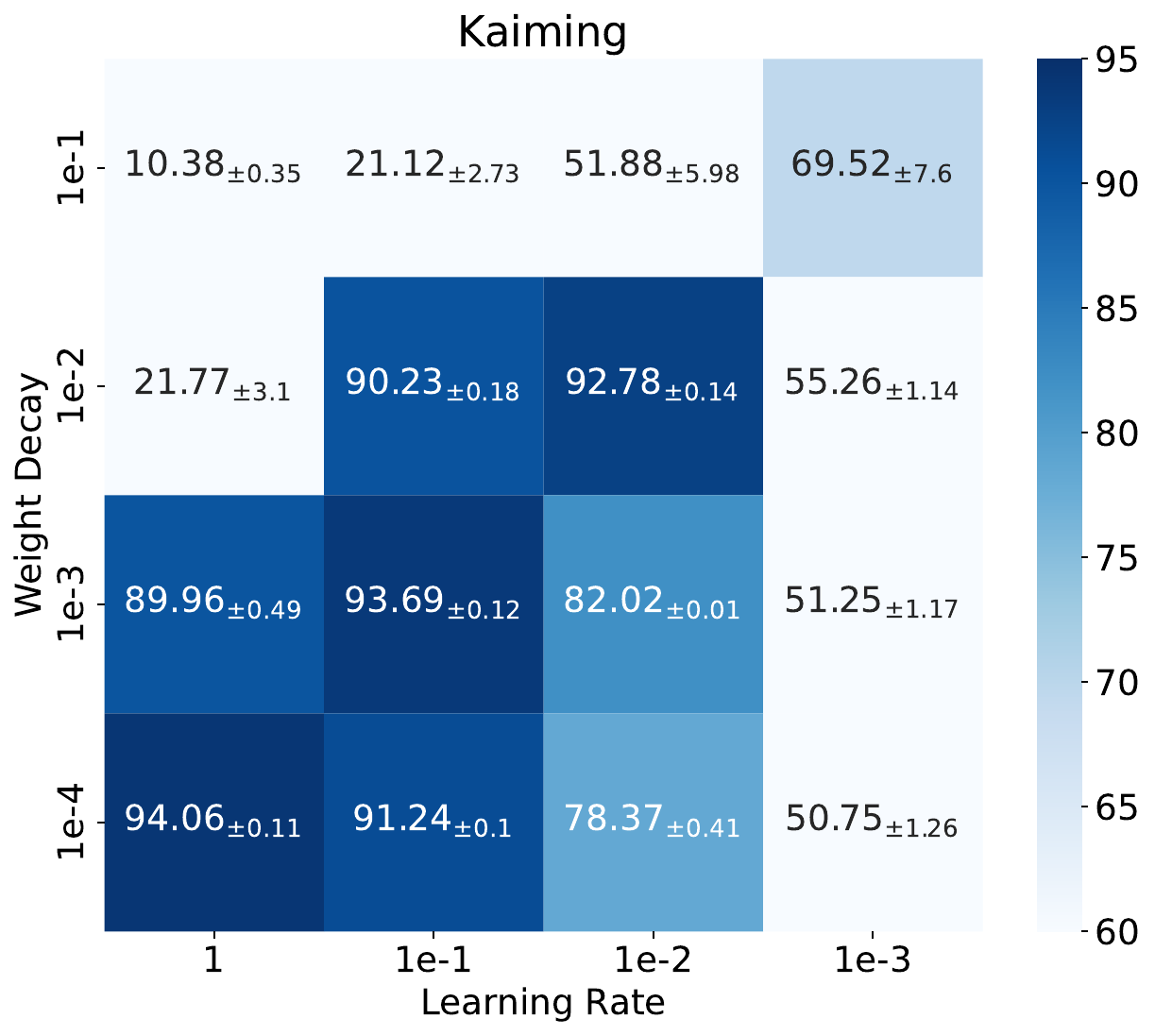}
	}
    \vspace{-10pt}
    \caption{The hyperparameter experiment on Cifar10. IDInit demonstrates superior adaptability across a broader range of training configurations compared to Kaiming initialization, exhibiting notable stability.}
    \label{fig:hypers}
\end{center}
\end{wrapfigure}

As shown in Figure~\ref{fig:hypers}, IDInit achieves a peak accuracy of 94.08\% with a weight decay of 1e-3 and a learning rate of 1e-1. In comparison to Kaiming, IDInit demonstrates superior stability, maintaining high accuracy even when the learning rate is reduced below 1e-1. Overall, IDInit consistently delivers robust performance while maintaining stability, making it a promising candidate for practical applications.

\begin{table*}[t]
\caption{Results on Cifar10. ZerO performs worse for zero down-sampling as mentioned in Sec.~\ref{sec:zero-preserving}. IDInit consistently facilitates rapid convergence when employed with SGD and Adam.}
\label{tbl:rescifar10}
\begin{adjustbox}{width=0.876\width,center}
\begin{tabular}{@{}lcccc@{}}
\toprule
\multirow{2}{*}{Initialization} & \multicolumn{2}{c}{56 Layer (SGD/Adam)}                                         & \multicolumn{2}{c}{110 Layer (SGD/Adam)}                                        \\ \cmidrule(l){2-5} 
                                & Acc.                                & Epochs to 80\% Acc.                       & Acc.                                & Epochs to 80\% Acc.                       \\ \midrule
Zero $\gamma$                   & 92.32$_{\pm0.19}$ / 87.37$_{\pm0.43}$ & 57$_{\pm7}$ / 63$_{\pm4}$                   & 93.07$_{\pm0.28}$ / 88.30$_{\pm0.31}$ & 36$_{\pm2}$ / 56$_{\pm7}$                   \\
ZerO                            & 90.57$_{\pm0.31}$ / 83.53$_{\pm0.42}$ & 57$_{\pm3}$ / 85$_{\pm4}$                   & 91.71$_{\pm0.21}$ / 84.24$_{\pm0.10}$ & 55$_{\pm3}$ / 76$_{\pm2}$                   \\
Fixup                           & 93.24$_{\pm0.82}$ / \underline{89.50}$_{\pm0.18}$ & \underline{31}$_{\pm3}$ / 55$_{\pm3}$       & 93.32$_{\pm0.23}$ / \textbf{90.67}$_{\pm0.12}$ & 33$_{\pm3}$ / 49$_{\pm2}$                   \\
SkipInit                        & 92.29$_{\pm0.30}$ / 85.45$_{\pm0.74}$ & \textbf{26}$_{\pm1}$ / 81$_{\pm3}$          & 92.67$_{\pm0.16}$ / 87.18$_{\pm0.94}$ & \underline{31}$_{\pm5}$ / 70$_{\pm7}$       \\
ReZero                          & 93.06$_{\pm0.54}$ / 89.26$_{\pm0.30}$ & 33$_{\pm2}$ / \underline{44}$_{\pm3}$       & 94.03$_{\pm0.26}$ / 90.25$_{\pm0.20}$ & 35$_{\pm5}$ / \underline{38}$_{\pm3}$       \\
Kaiming                         & \underline{93.36}$_{\pm0.14}$ / 87.55$_{\pm0.32}$ & 34$_{\pm3}$ / 50$_{\pm2}$                   & \textbf{94.06}$_{\pm0.18}$ / 87.89$_{\pm0.41}$ & 33$_{\pm4}$ / 56$_{\pm3}$                   \\
IDInit                         & \textbf{93.41}$_{\pm0.10}$ / \textbf{90.01}$_{\pm0.32}$ & \textbf{26}$_{\pm1}$ / \textbf{34}$_{\pm1}$ & \underline{94.04}$_{\pm0.24}$ / \underline{90.53}$_{\pm0.10}$ & \textbf{27}$_{\pm1}$ / \textbf{36}$_{\pm2}$ \\ \bottomrule
\end{tabular}
\end{adjustbox}
\end{table*}

\subsection{Ablation Experiment}
\label{sec:ablation}

We conduct this experiment to validate the effect of the proposed two improvements. The dataset is Cifar10 and the backbone is ResNet-20~\citep{DBLP:conf/cvpr/HeZRS16}. We run four times following settings: ({\romannumeral 1}) IDInit w/o $\operatorname{IDIC}_{\tau}$ and w/o $\operatorname{IDIZC}_{\varepsilon}$; ({\romannumeral 2}) IDInit w/o $\operatorname{IDIC}_{\tau}$ and w/ $\operatorname{IDIZC}_{\varepsilon}$; ({\romannumeral 3}) IDInit w/ $\operatorname{IDIC}_{\tau}$ and w/o $\operatorname{IDIZC}_{\varepsilon}$; ({\romannumeral 4}) IDInit. For model training for 200 epochs, we employ SGD with a momentum of 0.9, a weight decay of 5e-5, and an initial learning rate of 0.1, which is adjusted using a cosine annealing schedule.
Additional details and results, including the Loose condition, can be found in Sec.~\ref{sec:extend-ablation} of the appendix.

\begin{wraptable}{r}{8.5cm}
\centering
\renewcommand{\arraystretch}{1.1}

\vspace{-20pt}
\caption{Results of the ablation experiment on ResNet-20.}
\label{tbl:ablation}
\scalebox{0.88}{
\begin{tabular}{lcccc}
\toprule
Setting & ({\romannumeral 1}) & ({\romannumeral 2}) & ({\romannumeral 3}) & ({\romannumeral 4}) \\ \hline
Accracy & $87.01_{\pm 0.29}$                                                               &  $92.9_{\pm 0.18}$                                                               &  $90.43_{\pm 0.14}$                                                              &  $\textbf{93.22}_{\pm 0.05}$                                                             \\ \hline
\end{tabular}
}
\vspace{-10pt}
\end{wraptable}

The results are shown in Table~\ref{tbl:ablation}. By applying the identity matrix directly, ({\romannumeral 1}) obtains the lowest accuracy of 87.01\% among all cases. Regarding results of ({\romannumeral 2}) and ({\romannumeral 3}), both the two settings can make significant improvements of nearly 5.89\% and 3.42\% from ({\romannumeral 1}), respectively. And $\operatorname{IDIZC}_{\varepsilon}$ can make a deeper effect than $\operatorname{IDIC}_{\tau}$. Equipping $\operatorname{IDIC}_{\tau}$ and $\operatorname{IDIZC}_{\varepsilon}$, IDInit will improve performance further, which demonstrates our modification is efficient.

\subsection{Image Classification on Cifar10}
\label{sec:residual}
In this experiment, we validate IDInit with the comparison with existing initialization, including (1) Fixup~\citep{DBLP:conf/iclr/ZhangDM19}; (2) SkipInit~\citep{DBLP:conf/nips/DeS20}; (3) ReZero~\citep{DBLP:conf/uai/BachlechnerMMCM21}; (4) Kaiming~\citep{DBLP:conf/iccv/HeZRS15}; (5) Zero $\gamma$ (Setting the scale in Batch Normalization (BN) to 0)~\citep{DBLP:journals/corr/GoyalDGNWKTJH17}; (6) ZerO. We use ResNet-56/110 as backbones on Cifar10. For analyzing convergence, we adopt both SGD and Adam optimizer for updating models. We set SGD, with the momentum 0.9, the weight decay 5e-4, and the learning rate 0.2. For Adam, the learning rate is 0.001, $\beta_1$ is 0.9 and $\beta_2$ is 0.999. The training epoch is 200.

Results are shown in Table~\ref{tbl:rescifar10}. Although ZerO uses the Hadamard matrix to break the rank constraint problem, it can be damaged by zero down-sampling as mentioned in Sec.~\ref{sec:zero-preserving}. Therefore, we reclaim the importance of using $\operatorname{IDIZ}_{\varepsilon}$ and $\operatorname{IDIZC}_{\varepsilon}$ for avoiding such potential damage. Compared with baselines, IDInit derives the best accuracies in most cases. In addition, IDInit can achieve the least epochs to reach 80\% accuracy in all settings, which shows a good convergence ability.

\subsection{Image Classification on ImageNet}
\label{sec:imagenet}

\begin{wrapfigure}[9]{r}{0.5\textwidth}
\small
\begin{center}
    \vspace{-31pt}
    \subfigure[ViT-B/32]{
		\includegraphics[width=0.22\textwidth]{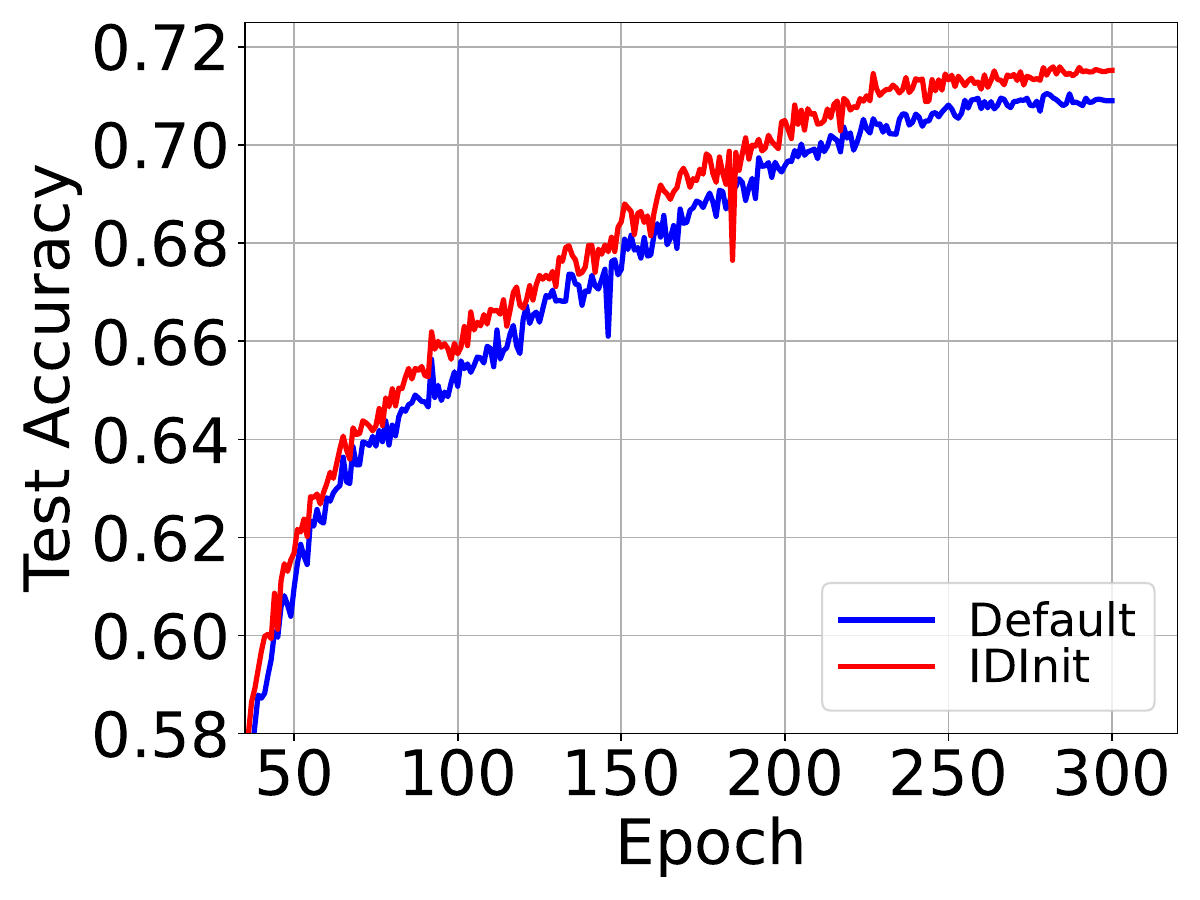}
		\label{fig:vit_val_acc}
	}
     \subfigure[RN-50]{
		\includegraphics[width=0.22\textwidth]{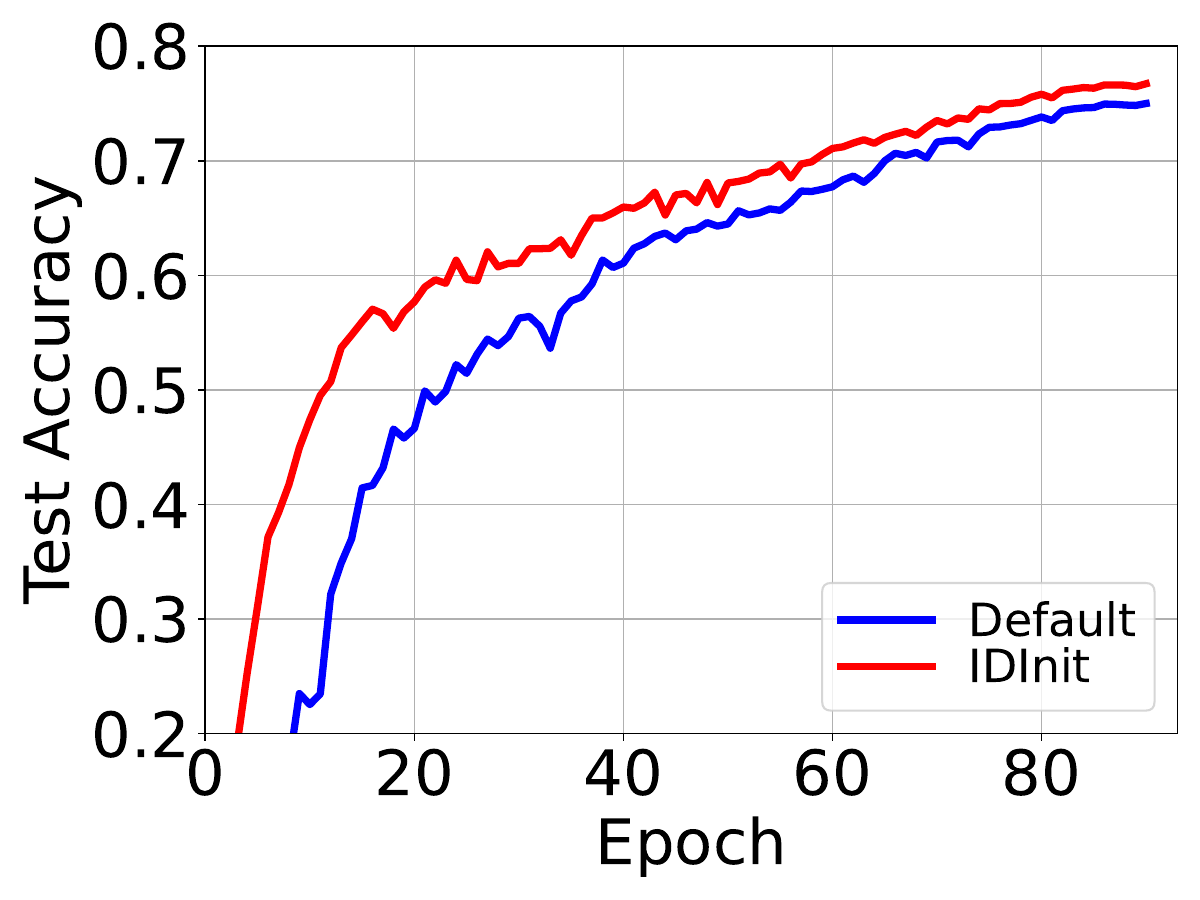}
		\label{fig:r50_val_acc}
	}
    \vspace{-10pt}
    \caption{Results on ImageNet. ``Default" means the default initialization of models.}
    \label{fig:vit-imagenet}
\end{center}
\end{wrapfigure}

We validate ViT-B/32~\citep{DBLP:conf/iclr/DosovitskiyB0WZ21}, ResNet-50/152 (RN-50/152)~\citep{DBLP:conf/cvpr/HeZRS16} and Se-ResNet-50 (SRN-50)~\citep{DBLP:journals/pami/HuSASW20} as backbones on ImageNet in this experiment.
For ViT-B/32, the optimizer is AdamW with a learning rate 1e-3 and a weight decay 5e-2. The training epochs is 300. We use 30 epochs for warm-up. For RN-50/152 and SRN-50, we use SGD with a learning rate 1e-1 and a weight decay 1e-4 for 90-epoch training. We use 9 epochs for warm-up. For all models, the batch size is 1024, and we apply data augment including cutmix~\citep{DBLP:conf/iccv/YunHCOYC19} with $\alpha=1.0$, mixup~\citep{DBLP:conf/iclr/ZhangCDL18} with $\alpha=0.8$, the switching probability is 0.5 and a label smoothing with 0.1. More details and results can be found in Sec.~\ref{sec:app-imagenet}.

\begin{wraptable}{r}{10.cm}
\centering
\renewcommand{\arraystretch}{1.1}

\vspace{-22pt}
\caption{Results on ImageNet. The value in brackets means ``Epochs to 60\% Acc''. On average, IDInit enhances accuracy by 0.55\% compared to the baseline and expedites model convergence by 7.4 epochs.}
\label{tbl:imagenet}
\scalebox{0.75}{
\begin{tabular}{@{}ccccccc@{}}
\toprule
Model   & ViT-B/32   & RN-50 (Adamw) & RN-50      & SRN-50     & RN-152     & Avg ($\Delta$)     \\ \midrule
Default & 71.05 (44) & 76.20 (20)    & 75.70 (38) & 76.30 (32) & 78.76 (28) & 0 (0)      \\ \midrule
IDInit  & 71.60 (42) & 76.71 (14)    & 76.72 (24) & 76.93 (22) & 79.10 (23) & 0.55 (7.4) \\ \bottomrule
\end{tabular}
}
\vspace{-10pt}
\end{wraptable}

\begin{table*}[t]

\caption{Results of text classification on SST2 and TREC-6. The subscript G denotes the embedding layer is initialized by Glove, while W indicates Word2Vec. ``Default" means the default initialization of models, specifically, Kaiming for TextCNN, and Xavier for both TextRNN and Transformer. Fixup is only applicable to the Transformer, as it is specifically designed for residual networks. Std values larger than 1.0 are marked in red. More results can be found in Table~\ref{tbl:textcomplete}.
}
\label{tbl:text}
\begin{adjustbox}{width=0.76\width,center}
\begin{tabular}{@{}llcccc@{}}
\toprule
Datasets                & Init.      & TextCNN\textsubscript{G/W}                            & TextRNN\textsubscript{G/W}                            & Transformer\textsubscript{G/W}  & Average\textsubscript{G/W}                        \\ \midrule
\multirow{5}{*}{SST2}   & Default    & 81.40$_{\pm0.66}$ / 84.56$_{\pm0.43}$                   & 81.69$_{\pm0.30}$ / 84.29$_{\pm0.70}$                   & 80.97$_{\pm\fatal{1.20}}$ / 83.36$_{\pm0.76}$ & 81.35$_{\pm0.72}$ / 84.07$_{\pm0.63}$          \\
                        & Orthogonal & 82.24$_{\pm0.44}$ / 84.37$_{\pm0.38}$                   & 81.86$_{\pm0.55}$ / 84.61$_{\pm0.78}$                   & 82.22$_{\pm0.87}$ / 83.99$_{\pm0.23}$  & 82.11$_{\pm0.62}$ / 84.32$_{\pm0.46}$                  \\
                        & Fixup & -                   & -                   & 78.72$_{\pm0.78}$ / 81.25$_{\pm0.27}$   & -                  \\
                        & ZerO & 82.05$_{\pm0.67}$ / 84.26$_{\pm0.39}$                   & 82.03$_{\pm0.41}$ / 84.80$_{\pm0.64}$                   & 82.28$_{\pm0.81}$ / 82.72$_{\pm0.55}$   & 82.12$_{\pm0.63}$ / 83.93$_{\pm0.53}$                    \\
                        & IDInit     & \textbf{82.60}$_{\pm0.24}$ / \textbf{85.67}$_{\pm0.41}$ & \textbf{82.66}$_{\pm0.16}$ / \textbf{85.49}$_{\pm0.33}$ & \textbf{82.48}$_{\pm0.55}$ / \textbf{84.51}$_{\pm0.24}$   & \textbf{82.58}$_{\pm0.32}$ / \textbf{85.22}$_{\pm0.33}$  \\ \midrule
\multirow{5}{*}{TREC-6} & Default    & 90.80$_{\pm0.94}$ / 92.06$_{\pm1.00}$                   & 86.34$_{\pm\fatal{1.04}}$ / 90.52$_{\pm\fatal{1.54}}$   & 86.68$_{\pm\fatal{2.68}}$ / 89.20$_{\pm\fatal{1.20}}$   & 87.94$_{\pm\fatal{1.55}}$ / 90.59$_{\pm\fatal{1.25}}$    \\
                        & Orthogonal & 90.34$_{\pm0.72}$ / 92.72$_{\pm0.84}$                   & 85.86$_{\pm0.90}$ / 89.88$_{\pm\fatal{1.54}}$           & 86.90$_{\pm\fatal{1.51}}$ / 89.26$_{\pm0.86}$  & 87.70$_{\pm0.71}$ / 90.62$_{\pm0.75}$           \\
                        & Fixup & -                   & -                   & 86.95$_{\pm0.35}$ / 89.35$_{\pm0.53}$   & -                 \\
                        & ZerO & 90.89$_{\pm0.41}$ / 92.90$_{\pm0.50}$                   & \textbf{87.24}$_{\pm0.64}$ / 88.71$_{\pm0.40}$                   & 86.97$_{\pm0.75}$ / 89.38$_{\pm0.64}$  & 88.37$_{\pm0.60}$ / 90.33$_{\pm0.51}$                    \\
                        & IDInit     & \textbf{91.22}$_{\pm0.54}$ / \textbf{92.94}$_{\pm0.48}$ & 87.04$_{\pm0.26}$ / \textbf{90.60}$_{\pm0.58}$ & \textbf{87.32}$_{\pm0.78}$ / \textbf{90.06}$_{\pm0.60}$  & \textbf{88.53}$_{\pm0.53}$ / \textbf{91.20}$_{\pm0.55}$ \\ \bottomrule
\end{tabular}
\end{adjustbox}

\end{table*}

Results are shown in Figure~\ref{fig:vit-imagenet} and Table~\ref{tbl:imagenet}. On three types of networks, i.e., ViT, ResNet and Se-ResNet, and multiple depths, IDInit always achieves faster convergence and better performance than the baseline. When training RN-50 with Adamw, the convergence of IDInit is consistently fast. Compared with RN-50, our initialization shows a faster convergence speed. IDInit has an average improvement of 0.55\%, which is significant to be in practice. This experiment shows the good practicability and promising probability of IDInit, which is beneficial to the artificial intelligence community.

\subsection{Text Classification}
\label{sec:determinacy}

We implement text classification on SST2~\citep{DBLP:conf/emnlp/SocherPWCMNP13} and TREC-6~\citep{DBLP:conf/coling/LiR02}  and select TextCNN~\citep{DBLP:conf/emnlp/Kim14}, TextRNN~\citep{DBLP:conf/aaai/LaiXLZ15} and Transformer \citep{DBLP:conf/nips/VaswaniSPUJGKP17} for comparison. For TextCNN and TextRNN, we use AdaDelta \citep{DBLP:journals/corr/abs-1212-5701} optimizer with a learning rate 1.0 and adopt Adam \citep{DBLP:journals/corr/KingmaB14} for Transformer with a learning rate 1e-4. For the embedding layer, we utilize Glove \citep{DBLP:conf/emnlp/PenningtonSM14} and Word2Vec\citep{DBLP:journals/corr/abs-1301-3781} to initialize the embedding weights. All models are trained up to 10 epochs for 5 times.

As shown in Table~\ref{tbl:text}, all the initialization methods can work normally. Default random initialization obtains the lowest accuracy in most cases on both SST2 and TREC-6. Orthogonal initialization always derives modest results. By contrast to baselines, IDInit can achieve the highest accuracy in all conditions. 
In addition, IDInit always obtains the smallest std values,  showing stable performance.

\subsection{Pre-Training on Language Model}
\label{sec:wudao}

\begin{wrapfigure}[10]{r}{0.38\textwidth}
\small
\begin{center}
    \vspace{-43pt}
    \includegraphics[width=0.36\textwidth]{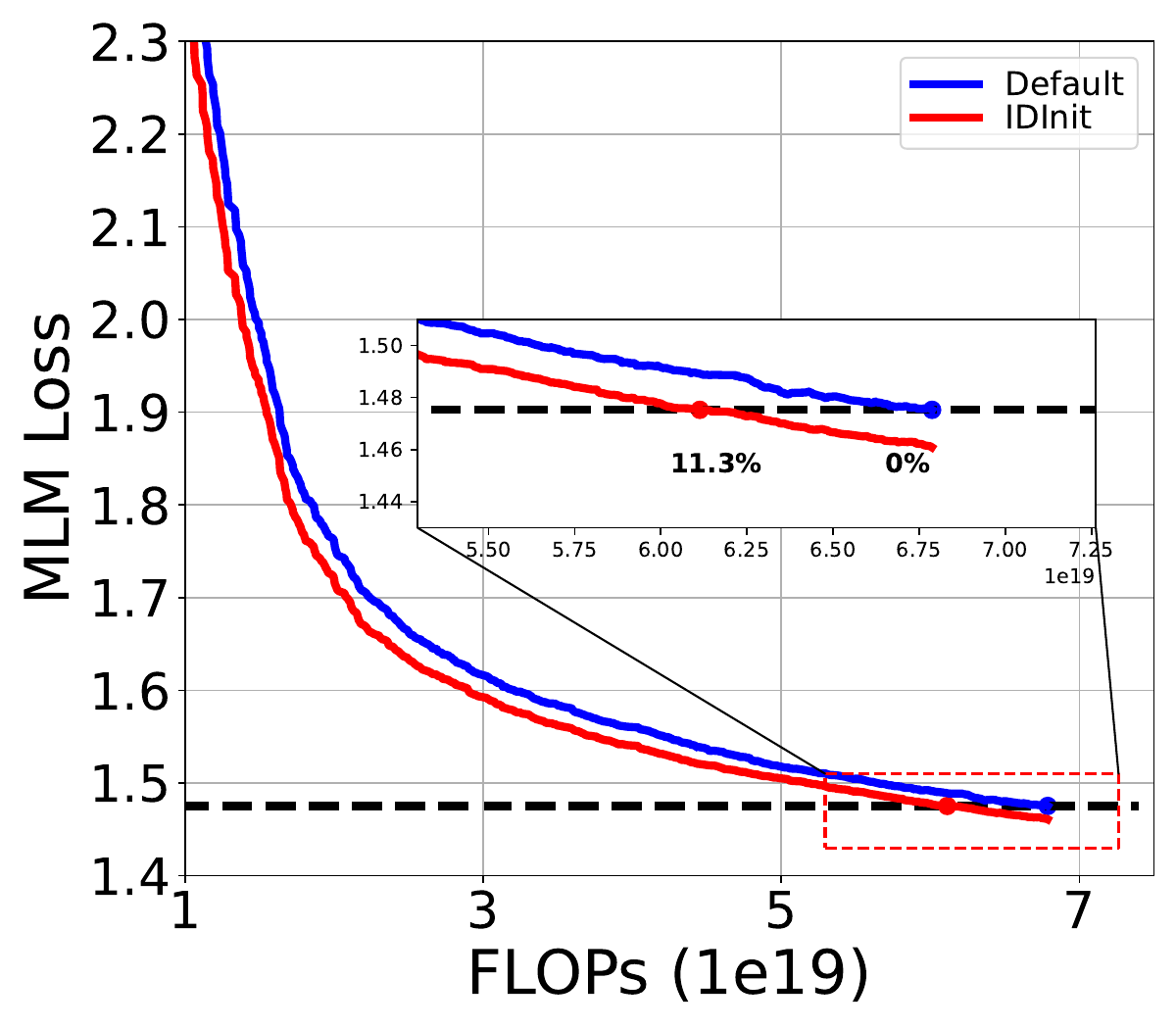}
    \vspace{-12pt}
    \caption{Results of BERT-Base.}
	\label{fig:bert-loss}
\end{center}
\end{wrapfigure}

Pre-training plays an important role in various applications. We conduct the experiment
to show the fast convergence on BERT~\citep{DBLP:conf/naacl/DevlinCLT19}. The dataset is the concatenation of English Wikipedia and Toronto Book Corpus~\citep{DBLP:conf/iccv/ZhuKZSUTF15}.  We train the BERT-Base for 40 epochs with 768 batch size. The optimizer is AdamW with learning rate 1e-4 and weight decay 1e-2. 32 NVIDIA V100s are used.

As shown in Figure~\ref{fig:bert-loss},  ``Default" means the default initialization of BERT-Base. IDInit achieves faster convergence. Specifically, IDInit shows an 11.3\% acceleration ratio in terms of FLOPs. Moreover, IDInit can derive a lower loss of 1.46 in the end. As a result, IDInit is promising used in practice for enhancing convergence ability and performance.

\section{Discussion}

\paragraph{The position of ReLU.} 
\citet{DBLP:conf/nips/PenningtonSG17} pointed out that non-residual networks cannot achieve dynamical isometry when using the ReLU activation function. However, in residual networks, such as $Y=W_2\text{ReLU}(W_1X)+X$, the non-linearity resides within the sub-stem of the residual block. As explored in \citet{DBLP:journals/neco/BartlettHL19} and \citet{DBLP:conf/iclr/HardtM17}, when the weights in the sub-stem are small, residual networks with ReLU can effectively approximate a linear network. This enables the model to follow dynamical isometry, as illustrated in Figure~\ref{fig:Jacbian}, where IDInit results in most $\chi$ values being close to 1. Moreover, \citet{DBLP:conf/aistats/TarnowskiWJTN19} provide a theoretical perspective, suggesting that any activation function within the residual stem can support dynamical isometry. However, this behavior changes if ReLU is instead placed in the main stem, such as in $Y=\text{ReLU}(W_2W_1X+X)$. In this configuration, the network fails to maintain isometry, as noted by \citet{DBLP:conf/nips/PenningtonSG17}. This indicates that placing ReLU in the main stem is not advisable.

\paragraph{The mechanism behind the identical initialization.} Figure~\ref{fig:idi-motivation} highlights the motivation behind the design of IDInit by demonstrating how identity-based initialization preserves structural bias while avoiding pitfalls such as rank constraints. However, the exact mechanism behind this improvement still remains an open question, which is a promising area for future research. Investigating this mechanism further could provide valuable insights and pave the way for the development of more efficient initialization strategies, benefiting the broader research community.

\paragraph{Theoretical analysis regarding the convergence rate.} Theoretical exploration of the convergence rate is a critical yet challenging aspect of initialization methods. The convergence process in deep neural networks is iterative and influenced by numerous factors beyond the initialization method, including the network architecture, optimization algorithm, learning rate, batch size, and data distribution. As this area holds significant importance, further research is necessary to gain deeper insights, which will contribute to a more comprehensive understanding of initialization.

\section{Conclusion}
An efficient initialization approach is crucial for training deep neural networks.
In this paper, we introduce a fully identical initialization (IDInit) that is based on the identity matrix. Addressing the problems encountered when developing IDInit, i.e., dead neurons and performance degeneration, we give two concise solutions, namely using small numerical values to wipe off dead neurons and reshaping an identity-like matrix into a tensor thus increasing feature diversity, leading to a performance improvement. With good performance on wide generality, high stability, and fast convergence, IDInit is promising to be applicable in practice. In the future, we hope that this identical design can motivate the AI community to implement more novel initialization methods.

\section*{Acknowledge}
\label{sec:acknowledge}
We extend our sincere gratitude to all the reviewers whose insightful comments and constructive feedback during previous submissions greatly enhanced the quality of our paper. This research was partially supported by the CFFF platform at Fudan University.

\section*{Impact Statements \& Limitation}
\textbf{Impact Statements.} This paper introduces IDInit, an initialization method designed to enhance stability and convergence of the training process for neural networks. This method is unlikely to have negative societal impacts.

\textbf{Limitation.} While IDInit demonstrates notable advancements in convergence speed and performance enhancement, it faces challenges in converging to ground truths that include negative eigenvalues. However, this drawback can be easily mitigated by incorporating momentum into the optimizer. Given that momentum is a commonly used setting, this limitation can be implicitly resolved as we show in the main context.

\bibliography{example_paper}
\bibliographystyle{iclr2025_conference}

\clearpage
\appendix


\section{IDInit Details}
\label{sec:appdetailidinit}

\subsection{Full IDInit Scheme}
\label{sec:appidinitsheme}
\begin{figure}[h]
	\centering
	\includegraphics[width=0.8\textwidth]{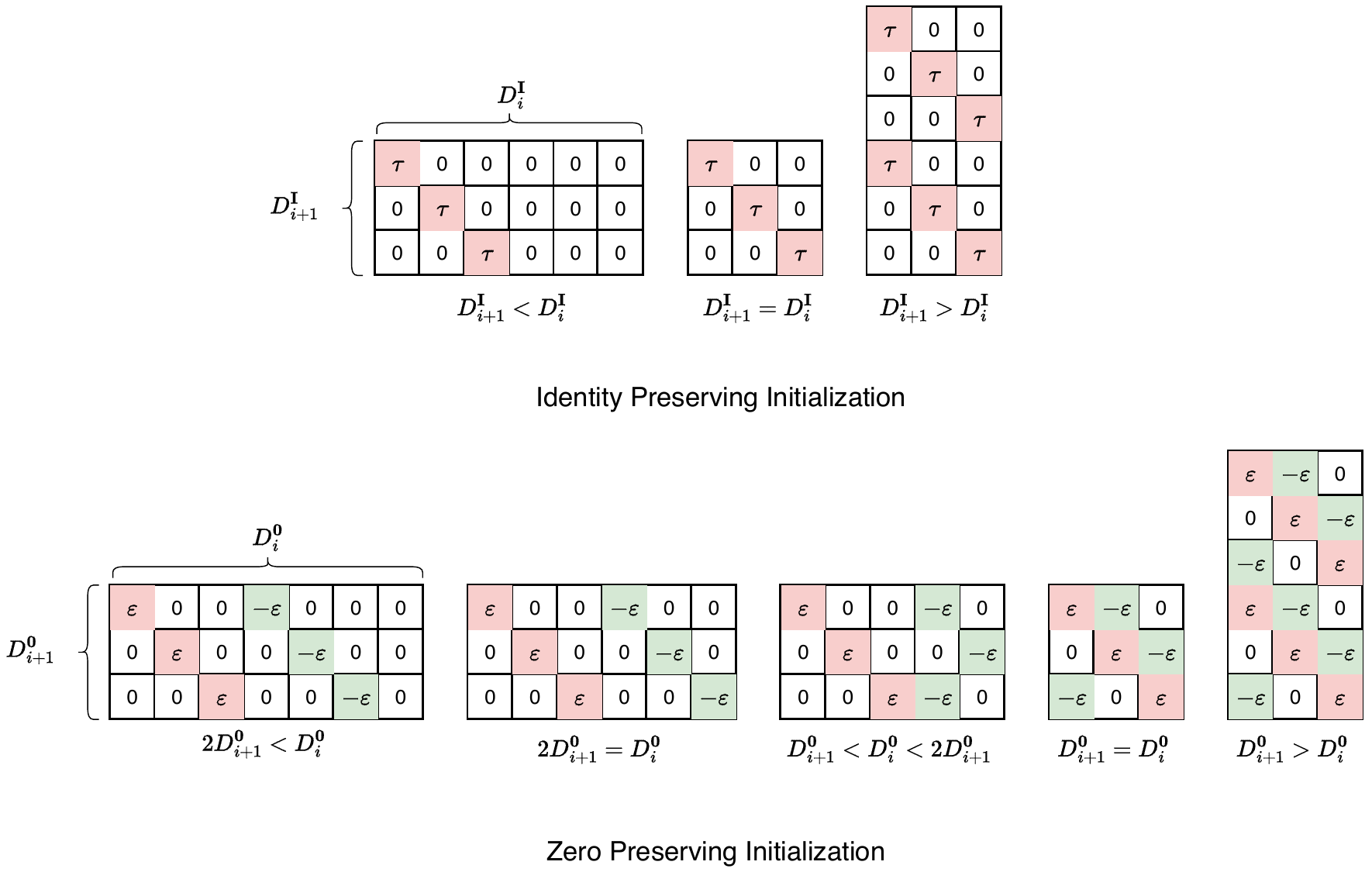}
	\caption{Illustration of IDInit with all conditions.}
	\label{fig:idifully}
\end{figure}

Here, we show the full IDInit scheme in Figure~\ref{fig:idifully}.

\subsection{Analysis on Convergence}
\label{sec:converge-analysis}

The issue of convergence was proposed by  \cite{DBLP:journals/neco/BartlettHL19}. According to their study, when layers in a neural network are initialized using the identity matrix, all the weight matrices of layers will be symmetric at each step of the training process. This persistent symmetry leads to the weights of layers being the same as each other at any step, posing a significant challenge in converging to the ground truth of which eigenvalues with negative values. Our findings indicate that employing a stochastic gradient descent (SGD) approach can effectively break the symmetry which facilitates convergence, and incorporating momentum can further accelerate the convergence process. In this context, we provide formal proof demonstrating that SGD with momentum can alleviate the convergence issue.

\begin{proof}

First of all, we present a training case for a single-layer network expressed as $y=\theta x$, where $x\in \mathbb{R}^d$ represents the input, $y\in \mathbb{R}^d$ denotes the output, and $\theta\in \mathbb{R}^{d\times d}$ is the weight matrix. The weight matrix $\theta$ is initialized to the identity matrix $I$, denoted as $\theta^{(0)}=I$. For our loss function, we employ the Mean Squared Error (MSE) and a learning rate denoted by $\eta$. Consider two training pairs $\{x_1, y_1\}$ and $\{x_2, y_2\}$ sampled from the same dataset $\mathcal{D}$. The network is initially trained with $\{x_1, y_1\}$, and trained with $\{x_2, y_2\}$ in the next step.

In the first step, we can get the prediction as
\begin{align}
\centering
    \hat{y}_1 = \theta^{(0)} x_1.
\end{align}
The updated $\theta^{(1)}$ can be derived by
\begin{align}
    \Delta \theta^{(0)} &= (\hat{y}_1 - y_1)x_1^T = ( \theta^{(0)}x_1 - y_1)x_1^T = (x_1 - y_1)x_1^T, \notag \\
    \theta^{(1)} &= \theta^{(0)} - \eta\Delta \theta^{(0)} = \theta^{(0)} - \eta (x_1 - y_1)x_1^T = I - \eta (x_1 - y_1)x_1^T.
\end{align}
Therefore, in the second step, the gradient $\Delta \theta^{(1)}$ can be calculated as
\begin{align}
\centering
    \Delta \theta^{(1)} &= (\hat{y}_2 - y_2)x_2^T, \notag \\
     &= (\theta^{(1)}x_2 - y_2)x_2^T, \notag \\
     &= ((I - \eta (x_1 - y_1)x_1^T)x_2 - y_2)x_2^T, \notag \\
     &= x_2x_2^T - \eta x_1x_1^Tx_2x_2^T + \eta y_1x_1^Tx_2x_2^T - y_2x_2^T.
\end{align}
While $x_2x_2^T$ is symmetric, $x_1x_1^Tx_2x_2^T$, $y_1x_1^Tx_2x_2^T$, and $y_2x_2^T$ can be asymmetric. To calculate the magnitude of the asymmetry in $\Delta \theta^{(1)}$, letting $\Omega = -\eta x_1x_1^Tx_2x_2^T + \eta y_1x_1^Tx_2x_2^T - y_2x_2^T$ denotes the asymmetric component, the magnitude of asymmetry that can be calculated as $\mathbb{E}(||\Omega - \Omega^T||_F^2)$. Assuming \( x_1, x_2, y_1, y_2 \in \mathbb{R}^d \) are random vectors with entries that are i.i.d. Gaussian random variables, following \( N(0, \sigma^2) \), then the magnitude of asymmetry is bounded as
\begin{align}
   4\eta^2d^3\sigma^8-4\eta^2d^2\sigma^8+2d^2\sigma^4 \leq \mathbb{E}(||\Omega - \Omega^T||_F^2) \leq 6\eta^2d^3\sigma^8 + 3d^2\sigma^4.
\end{align}
The proof can be found in Sec.~\ref{sec:appasymmetry}. As $\eta$ is usually $1e-1$, and both training pairs $\{x_1, y_1\}$ and $\{x_2, y_2\}$ can be generally normalized to $\mathcal{N}\sim(0, 1)$, thereby, the symmetry of the weight can be sufficiently influenced as 
\begin{align}
    \theta^{(2)} &= \theta^{(1)} - \eta \Delta \theta^{(1)}.
\end{align}
When introducing a momentum $m^{(0)}$ initialized to $\Delta \theta^{(0)}$, assuming the coefficient of $m$ is $\gamma$, $\theta^{(2)}$ will be updated as
\begin{align}
\centering
    m^{(1)} &= \gamma m^{(0)} + \eta \Delta \theta^{(1)}, \notag \\ 
    \theta^{(2)} &= \theta^{(1)} - m^{(1)} = \theta^{(1)} - \gamma m^{(0)} - \eta \Delta \theta^{(1)}.
\end{align}
Therefore, momentum can promote the weight to become asymmetric by accumulating the asymmetry of gradients in steps and impact more when samples are increased.

As for networks of multiple layers, when their layers are asymmetric, each layer can be updated differently which breaks the convergence problem caused by the same gradients in each step (which is stated in Lemma 5 of \citet{DBLP:journals/neco/BartlettHL19}).
\end{proof}

\begin{figure}[t]
	\centering
 \subfigure[GD w/o momentum]{
		\includegraphics[width=0.38\textwidth]{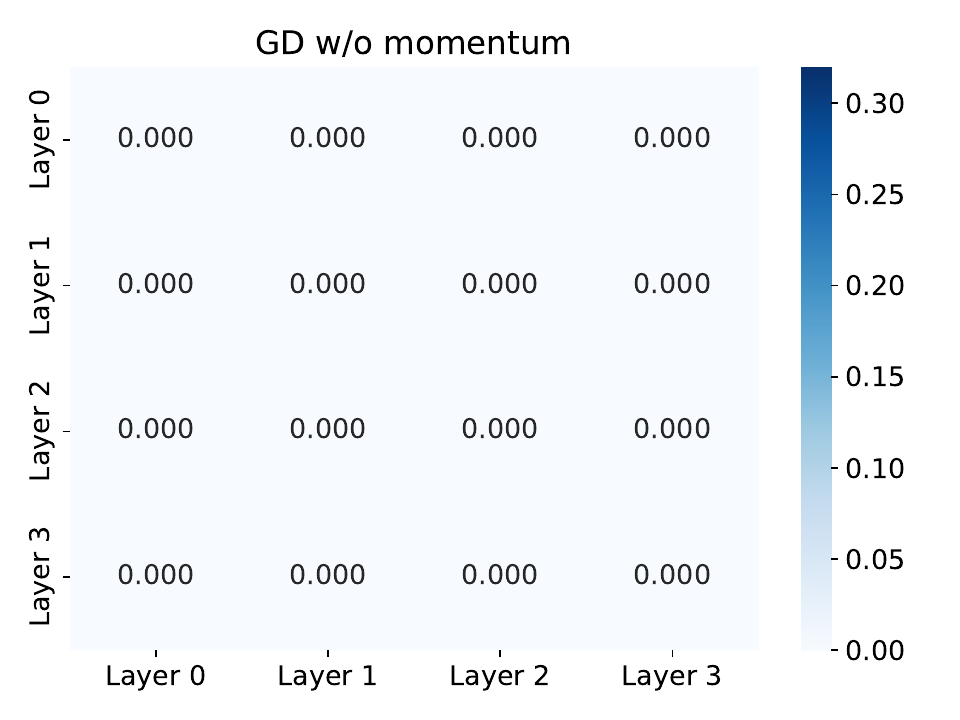}
	}~~~~~~~~~~~~~~~
 \subfigure[GD w/ momentum]{
		\includegraphics[width=0.38\textwidth]{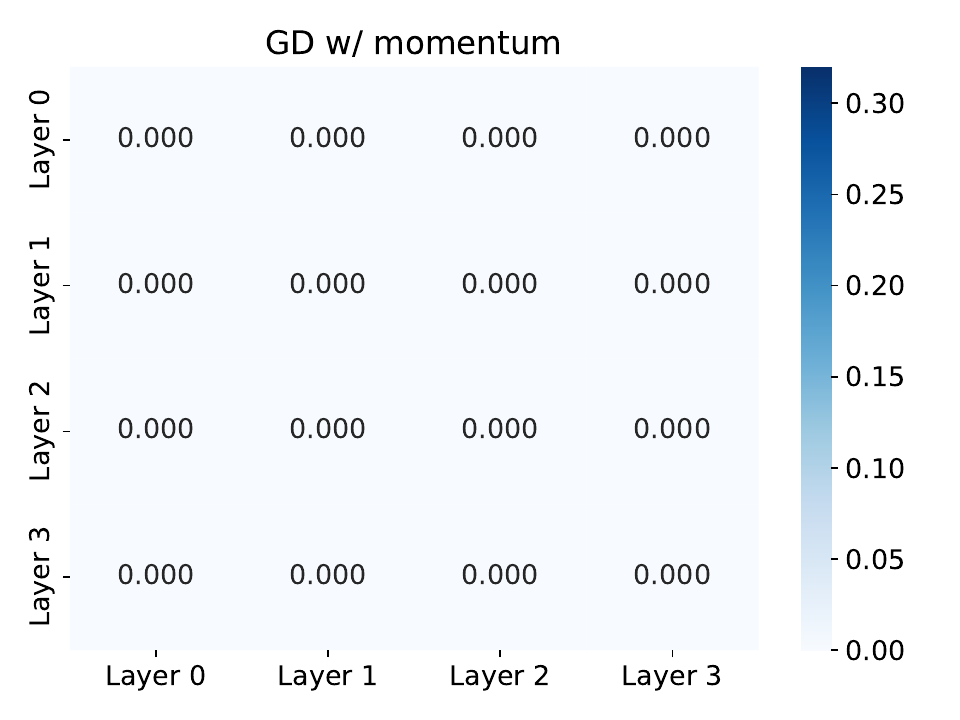}
	}
 
	\subfigure[SGD w/o momentum]{
		\includegraphics[width=0.38\textwidth]{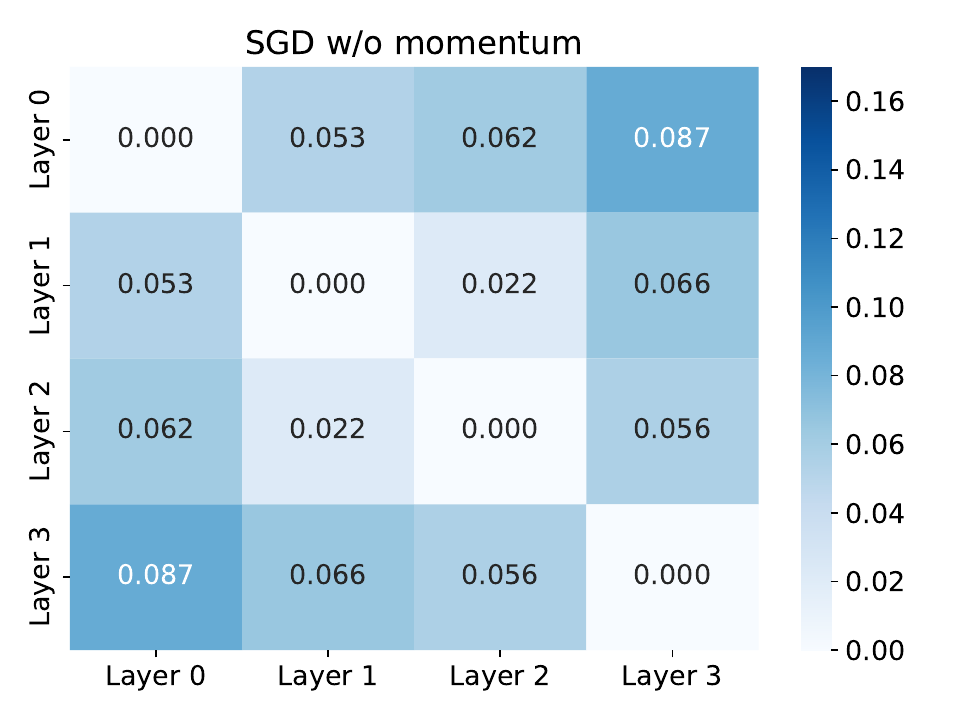}
	}~~~~~~~~~~~~~~~
 \subfigure[SGD w/ momentum]{
		\includegraphics[width=0.38\textwidth]{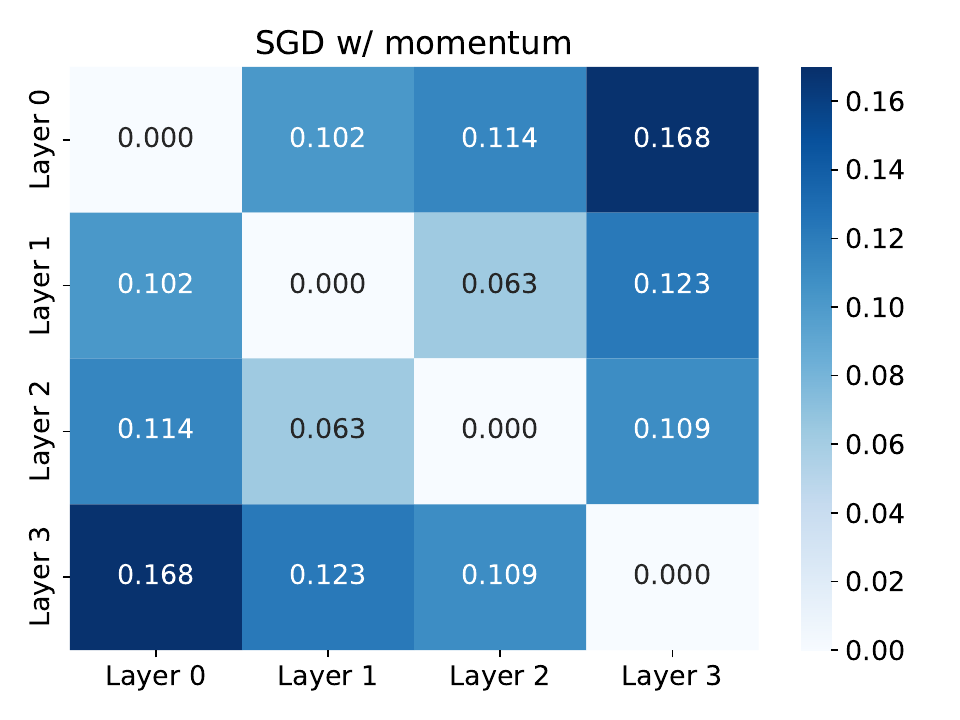}
	}
 
	\caption{The distance between two layers in a 4-layer network after training. In this experiment, we set the target matrix as $-I\in \mathbb{R}^{10\times10}$. The weights are $W_0, W_1, W_2, W_3\in \mathbb{R}^{10\times 10}$. We randomly generated 4000  data pairs $\{X_i, Y_i\}$ by $Y_i = -IX_i + \xi$, where $X_i, Y_i \in \mathbb{R}^{10}$, and $\xi$ is noise with mean 0 and std 1e-2. We use 2000 samples for training the network. We use the other 2000 samples for testing. Batch size is 4. Mean squared error (MSE) is used as the loss function. We calculate the distance by averaging the absolute value from the difference value of two layers. Layers trained using SGD display distinct differences from one another, and the incorporation of momentum significantly increases these differences, thereby accelerating the convergence speed.}
	\label{fig:differlayers}
\end{figure}

This proof primarily demonstrates that SGD with momentum can effectively resolve the issue of layers being the same in networks initialized with the identity matrix during training, which facilitates the convergence process. As illustrated in Figure~\ref{fig:differlayers}, it is evident that layers trained using SGD are different from each other, with the momentum component amplifying the degree of this difference. By theoretically and empirically demonstrating that SGD with momentum can efficiently address this convergence problem, we hope this finding can offer valuable insights for the research community, encouraging further investigation into identity initialization and its significant role in model training.

\subsection{Analysis on Asymmetry}
\label{sec:appasymmetry}
In this section, we analyze the magnitude of asymmetry in the gradient. 

\textbf{Setup and Target.} Here, we assume \( x_1, x_2, y_1, y_2 \in \mathbb{R}^d \) are random vectors with entries that are i.i.d. Gaussian random variables, following \( N(0, \sigma^2) \). According to Eq.~(\ref{eq:step2-grad}), the asymmetry in the gradient arises from: 
\begin{align}
    \Omega = -\eta x_1x_1^Tx_2x_2^T + \eta y_1x_1^Tx_2x_2^T - y_2x_2^T.
\end{align}

Our target is to compute the magnitude of asymmetry that can be calculated as
\begin{align}
    \mathbb{E}&(||\Omega - \Omega^T||_F^2) \notag \\
=&\mathbb{E}\{||[-\eta x_1x_1^Tx_2x_2^T+(\eta x_1x_1^Tx_2x_2^T)^T]+ [\eta y_1x_1^Tx_2x_2^T-(\eta y_1x_1^Tx_2x_2^T)^T] \notag \\
&+ [-y_2x_2^T+(y_2x_2^T)^T]||_F^2\} 
\end{align}

\textbf{Lower Bound.} Introducing substitutions $u=y_1-x_1$, and $s=x_1^Tx_2=x_2^Tx_1$, we rewrite:

\begin{align}
\mathbb{E}(||\Omega - \Omega^T||_F^2) = \mathbb{E}\{||\eta s(ux_2^T-x_2u^T)-(y_2x_2^T-x_2y_2^T)||_F^2\},
\end{align}

Let $w=\eta su-y_2$, then:
\begin{align}
\mathbb{E}(||\Omega - \Omega^T||_F^2) &= \mathbb{E}\{||wx_2^T-x_2w^T||_F^2\}, \\
&= \mathbb{E}\{2(||w||^2 ||x_2||^2 - (w^T x_2)^2)\}, \\
&= 2 \left( \mathbb{E}[||w||^2] \mathbb{E}[||x_2||^2] - \mathbb{E}[(w^T x_2)^2] \right),
\end{align}

Expanding and computing expectations:
\begin{align}
\mathbb{E}(||\Omega - \Omega^T||_F^2) &\geq 2 ((\eta^2(2d^2-d)\sigma^6+d\sigma^2)d\sigma^2 - \eta^2d^2\sigma^8), \\
&= 4\eta^2d^3\sigma^8-4\eta^2d^2\sigma^8+2d^2\sigma^4.
\end{align}

\textbf{Upper Bound.} We derive the upper bound as
\begin{align}
\mathbb{E}&(||\Omega - \Omega^T||_F^2) \\
=&\mathbb{E}\{||[-\eta x_1x_1^Tx_2x_2^T+(\eta x_1x_1^Tx_2x_2^T)^T]+ [\eta y_1x_1^Tx_2x_2^T-(\eta y_1x_1^Tx_2x_2^T)^T] \notag \\
&+ [-y_2x_2^T+(y_2x_2^T)^T]||_F^2\}\\
&\text{According to Relaxed Triangle Inequality, there is} \notag \\
\leq& 3\{\eta^2\mathbb{E}[||- x_1x_1^Tx_2x_2^T+ (x_1x_1^Tx_2x_2^T)^T||_F^2]+\eta^2\mathbb{E}[|| y_1x_1^Tx_2x_2^T- ( y_1x_1^Tx_2x_2^T)^T||_F^2] \notag \\
&+\mathbb{E}[||-y_2x_2^T+(y_2x_2^T)^T||_F^2]\} \\
 \leq& 3(\eta^2d^3\sigma^8 + \eta^2d^3\sigma^8+d^2\sigma^4) \\
 =& 6\eta^2d^3\sigma^8 + 3d^2\sigma^4
\end{align}

This shows that a higher learning rate promotes greater asymmetry, further explaining the observed differences. However, a high learning rate can affect training stability. Therefore, while using a higher learning rate to reduce symmetry, it is crucial to carefully select its magnitude to maintain stability.

\subsection{Proof for Theorem~\ref{thm:dimup}.}
\label{sec:thmdimpuproof}

\begin{proof}

Consider a network with a single hidden layer (i.e. $L = 3$) and a batch of linearly independent samples, $x_1^{(0)} = \big\{x_1^{(0, 1)}, \ldots, x_1^{(0, N)}\big\},$ with $N = D_0.$ Using $\Pi_1 = \sum_{i = 1}^N \frac{\partial \mathcal{L}}{\partial x_1^{(3, i)}} \times x_1^{(0, i)},$ the gradients for the first update step can be written as
\begin{align}
\frac{\partial \mathcal{L}}{\partial \theta^{(0)}} &= \begin{pmatrix}\Pi_1 \\ \boldsymbol{0} \end{pmatrix} & \frac{\partial \mathcal{L}}{\partial \theta^{(1)}} &= \begin{pmatrix}\Pi_1 & \Pi_1 \\ \boldsymbol{0} & \boldsymbol{0} \end{pmatrix} & \frac{\partial \mathcal{L}}{\partial \theta^{(2)}} &= \begin{pmatrix}\Pi_1 & \Pi_1 \end{pmatrix}.
\end{align}
After updating the weights with learning rate $\eta > 0$, we have
\begin{align}
\theta^{(0)} &= \begin{pmatrix}\boldsymbol{I} - \eta\,\Pi_1 \\ \boldsymbol{I} \end{pmatrix} & \theta^{(1)} &= \begin{pmatrix}\boldsymbol{I} - \eta\,\Pi_1 & -\eta\,\Pi_1 \\ \boldsymbol{0} & \boldsymbol{I} \end{pmatrix} & \theta^{(2)} &= \begin{pmatrix}\boldsymbol{I} - \eta\,\Pi_1 & -\eta\,\Pi_1 \end{pmatrix}.
\end{align}
As a result, the gradients of $\theta^{(1)}$ for the second update with a second batch of linearly independent samples $x_2^{(0)} = \big\{x_2^{(0, 1)}, \ldots, x_2^{(0, N)}\big\},$ are given by 
\begin{align}
\frac{\partial \mathcal{L}}{\partial \theta^{(1)}} &= \sum_{i=1}^N \big({\theta^{(0)}} \cdot x_2^{(0, i)}\big) \times \bigg(\frac{\partial \mathcal{L}}{\partial x_2^{(3, i)}} \cdot {\theta^{(2)}}^\mathsf{T}\bigg) \\ &= \begin{pmatrix} (\boldsymbol{I} - \eta\, \Pi_1^\mathsf{T})\, \Pi_2\, (\boldsymbol{I} - \eta\, \Pi_1^\mathsf{T}) & (\boldsymbol{I} - \eta\, \Pi_1^\mathsf{T})\, \Pi_2 \\ -\eta\, \Pi_1^\mathsf{T}\, \Pi_2\, (\boldsymbol{I} - \eta\, \Pi_1^\mathsf{T}) & -\eta\, \Pi_1^\mathsf{T}\, \Pi_2 \end{pmatrix},
\end{align}
with $\Pi_2 = \sum_{i = 1}^N \frac{\partial \mathcal{L}}{\partial x_2^{(3, i)}} \times x_2^{(0, i)}.$ Using the gradients of the second batch to update the parameters with the same learning rate, we obtain
$$\theta^{(1)} = \begin{pmatrix}\boldsymbol{I} - \eta\,\Pi_1 - \eta\, (\boldsymbol{I} - \eta\, \Pi_1^\mathsf{T})\, \Pi_2\, (\boldsymbol{I} - \eta\, \Pi_1^\mathsf{T}) & -\eta\,\Pi_1 - \eta\, (\boldsymbol{I} - \eta\, \Pi_1^\mathsf{T})\, \Pi_2 \\ \eta^2\, \Pi_1^\mathsf{T}\, \Pi_2\, (\boldsymbol{I} - \eta\, \Pi_1^\mathsf{T}) & \boldsymbol{I} + \eta^2\, \Pi_1^\mathsf{T}\, \Pi_2 \end{pmatrix}.
$$
Consequently, the difference of the weights after two updates to the initial value, $\boldsymbol{I}$, is given by
$$\Delta \theta^{(1)} = \begin{pmatrix} -\eta\,\Pi_1 - \eta\, (\boldsymbol{I} - \eta\, \Pi_1^\mathsf{T})\, \Pi_2\, (\boldsymbol{I} - \eta\, \Pi_1^\mathsf{T}) & -\eta\,\Pi_1 - \eta\, (\boldsymbol{I} - \eta\, \Pi_1^\mathsf{T})\, \Pi_2 \\ \eta^2\, \Pi_1^\mathsf{T}\, \Pi_2\, (\boldsymbol{I} - \eta\, \Pi_1^\mathsf{T}) & \eta^2\, \Pi_1^\mathsf{T}\, \Pi_2 \end{pmatrix}.
$$

Assuming that the gradients $\frac{\partial \mathcal{L}}{\partial x_1^{(3)}}$ and $\frac{\partial \mathcal{L}}{\partial x_2^{(3)}}$ are also linearly independent, $\operatorname{rank}(\Pi_1) = \operatorname{rank}(\Pi_2) = D_0$. Due to Sylvester's rank inequality, we can conclude that also $\operatorname{rank}(\Pi_1 \, \Pi_2) = D_0$. As a result, the lower-right part of the difference has rank $D_0$, from which we can conclude that $\operatorname{rank}(\Delta \theta^{(1)}) \geq D_0$.
\end{proof}

\subsection{Implementing IDInit on Attention Layer in Transformer}
In this part, we show the way to initialize the attention layer with IDInit. Prior to that, formulating an attention layer as
\begin{align}
    \operatorname{Att}(Q, K, V) = \operatorname{softmax}(\frac{QW^QW^KK}{\sqrt{d}})VW^VW^O,
\end{align}
where $Q$ is the query matrix, $K$ means the key matrix, $V$ denotes the value matrix, $W^Q$, $W^K$ and $W^V$ represents the weights for $Q$, $K$, and $V$ respectively, and $W^O$ is the output transformation. Following the instruction of IDInit in Sec.~\ref{sec:idinit}, we firstly use $\operatorname{IDI}_{\tau}$ to initialize $W^Q$, $W^K$, $W^V$ and $W^O$. And then, we use $\operatorname{IDIZ}_{\varepsilon}$ to initialize the last fully-connected layer $W^O$. The $\tau$ and $\varepsilon$ are consistently set with the paper content to 1 and 1e-6, respectively.

\subsection{Details of Patch-Maintain Convolution}
\begin{figure*}[h]
	\centering
	\includegraphics[width=0.85\textwidth]{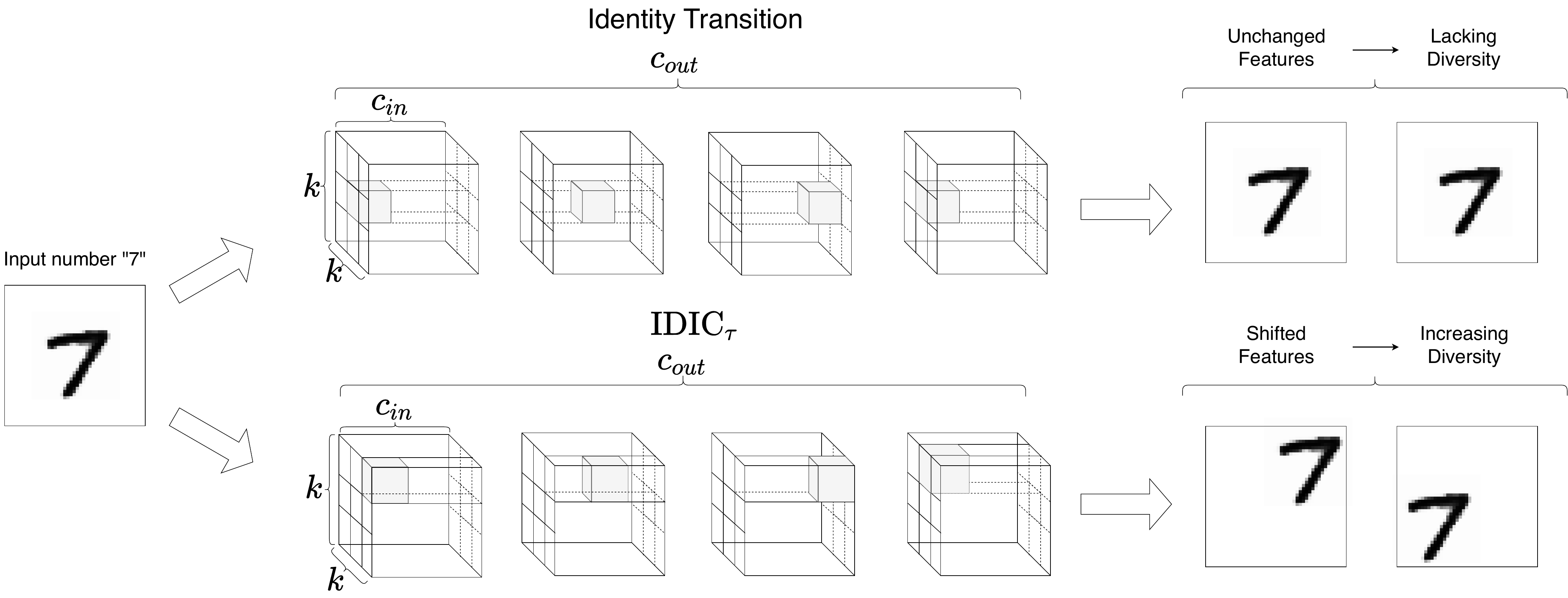}
	\caption{A case of number ``7" on Identical Convolution Layer. The upper sub-figure maintains the identity transition. The under sub-figure is $\operatorname{IDIC}_{\tau}$ initialization that shifts features for increasing diversity. More feature diversity from $\operatorname{IDIC}_{\tau}$ is beneficial for improving model performance.}
	\label{fig:idi-conv}
\end{figure*}

We illustrate the figure to show the comparison between channel-maintain convolution and patch-maintain convolution in Figure~\ref{fig:idi-conv}.

\section{Detailed Settings of Experiments}
\label{sec:appdetailexp}

In this paper, for ReLU activated networks, $\tau$ is set to $\sqrt{2}$ for the first layer in a network and 1 for other $\operatorname{IDI}_{\tau}/\operatorname{IDIC}_{\tau}$ initializing layers, while for tanh-activated networks, all $\operatorname{IDI}_{\tau}$ is set to 1, and $\varepsilon$ is $1e-6$ for all $\operatorname{IDIZ}_{\varepsilon}/\operatorname{IDIZC}_{\varepsilon}$ initializing layers.

\subsection{Experiment for Hyperparameters}
\label{sec:hyper-app}
In this experiment, we compare IDInit with other initialization methods, including (1) Fixup; (2) ReZero; (3) Kaiming; and (4) Zero, by analyzing the training hyperparameters, i.e., the weight decay and the learning rate. We use Cifar10. The backbone is ResNet-32, we use SGD with a momentum of 0.9. The batch size is 1024. We train models for 200 epochs. The learning rate is reduced with a cosine function. Each setting is trained 3 times to calculate the standard deviation.

\begin{figure}[h]
	\centering
        \subfigure[IDInit]{
		\includegraphics[width=0.3\textwidth]{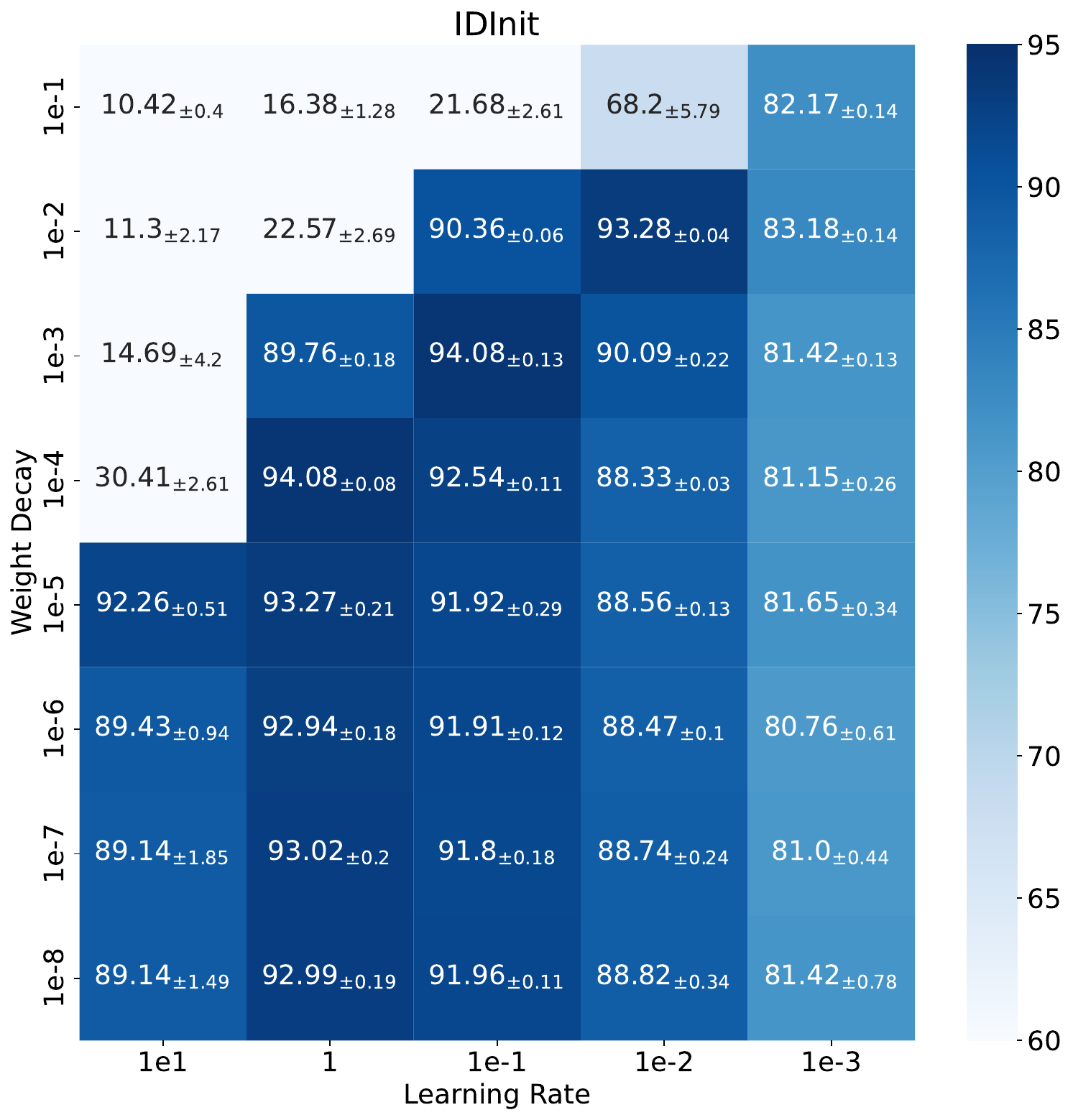}
	}
	\subfigure[ZerO]{
		\includegraphics[width=0.3\textwidth]{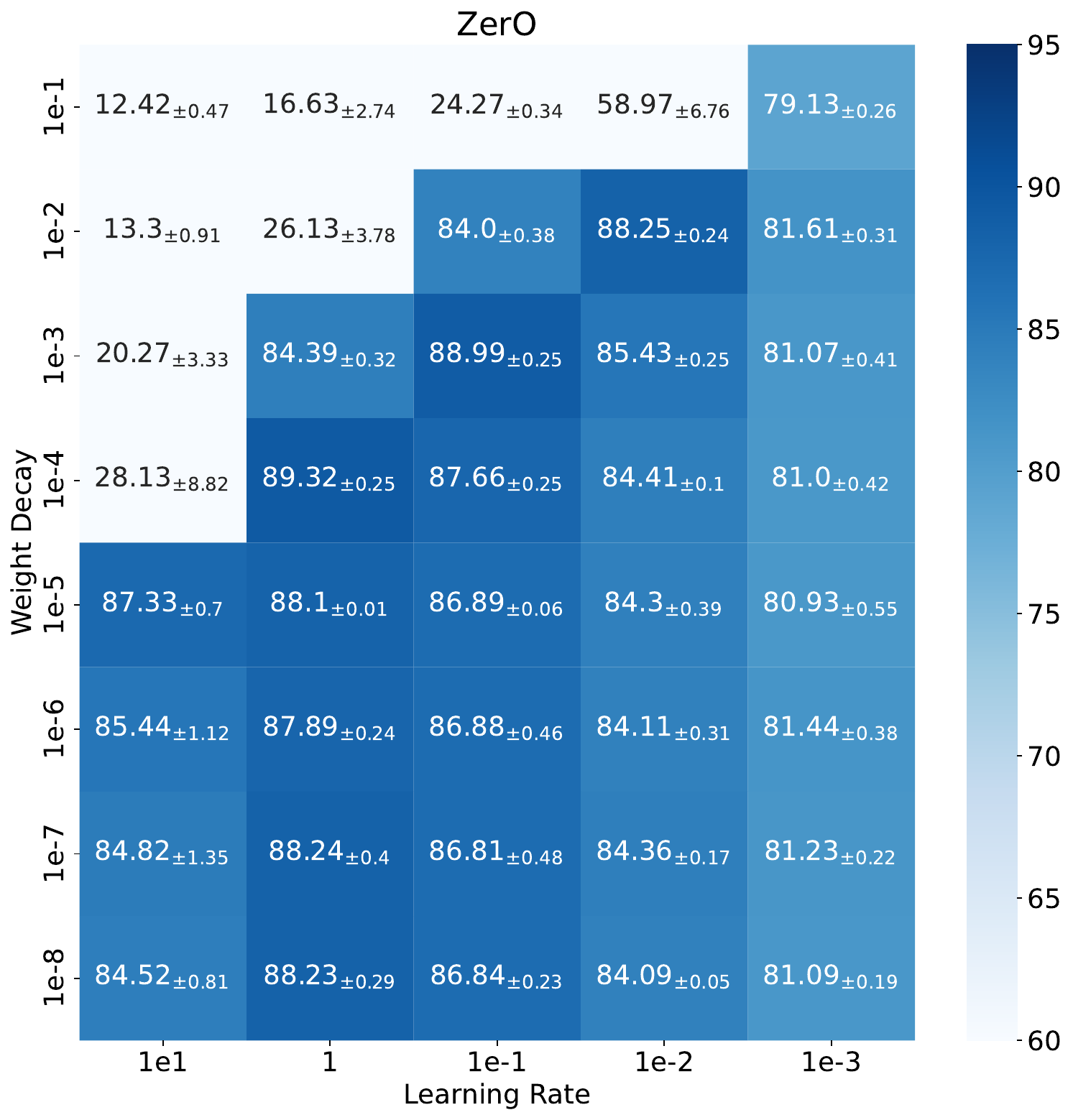}
	}
 \subfigure[Kaiming]{
		\includegraphics[width=0.3\textwidth]{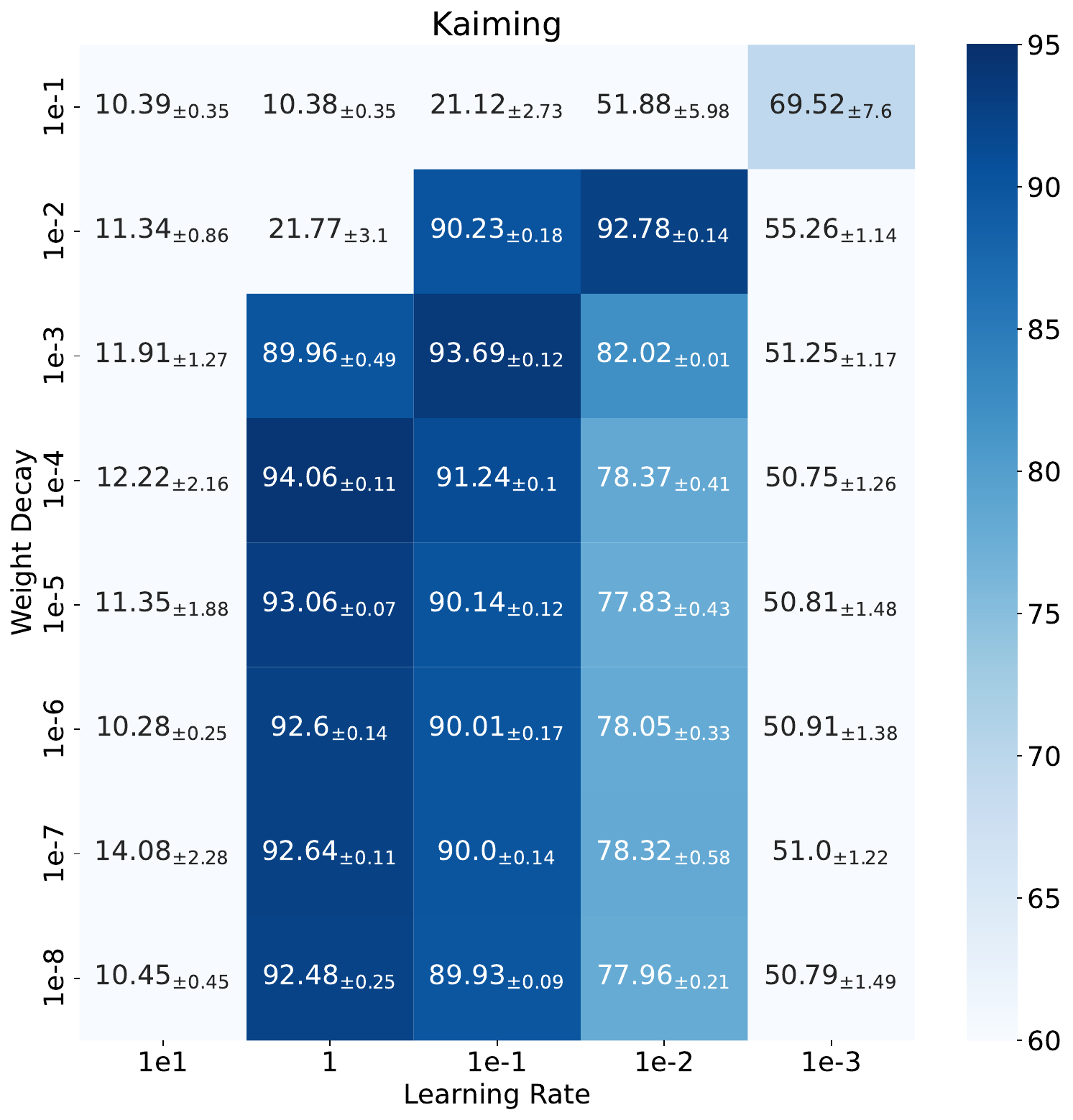}
	}
 
 \subfigure[Fixup]{
		\includegraphics[width=0.3\textwidth]{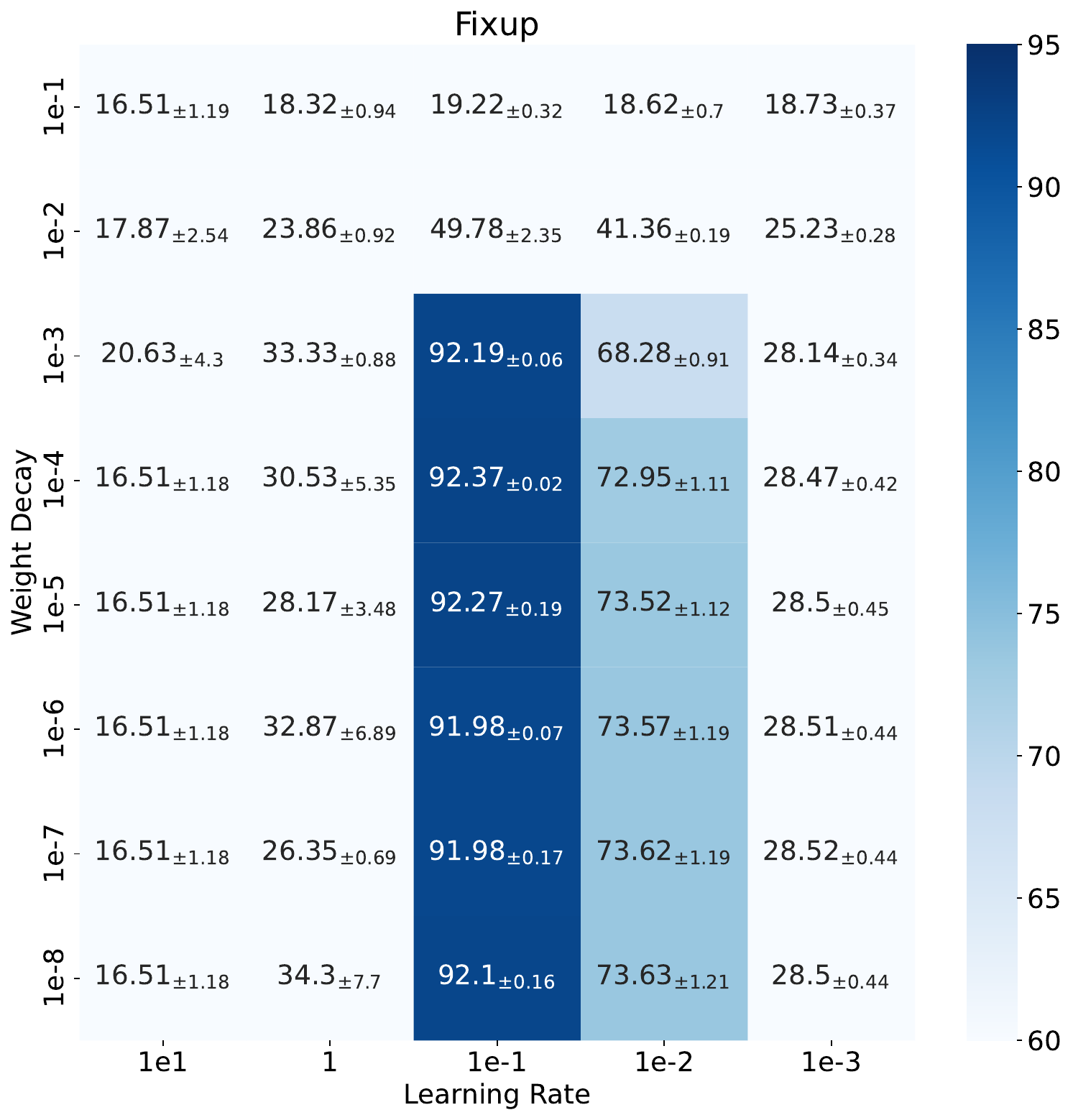}
	}
  \subfigure[Rezero]{
		\includegraphics[width=0.3\textwidth]{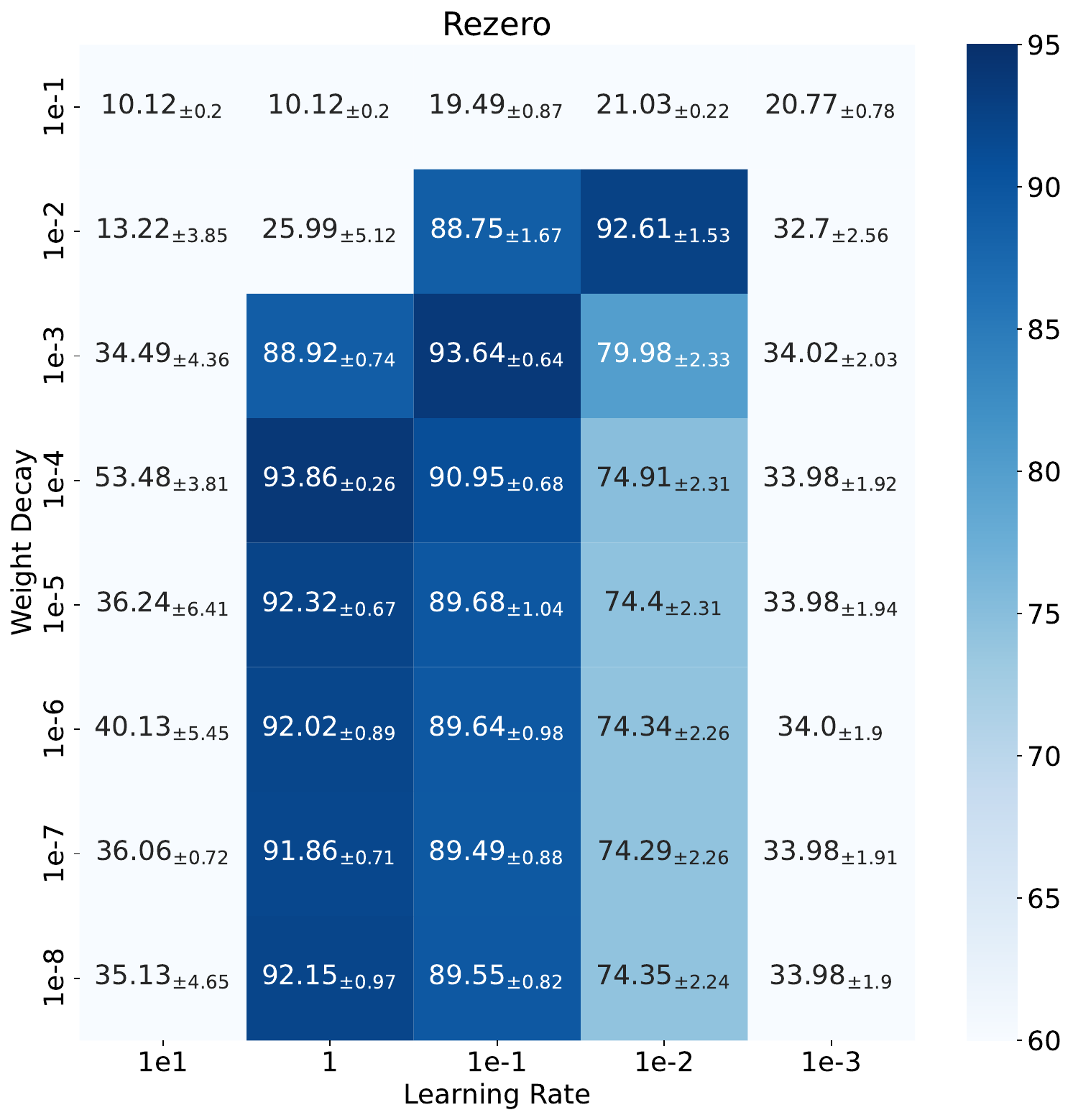}
	}
	\caption{The expanded hyperparameter experiment on Cifar10 on ResNet-32.}
	\label{fig:hypers-app-long}
\end{figure}

We scanned the learning rate from 1e-3 to 1e1 and weight decay from 1e-8 to 1e-1, ensuring that the best-performing hyperparameters are not at the corners or edges of the grid. As shown in Figure~\ref{fig:hypers-app-long}, IDInit achieves a peak accuracy of 94.08\% with a weight decay of 1e-3 and a learning rate of 1e-1. In comparison to other initialization methods including Kaiming, Fixup, and Rezero, IDInit demonstrates superior stability, maintaining high accuracy even when the learning rate is reduced below 1e-1. Although ZerO exhibits comparable stability at lower learning rates owing to its Hadamard matrix's ability to sustain dynamics, it underperforms at higher learning rates due to the dead neurons caused by the zero weights in its residual stems. Fixup, on the other hand, lacks stability by eliminating batch normalization, rendering it unsuitable for high learning rates. Overall, IDInit consistently delivers robust performance while maintaining stability, making it a promising candidate for practical applications.

\subsection{Details of Image Classification on Cifar10 Experiment}
\label{sec:appresidual}
In this experiment, we validate the proposed initialization with the comparison with existing initialization, including (1) Fixup; (2) SkipInit; (3) ReZero; (4) Kaiming; (5) Zero $\gamma$ (Setting the scale in Batch Normalization (BN) to 0). We use ResNet-56/110 as backbones on Cifar10.
For analyzing convergence, we adopt both SGD and Adam optimizer for updating models. We set SGD, with the momentum 0.9, the weight decay 5e-4, and the learning rate 0.2. For Adam, the learning rate is 0.001, $\beta_1$ is 0.9 and $\beta_2$ is 0.999. We train models for 200 epochs. The learning rate is reduced with a cosine function. The experiment is conducted on one Nvidia A100.

\begin{figure}[h]
	\centering
        \subfigure[Kaiming]{
		\includegraphics[width=0.25\textwidth]{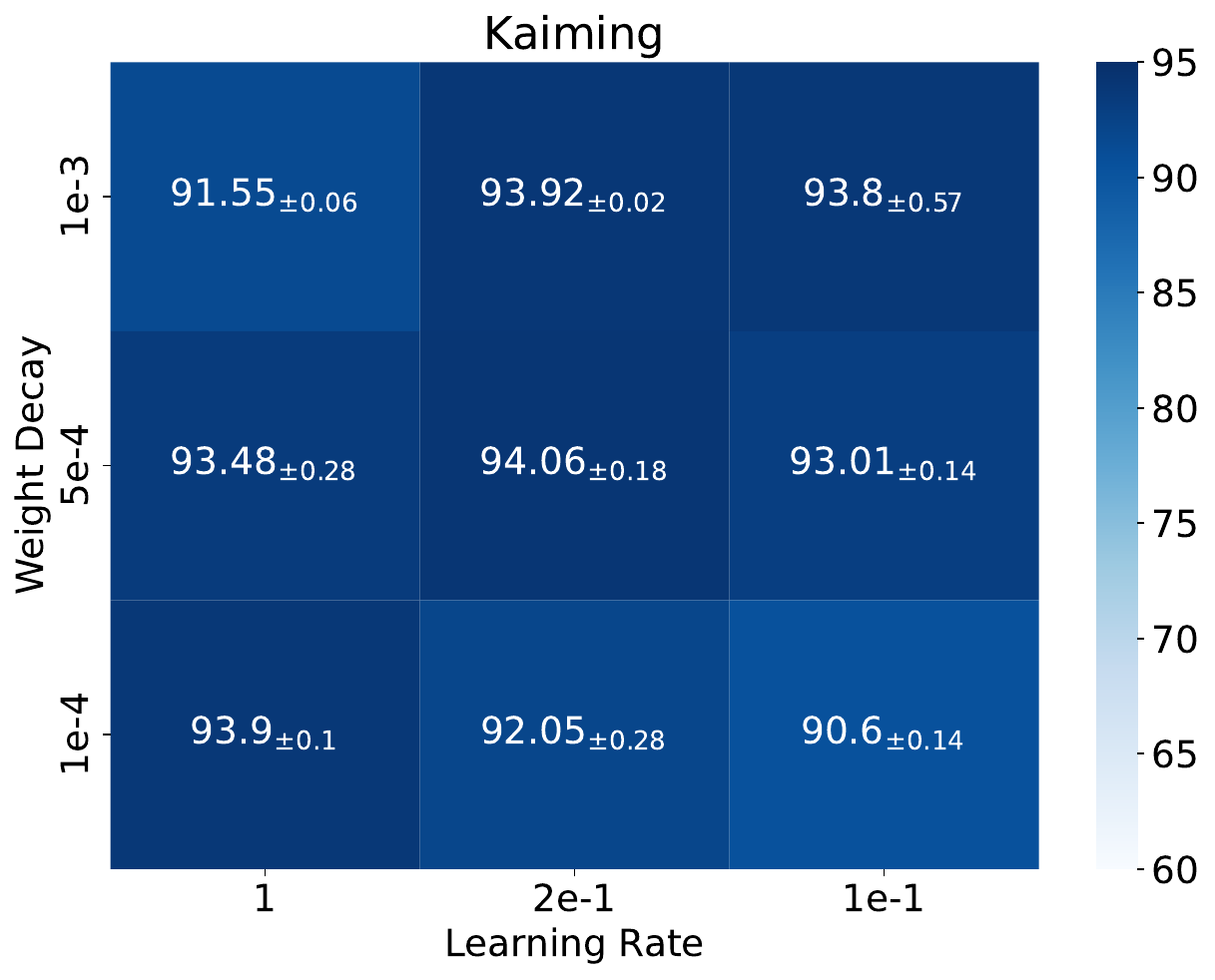}
	}
	\subfigure[Fixup]{
		\includegraphics[width=0.25\textwidth]{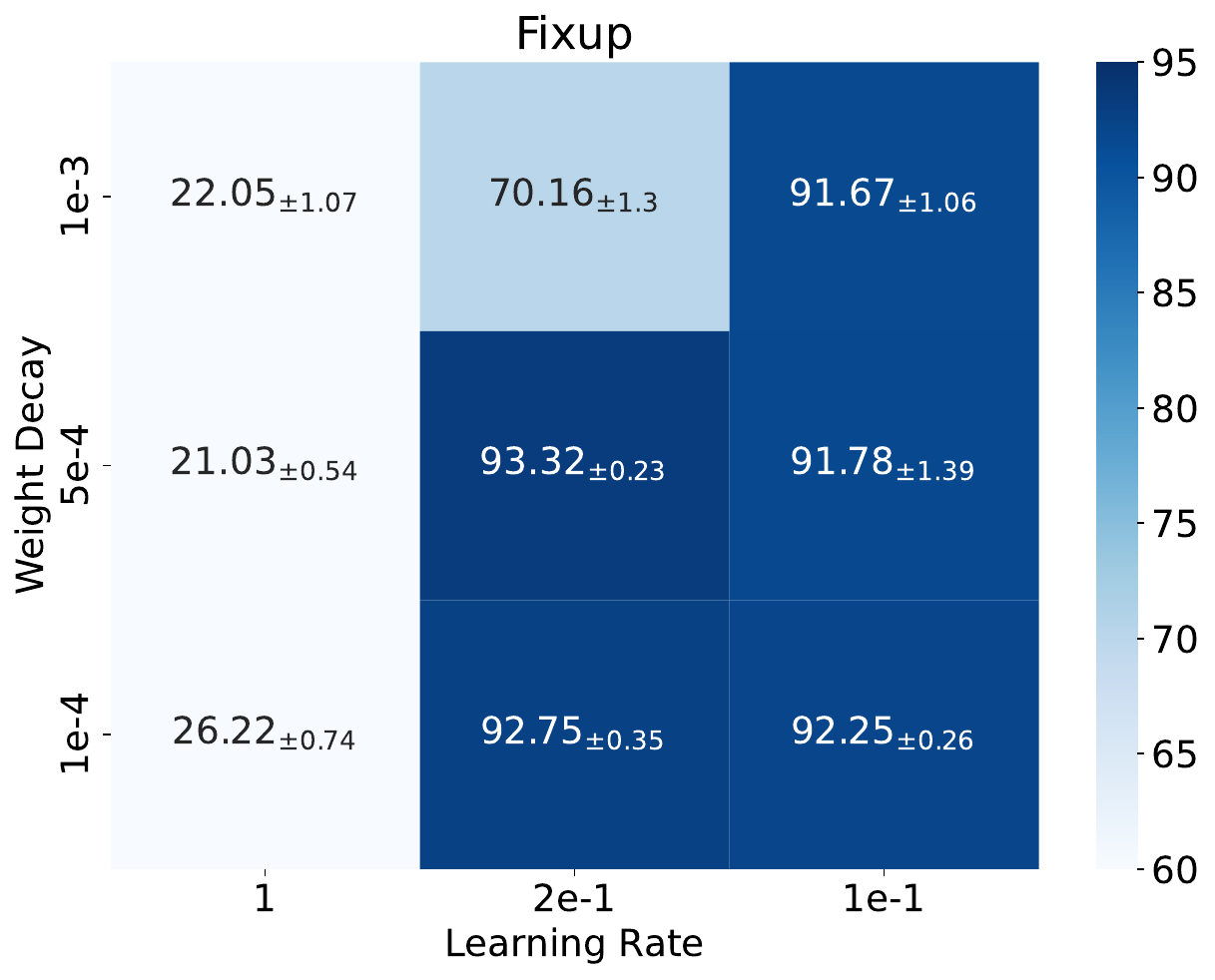}
	}
	\caption{The tuning hyperparameters on Cifar10 on ResNet-110.}
	\label{fig:hypers-app110-long}
\end{figure}

We perform a detailed hyperparameter analysis for ResNet-110, evaluating the learning rates \{1, 2e-1, 1e-1\} and weight decays \{1e-4, 5e-4, 1e-3\} on the standard baseline Kaiming and the more fragile Fixup method. As shown in Figure~\ref{fig:hypers-app110-long}, both Kaiming and Fixup achieve optimal accuracy with a learning rate of 2e-1 and a weight decay of 5e-4. However, Fixup fails to train with a learning rate of 1. Consequently, selecting a learning rate of 2e-1 and a weight decay of 5e-4 as the training hyperparameters in Section~\ref{sec:residual} is justified.

\subsection{Details of Ablation Experiment}
\label{sec:extend-ablation}
The dataset is Cifar10 and the backbone is ResNet-20. We choose SGD with momentum 0.9, weight decay 5e-4, and learning rate 0.1 to train the models for 200 epochs. The learning rate is reduced with a cosine function. And data-augment mixup is applied. The experiment is conducted on one Nvidia A100.

The extended analysis in Table~\ref{tbl:more-ablation} shows that the Loose condition, along with the components IDIC and IDIZ, contributes independently to performance improvements. Furthermore, the combination of these components yields the most significant results. Across all comparison pairs—specifically, settings 1/4, 2/6, 3/7, and 5/8—the Loose condition consistently demonstrates performance improvements. This highlights its practical value and its role in enhancing the overall effectiveness of the initialization methods.

\begin{table*}[h]
\caption{Analysis of components.}
\label{tbl:more-ablation}
\begin{center}
\scalebox{0.8}{
\begin{tabular}{@{}ccccccccc@{}}
\toprule
Component & Setting 1      & Setting 2      & Setting 3      & Setting 4      & Setting 5      & Setting 6     & Setting 7      & Setting 8      \\ \midrule
Loose     &                &                &                & $\checkmark$   &                & $\checkmark$  & $\checkmark$   & $\checkmark$   \\ \midrule
IDIC      &                &                & $\checkmark$   &                & $\checkmark$   &               & $\checkmark$   & $\checkmark$   \\ \midrule
IDIZ      &                & $\checkmark$   &                &                & $\checkmark$   & $\checkmark$  &                & $\checkmark$   \\ \midrule
Accuracy & $86.12_{\pm 0.52}$ & $92.68_{\pm 0.08}$ & $89.47_{\pm 0.24}$ & $87.01_{\pm 0.29}$ & $92.95_{\pm 0.21}$ & $92.9_{\pm 0.18}$ & $90.43_{\pm 0.14}$ & $\mathbf{93.22}_{\pm 0.05}$ \\ \bottomrule
\end{tabular}
}
\end{center}
\end{table*}

\subsection{Details of Text Classification Experiment}
We also explore performance networks on text classification datasets including SST2, SST5~\citep{DBLP:conf/emnlp/SocherPWCMNP13} and TREC-6, and we select TextCNN~\citep{DBLP:conf/emnlp/Kim14}, TextRNN~\citep{DBLP:conf/aaai/LaiXLZ15} and Transformer \citep{DBLP:conf/nips/VaswaniSPUJGKP17} for comparison. For TextCNN and TextRNN, we use AdaDelta~\citep{DBLP:journals/corr/abs-1212-5701} optimizer with a learning rate 1.0 and adopt Adam~\citep{DBLP:journals/corr/KingmaB14} for Transformer with a learning rate 1e-4. For the embedding layer, we utilize Glove~\citep{DBLP:conf/emnlp/PenningtonSM14} and Word2Vec~\citep{DBLP:journals/corr/abs-1301-3781} to initialize the embedding weights. All models are trained up to 10 epochs, and we run all the random initialization 5 times. The experiment is conducted on one Nvidia A100.

\begin{table*}[t]

\caption{Results of text classification on SST2 and TREC-6. The subscript G denotes the embedding layer is initialized by Glove, while W indicates Word2Vec. ``Default" means the default initialization of models, specifically, Kaiming for TextCNN, and Xavier for both TextRNN and Transformer. Fixup, ReZero and SkipInit are only applicable to the Transformer, as it is specifically designed for residual networks. Std values larger than 1.0 are marked in red.
}
\label{tbl:textcomplete}
\begin{adjustbox}{width=0.76\width,center}
\begin{tabular}{@{}llcccc@{}}
\toprule
Datasets                & Init.      & TextCNN\textsubscript{G/W}                            & TextRNN\textsubscript{G/W}                            & Transformer\textsubscript{G/W}  & Average\textsubscript{G/W}                        \\ \midrule
\multirow{5}{*}{SST2}   & Default    & 81.40$_{\pm0.66}$ / 84.56$_{\pm0.43}$                   & 81.69$_{\pm0.30}$ / 84.29$_{\pm0.70}$                   & 80.97$_{\pm\fatal{1.20}}$ / 83.36$_{\pm0.76}$ & 81.35$_{\pm0.72}$ / 84.07$_{\pm0.63}$          \\
                        & Orthogonal & 82.24$_{\pm0.44}$ / 84.37$_{\pm0.38}$                   & 81.86$_{\pm0.55}$ / 84.61$_{\pm0.78}$                   & 82.22$_{\pm0.87}$ / 83.99$_{\pm0.23}$  & 82.11$_{\pm0.62}$ / 84.32$_{\pm0.46}$                  \\
                        & Fixup & -                   & -                   & 78.72$_{\pm0.78}$ / 81.25$_{\pm0.27}$   & -                  \\
                        & ReZero & -                   & -                   & 81.67$_{\pm0.77}$ / 82.32$_{\pm0.51}$   & -                  \\
                        & SkipInit & -                   & -                   & 82.30$_{\pm0.47}$ / 84.12$_{\pm0.75}$   & -                  \\
                        & ZerO & 82.05$_{\pm0.67}$ / 84.26$_{\pm0.39}$                   & 82.03$_{\pm0.41}$ / 84.80$_{\pm0.64}$                   & 82.28$_{\pm0.81}$ / 82.72$_{\pm0.55}$   & 82.12$_{\pm0.63}$ / 83.93$_{\pm0.53}$                    \\
                        & IDInit     & \textbf{82.60}$_{\pm0.24}$ / \textbf{85.67}$_{\pm0.41}$ & \textbf{82.66}$_{\pm0.16}$ / \textbf{85.49}$_{\pm0.33}$ & \textbf{82.48}$_{\pm0.55}$ / \textbf{84.51}$_{\pm0.24}$   & \textbf{82.58}$_{\pm0.32}$ / \textbf{85.22}$_{\pm0.33}$  \\ \midrule
\multirow{5}{*}{TREC-6} & Default    & 90.80$_{\pm0.94}$ / 92.06$_{\pm1.00}$                   & 86.34$_{\pm\fatal{1.04}}$ / 90.52$_{\pm\fatal{1.54}}$   & 86.68$_{\pm\fatal{2.68}}$ / 89.20$_{\pm\fatal{1.20}}$   & 87.94$_{\pm\fatal{1.55}}$ / 90.59$_{\pm\fatal{1.25}}$    \\
                        & Orthogonal & 90.34$_{\pm0.72}$ / 92.72$_{\pm0.84}$                   & 85.86$_{\pm0.90}$ / 89.88$_{\pm\fatal{1.54}}$           & 86.90$_{\pm\fatal{1.51}}$ / 89.26$_{\pm0.86}$  & 87.70$_{\pm0.71}$ / 90.62$_{\pm0.75}$           \\
                        & Fixup & -                   & -                   & 86.95$_{\pm0.35}$ / 89.35$_{\pm0.53}$   & -                 \\
                        & ReZero & -                   & -                   & 86.92$_{\pm0.98}$ / 89.36$_{\pm0.52}$   & -                 \\
                        & SkipInit & -                   & -                   & 83.59$_{\pm0.61}$ / 87.10$_{\pm0.41}$   & -                 \\
                        & ZerO & 90.89$_{\pm0.41}$ / 92.90$_{\pm0.50}$                   & \textbf{87.24}$_{\pm0.64}$ / 88.71$_{\pm0.40}$                   & 86.97$_{\pm0.75}$ / 89.38$_{\pm0.64}$  & 88.37$_{\pm0.60}$ / 90.33$_{\pm0.51}$                    \\
                        & IDInit     & \textbf{91.22}$_{\pm0.54}$ / \textbf{92.94}$_{\pm0.48}$ & 87.04$_{\pm0.26}$ / \textbf{90.60}$_{\pm0.58}$ & \textbf{87.32}$_{\pm0.78}$ / \textbf{90.06}$_{\pm0.60}$  & \textbf{88.53}$_{\pm0.53}$ / \textbf{91.20}$_{\pm0.55}$ \\ \bottomrule
\end{tabular}
\end{adjustbox}

\end{table*}

\begin{figure*}[h]
	\centering
	\subfigure[ViT-B/32]{
		\includegraphics[width=0.17\textwidth]{figure/vit_val_acc.pdf}
	}
	\subfigure[RN-50 (Adamw)]{
		\includegraphics[width=0.17\textwidth]{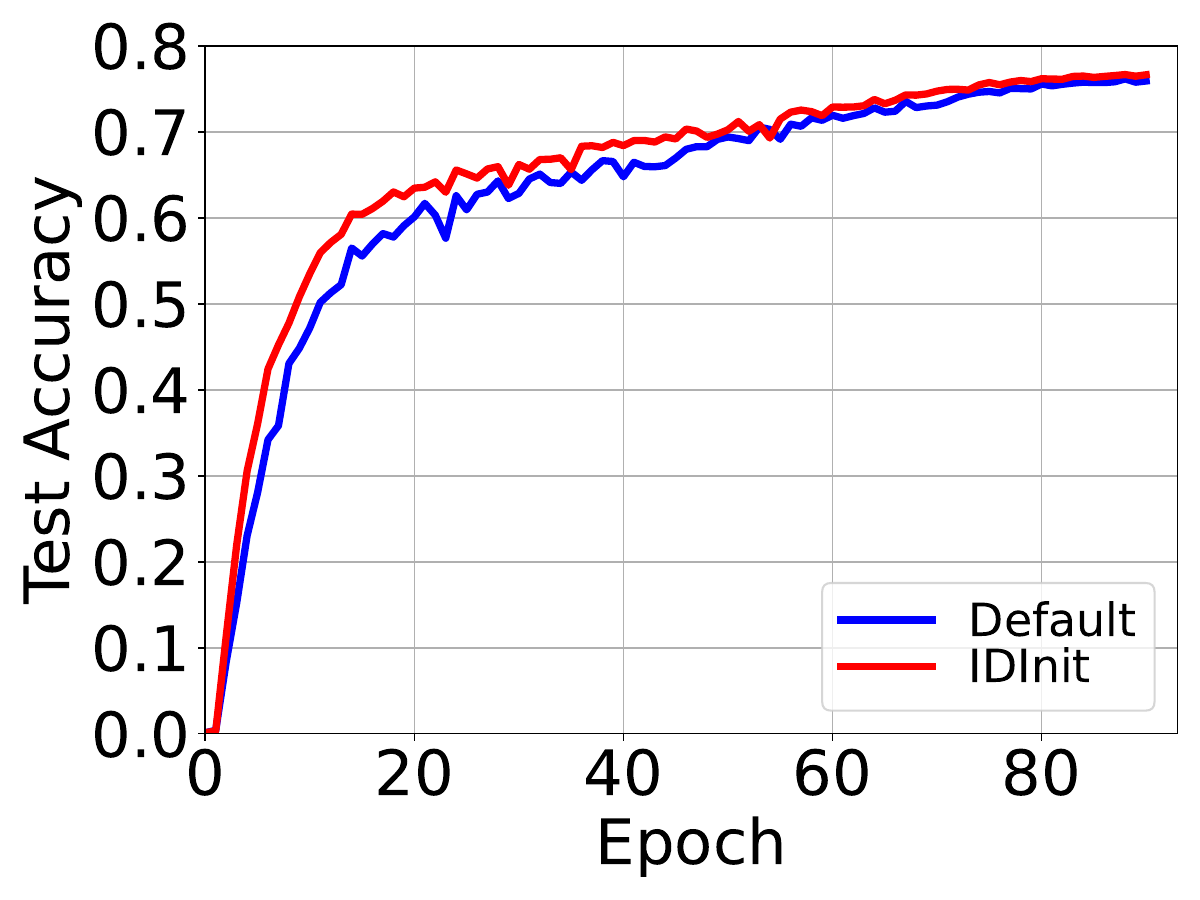}
	}
	\subfigure[RN-50]{
		\includegraphics[width=0.17\textwidth]{figure/resnet50_val_acc.pdf}
	}
	\subfigure[SRN-50]{
		\includegraphics[width=0.17\textwidth]{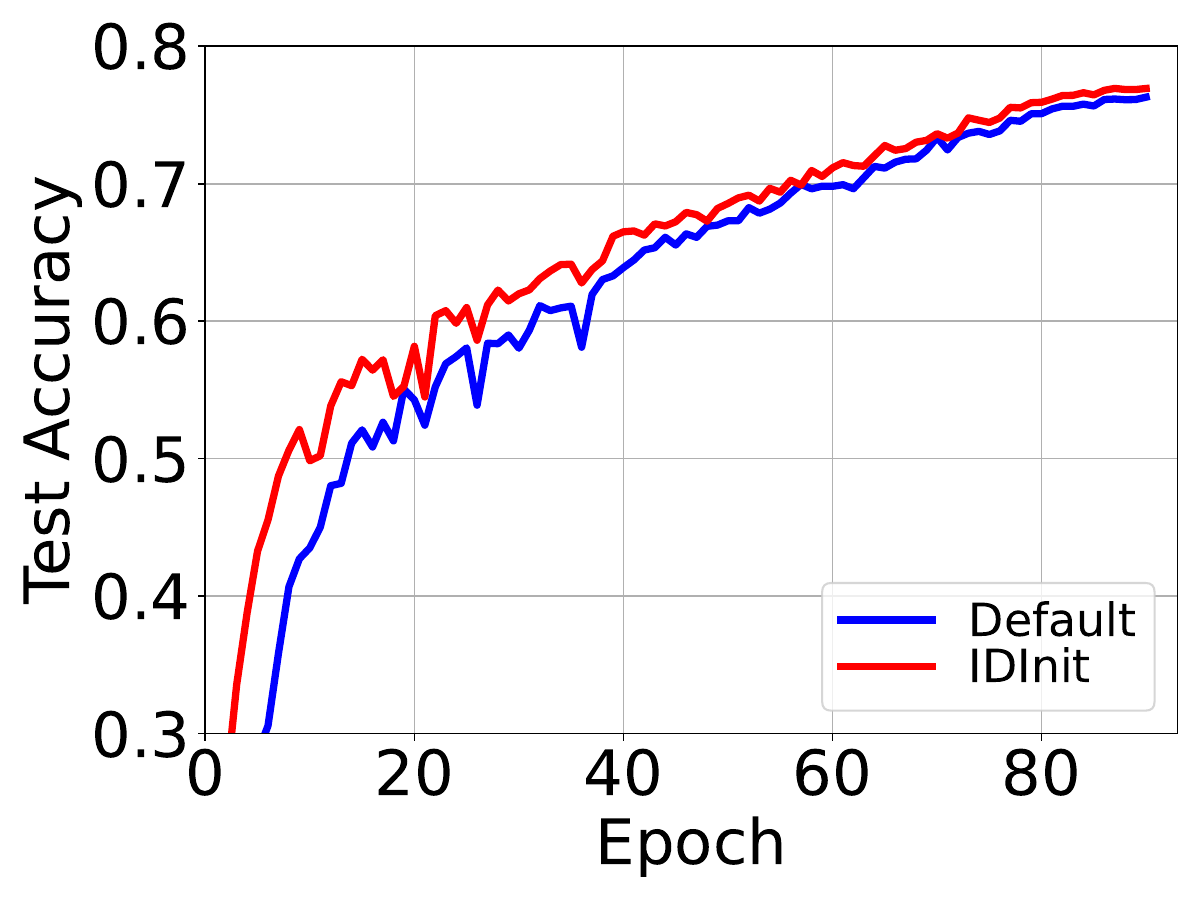}
	}
	\subfigure[RN-152]{
		\includegraphics[width=0.17\textwidth]{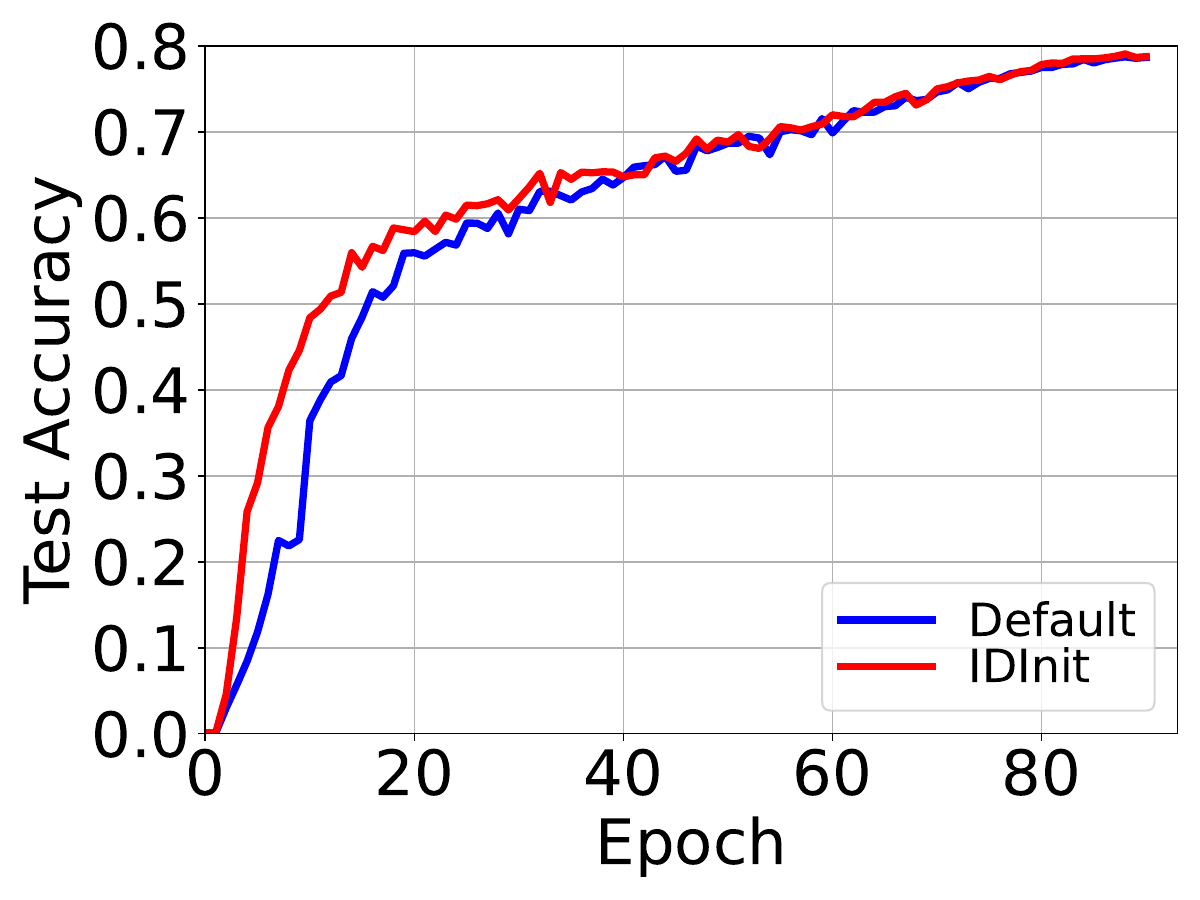}
	}
	\caption{Results on ImageNet. ``Default" means the default initialization of models. RN-50 (Adamw) means that ResNet-50 is trained with the same optimizer Adamw as the ViT-B/32.}
	\label{fig:vit-all-imagenet}
\end{figure*}

\subsection{Details of Image Classification on ImageNet Experiment}
\label{sec:app-imagenet}
In this experiment, we use ImageNet for validation. We use ViT-B/32~\citep{DBLP:conf/iclr/DosovitskiyB0WZ21}, ResNet-50/152 (RN-50/152) and Se-ResNet-50 (SRN-50) as backbones. For ViT-B/32 that inputs $32 \times 32$ patch window, the optimizer is AdamW with a learning rate 1e-3 and a weight decay of 5e-2. And the batch size is 1024. The epoch for training is 300. We use 30 epochs for warm-up. The input image size is $224 \times 224$. The dropout rates of the embedding layer and the network layer are all 0.1. For RN-50/152 and SRN-50, the optimizer is SGD with a learning rate 1e-1 and a weight decay of 1e-4. And the batch size is 1024. The epoch for training is 90. We use 9 epochs for warm-up. The input image size is $160 \times 160$ for the front 35 epochs and $224 \times 224$ for the remaining epochs. For all models, we apply data-augment including cutmix~\citep{DBLP:conf/iccv/YunHCOYC19} with $\alpha=1.0$, mixup~\citep{DBLP:conf/iclr/ZhangCDL18} with $\alpha=0.8$, the switching probability is 0.5 and a label smoothing with 0.1. The experiment is conducted on 4 Nvidia A100.

To further compare with other identity-control methods, we conducted experiments on ResNet-50 using Fixup and Zero. As shown in Table~\ref{tbl:imagenet-inits}, IDInit outperforms both Fixup and ZerO, demonstrating its superior performance on large-scale datasets.

\begin{table*}[h]
\caption{Comparison among Default, Fixup, ZerO and IDInit initialized ResNet-50 on ImageNet.}
\label{tbl:imagenet-inits}
\begin{center}
\begin{tabular}{@{}lcc@{}}
\toprule
Init.   & Epochs to 60\% Accuracy & Accuracy                                             \\ \midrule
Default & 38                 & 75.70 \\
Fixup   & 24                 & 75.83                                                \\
ZerO    & 30                 & 75.64                                                \\
IDInit  & 24                 & 76.72                                                \\ \bottomrule
\end{tabular}
\end{center}
\end{table*}

\subsection{Details of Pre-Training on Language Model}

Pre-training plays an important role in various applications. We conduct the experiment to show the fast convergence on BERT~\citep{DBLP:conf/naacl/DevlinCLT19}. The dataset is the concatenation of English Wikipedia and Toronto Book Corpus~\cite{DBLP:conf/iccv/ZhuKZSUTF15}.  We train the BERT-Base for 40 epochs with 768 batch size. The optimizer is set to AdamW with learning rate 1e-4 and weight decay 1e-2. 32 NVIDIA V100s are used.

\begin{table}[h]
\setlength{\tabcolsep}{8pt}
\renewcommand{\arraystretch}{1.6}

\caption{Results of Linear-5 on MNIST. ``Default" means the default initialization of  models where Xavier is for Linear-5-tanh and Kaiming is adopted for Linear-5-ReLU.}
\label{tbl:liner5}
\begin{center}
\begin{tabular}{ccc}
\toprule
Init.  & Linear-5-tanh  & Linear-5-ReLU      \\ \hline
Default & 98.26 & 98.21\\ \hline
IDInit & \textbf{98.32} & \textbf{98.4} \\
\hline
\end{tabular}
\end{center}
\end{table}

\section{Additional Experiments}
We provide additional experiments to further validate IDInit. $\tau$ and $\varepsilon$ are set the same as Sec.~\ref{sec:appdetailexp}.

\subsection{Validation on the Linear Structure}
This experiment is conducted on MNIST. We use five linear layers named Liner-5 whose hidden layers are all of dimension 512. The optimizer is SGD  with momentum 0.9, weight decay 5e-4, and a learning rate 1e-1. The learning rate scheduler adopts a cosine reduction strategy. We run the model in 30 epochs on one Nvidia A100. We both consider Linear-5-tanh and Linear-5-ReLU which consist of Linear-5, and tanh and ReLU activation functions, respectively. The experiment is conducted on one Nvidia A100.

As shown in Table~\ref{tbl:liner5}, IDInit can achieve the highest accuracy in both different tanh and ReLU conditions. The results show the ability of our proposed method to train a model with only fully-connected layers.

\begin{figure}[t]
	\centering
	\subfigure[FC-0.00]{
		\includegraphics[width=0.23\textwidth]{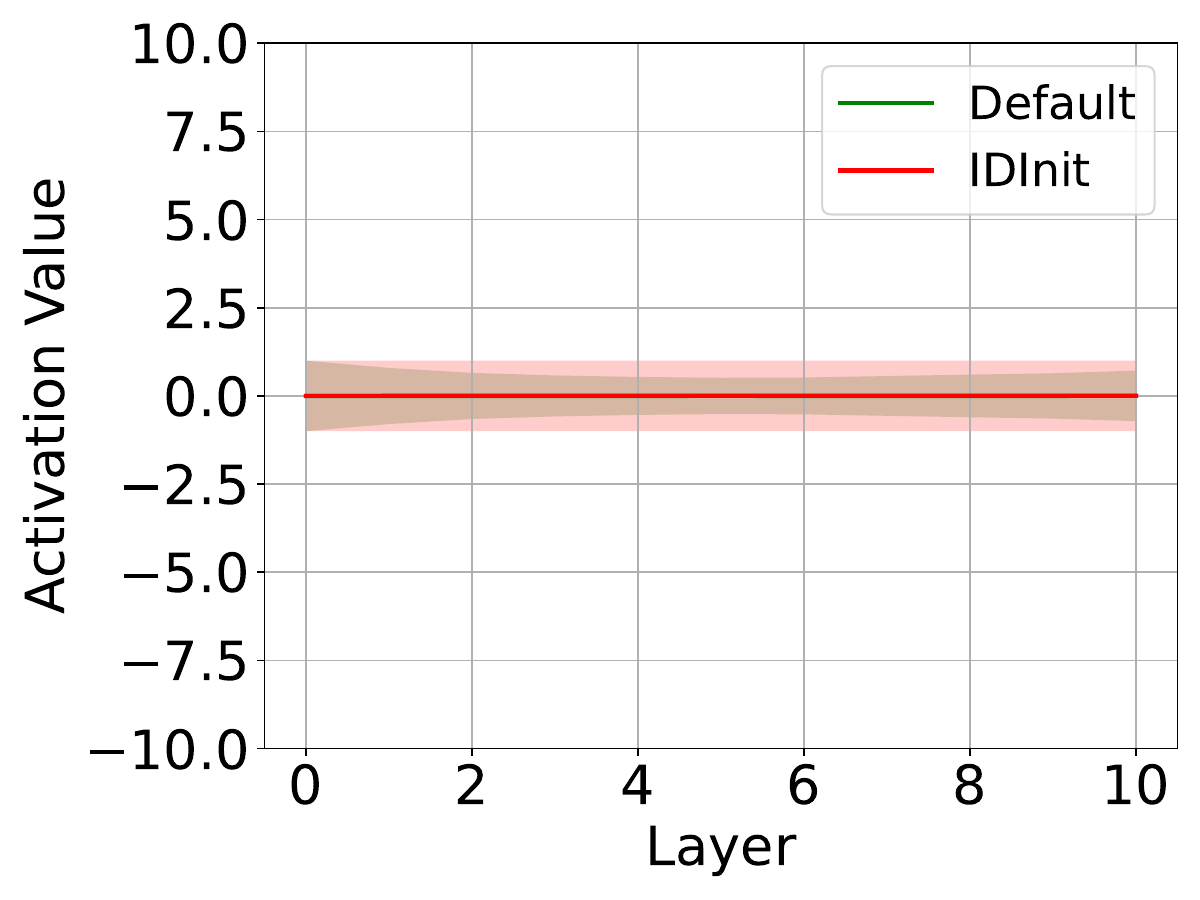}
	}
	\subfigure[ResFC-0.00]{
		\includegraphics[width=0.23\textwidth]{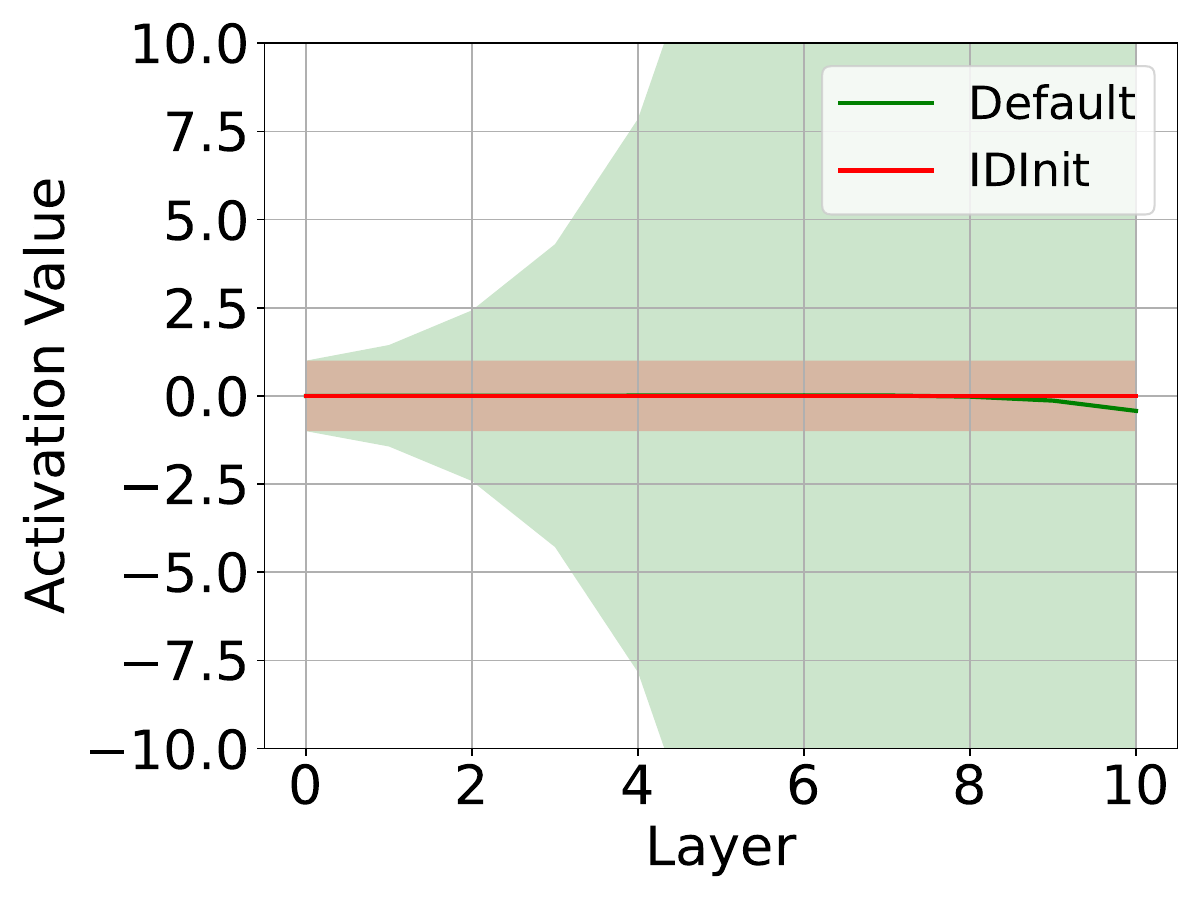}
	}
	\subfigure[Conv-0.00]{
		\includegraphics[width=0.23\textwidth]{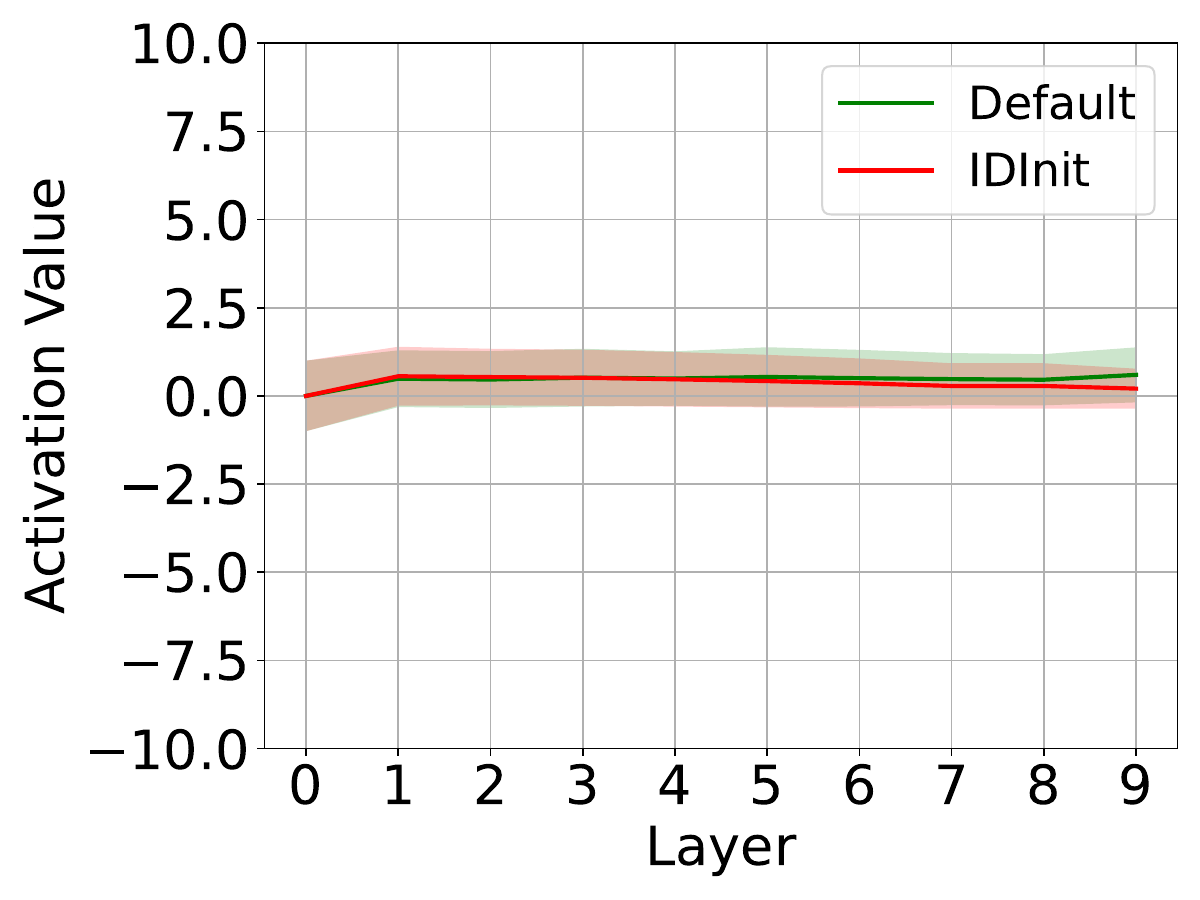}
	}
	\subfigure[ResConv-0.00]{
		\includegraphics[width=0.23\textwidth]{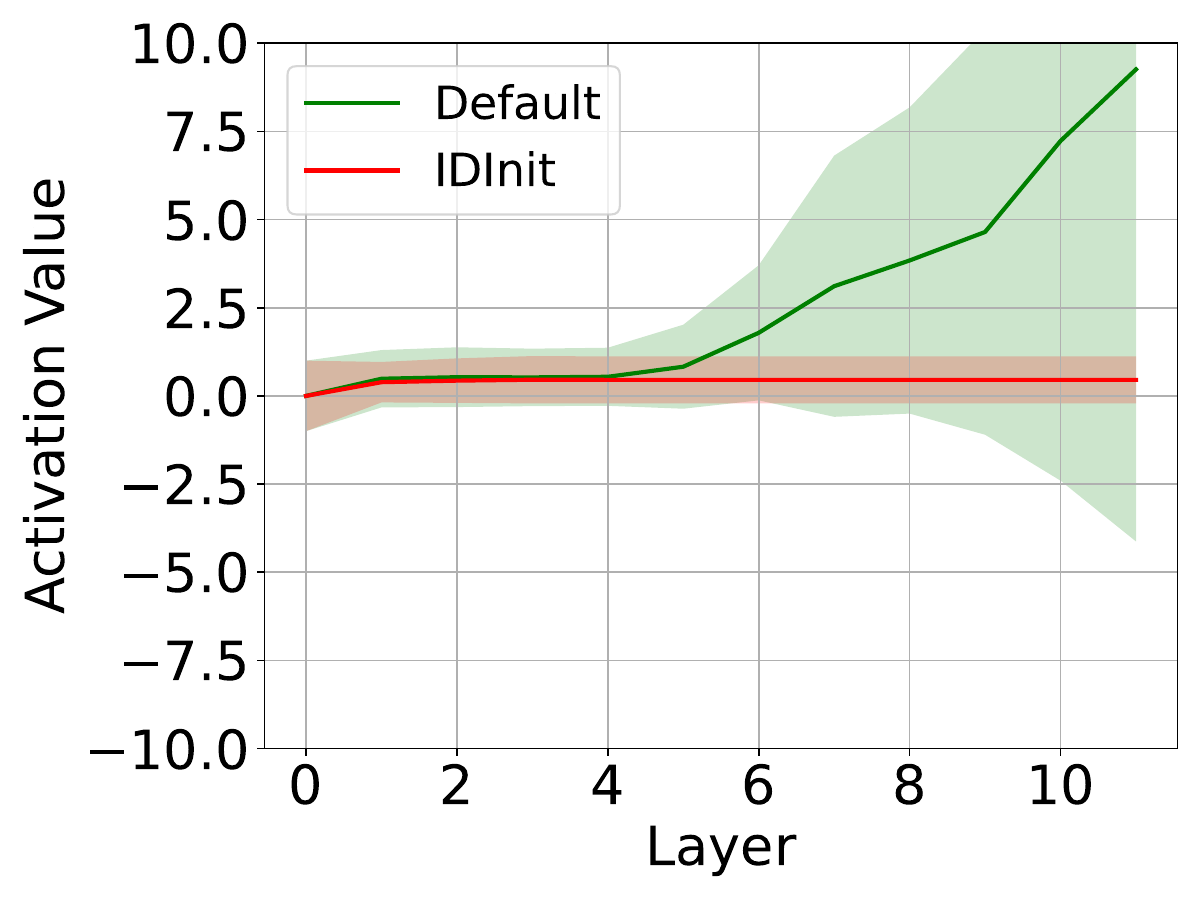}
	}
	
	\subfigure[FC-0.01]{
		\includegraphics[width=0.23\textwidth]{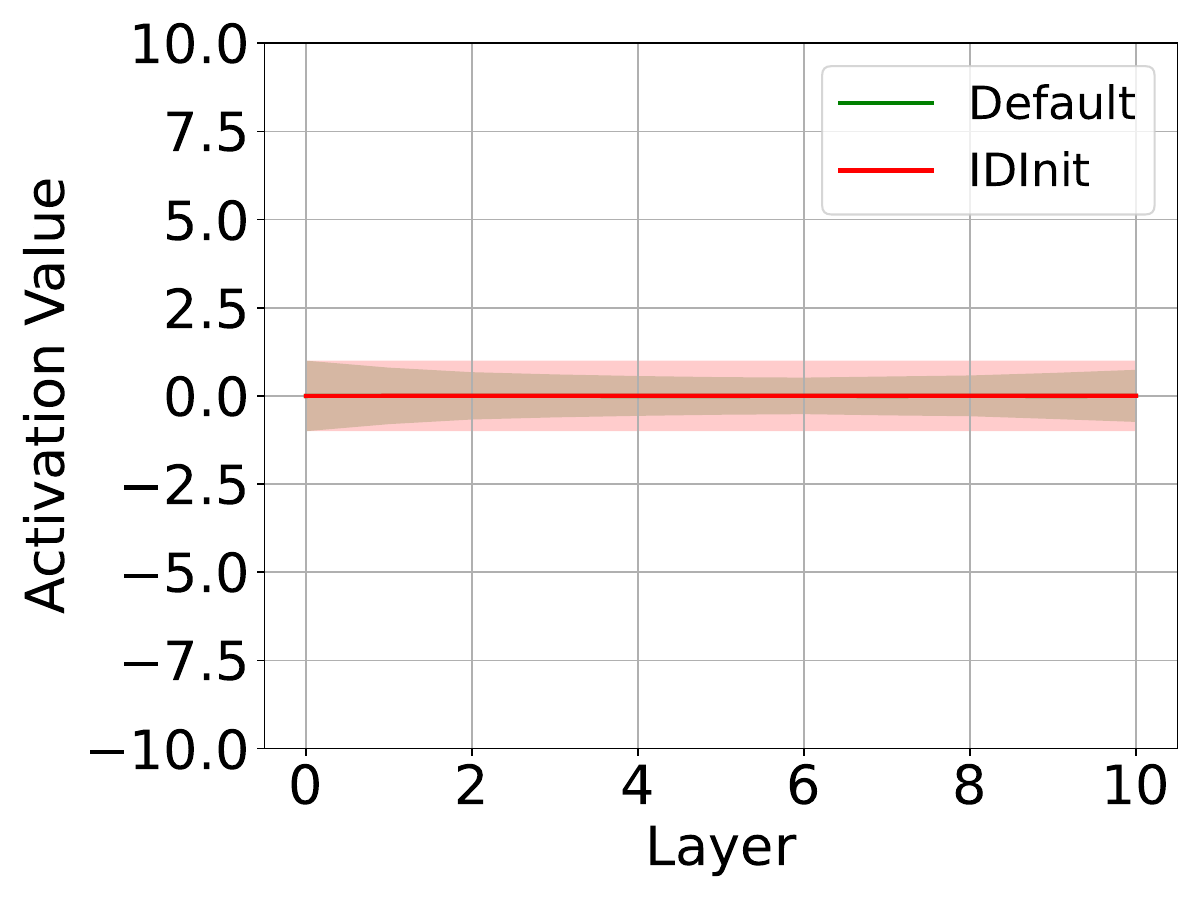}
	}
	\subfigure[ResFC-0.01]{
		\includegraphics[width=0.23\textwidth]{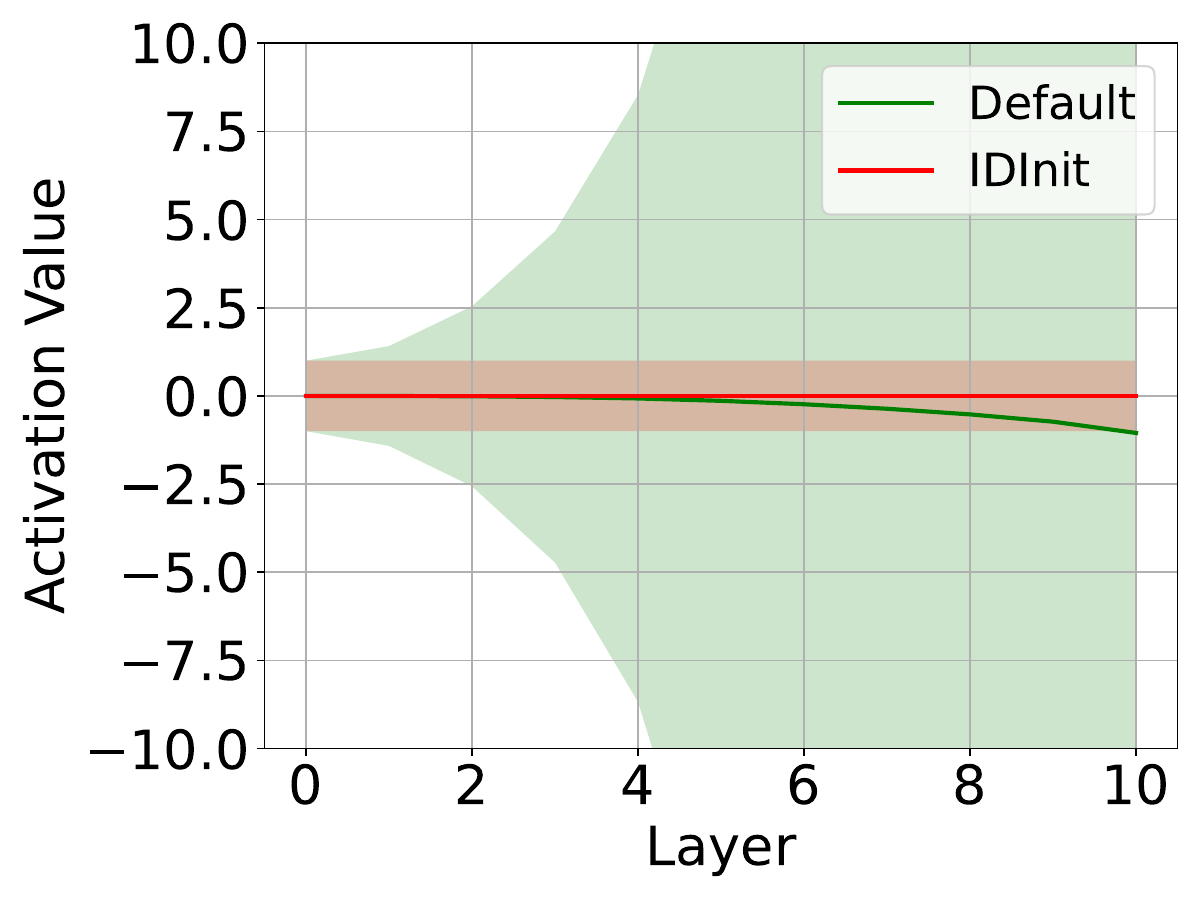}
	}
	\subfigure[Conv-0.01]{
		\includegraphics[width=0.23\textwidth]{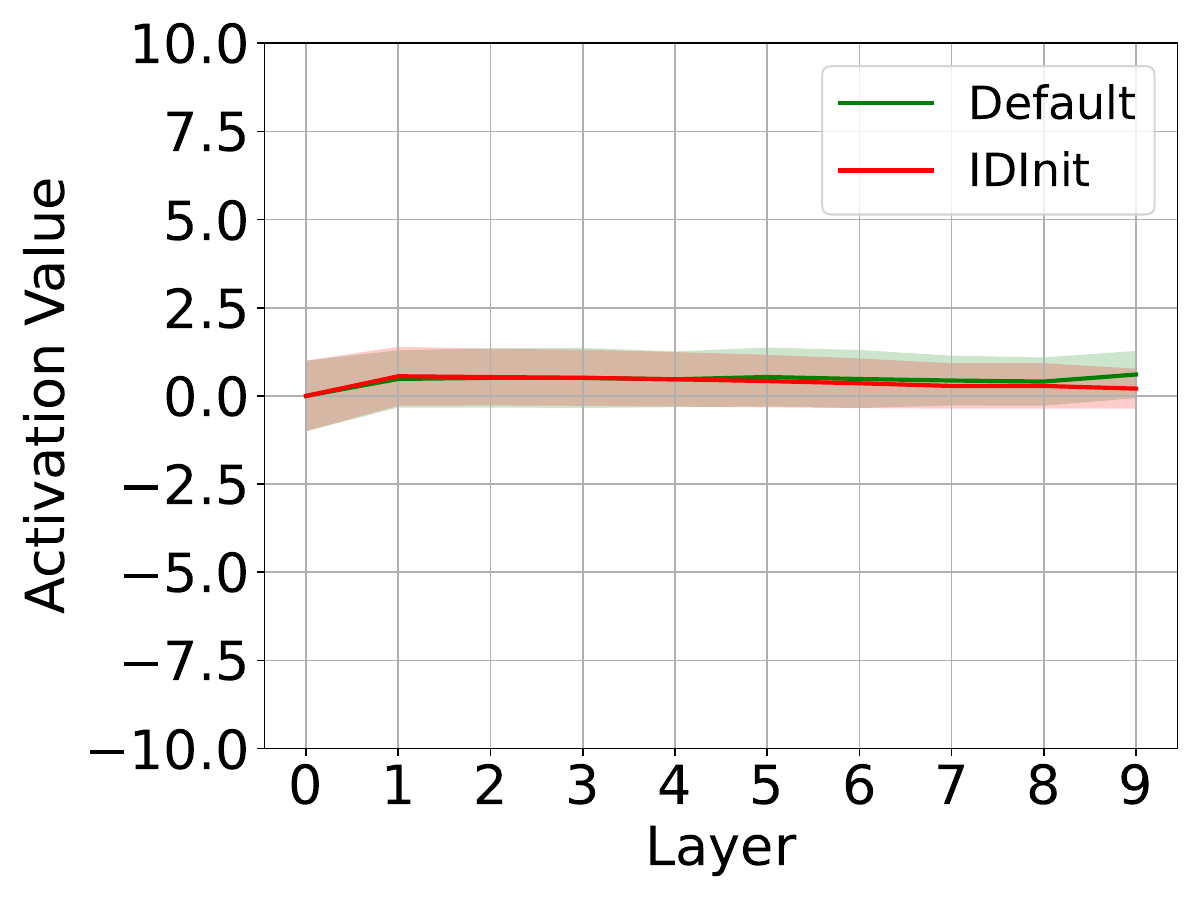}
	}
	\subfigure[ResConv-0.01]{
		\includegraphics[width=0.23\textwidth]{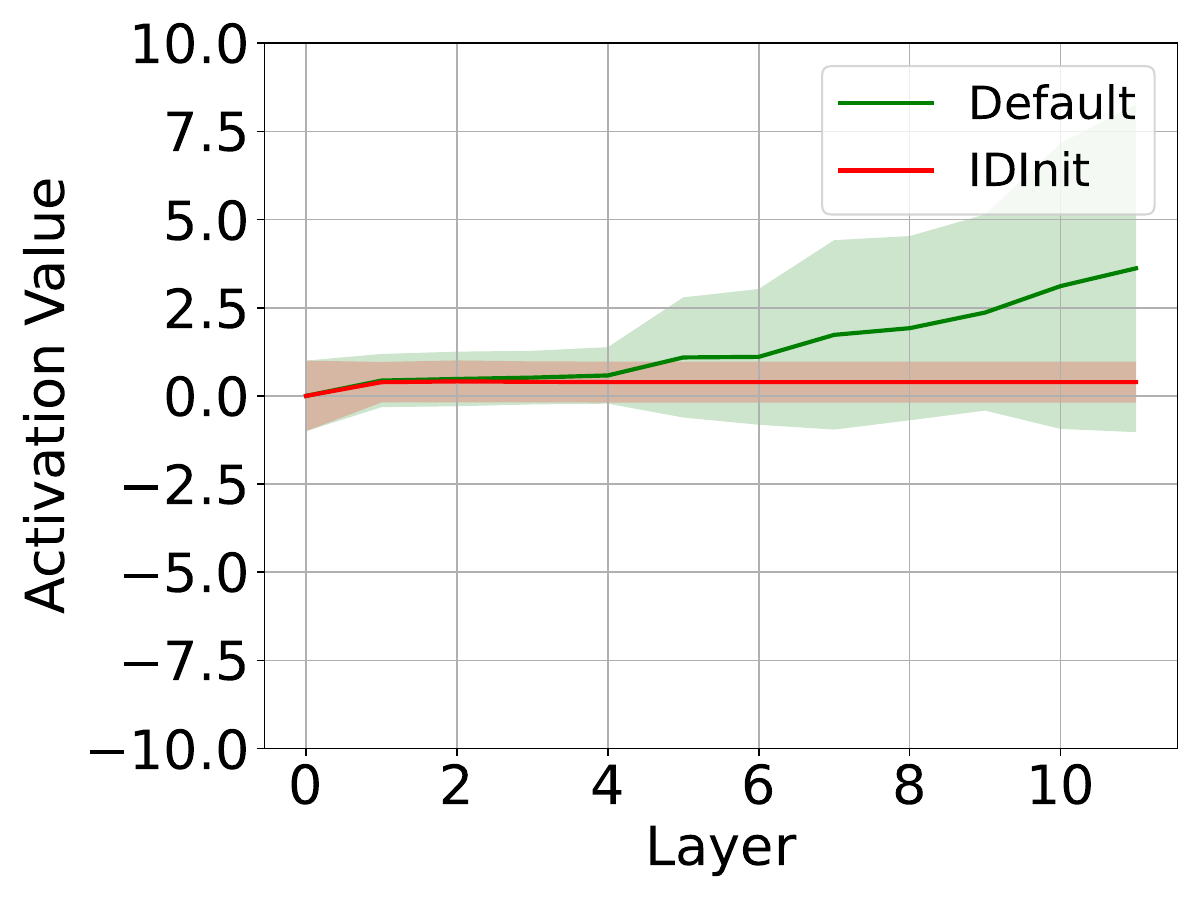}
	}
	
	\subfigure[FC-0.10]{
		\includegraphics[width=0.23\textwidth]{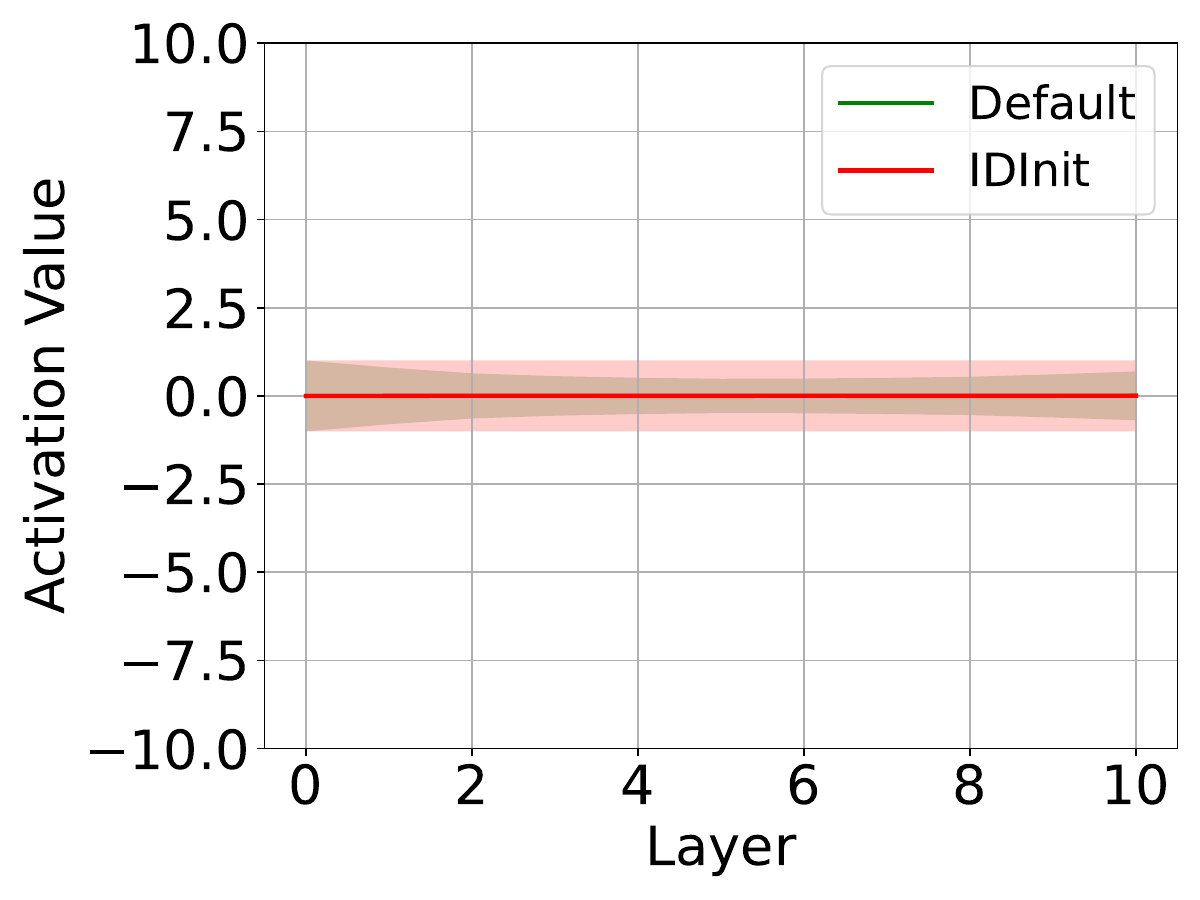}
	}
	\subfigure[ResFC-0.10]{
		\includegraphics[width=0.23\textwidth]{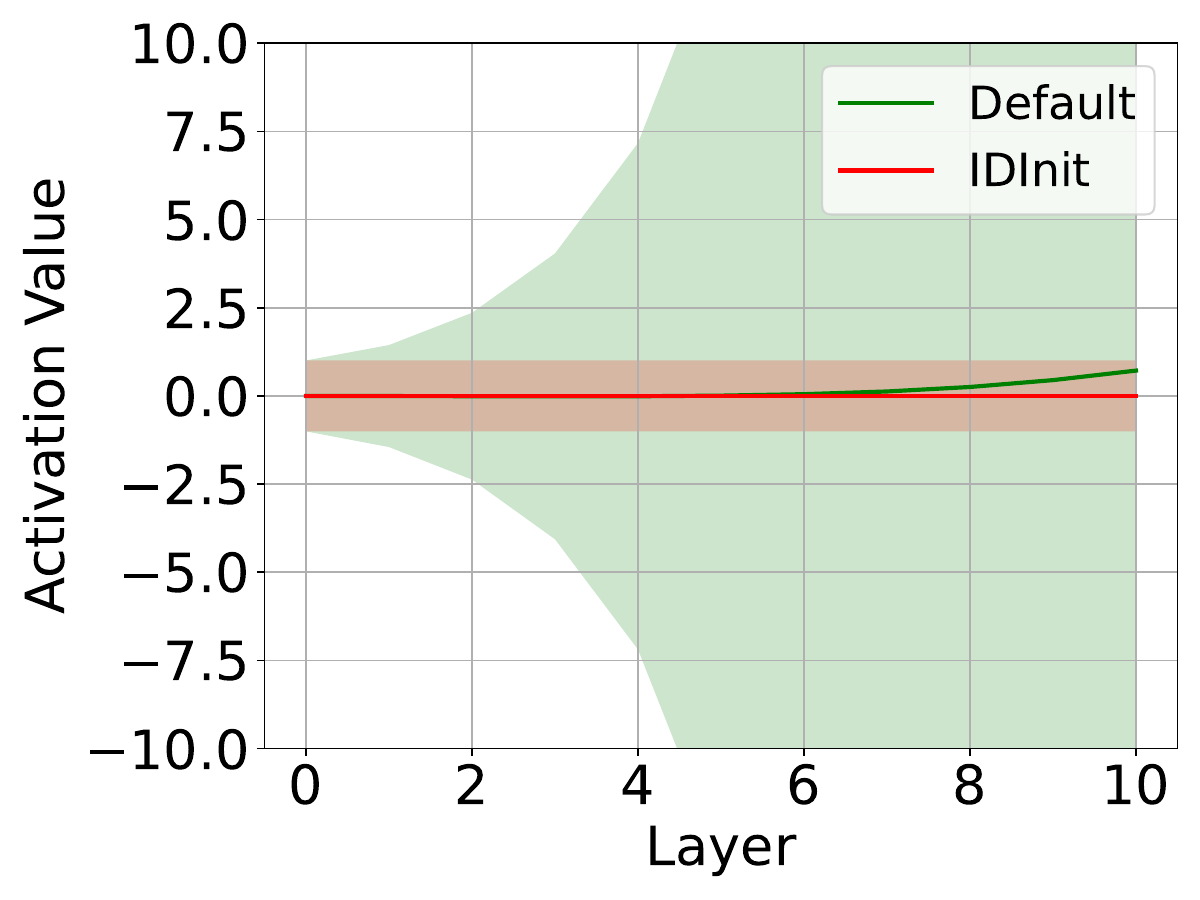}
	}
	\subfigure[Conv-0.10]{
		\includegraphics[width=0.23\textwidth]{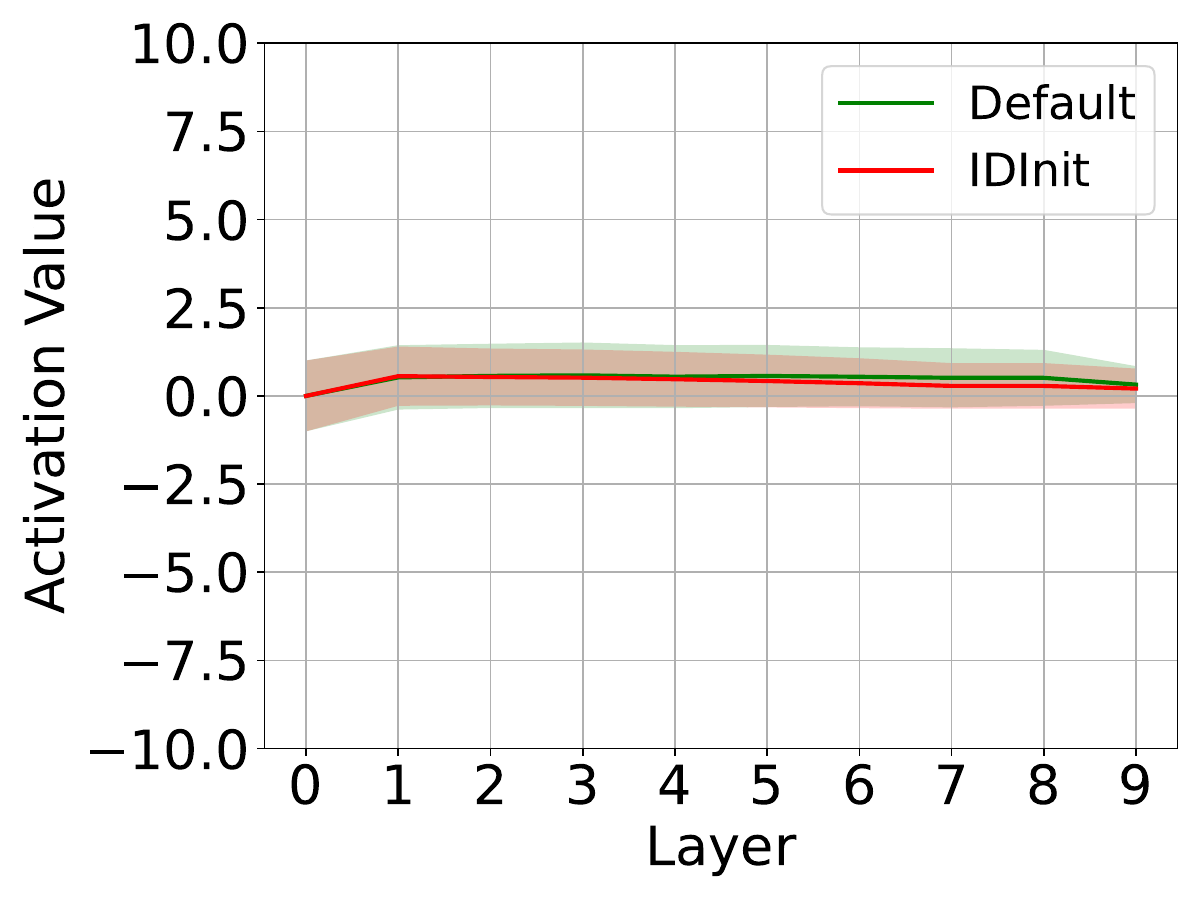}
	}
	\subfigure[ResConv-0.10]{
		\includegraphics[width=0.23\textwidth]{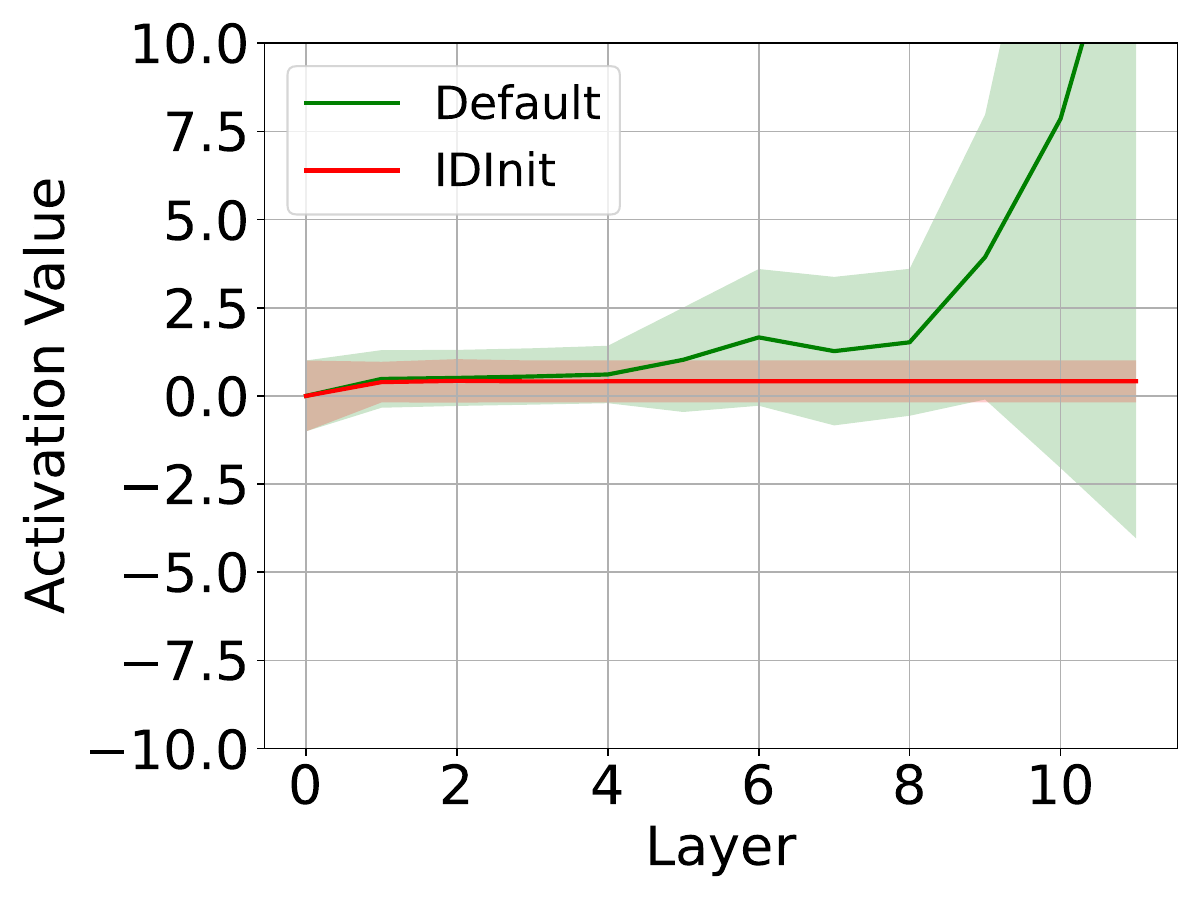}
	}
	
	\subfigure[FC-1.00]{
		\includegraphics[width=0.23\textwidth]{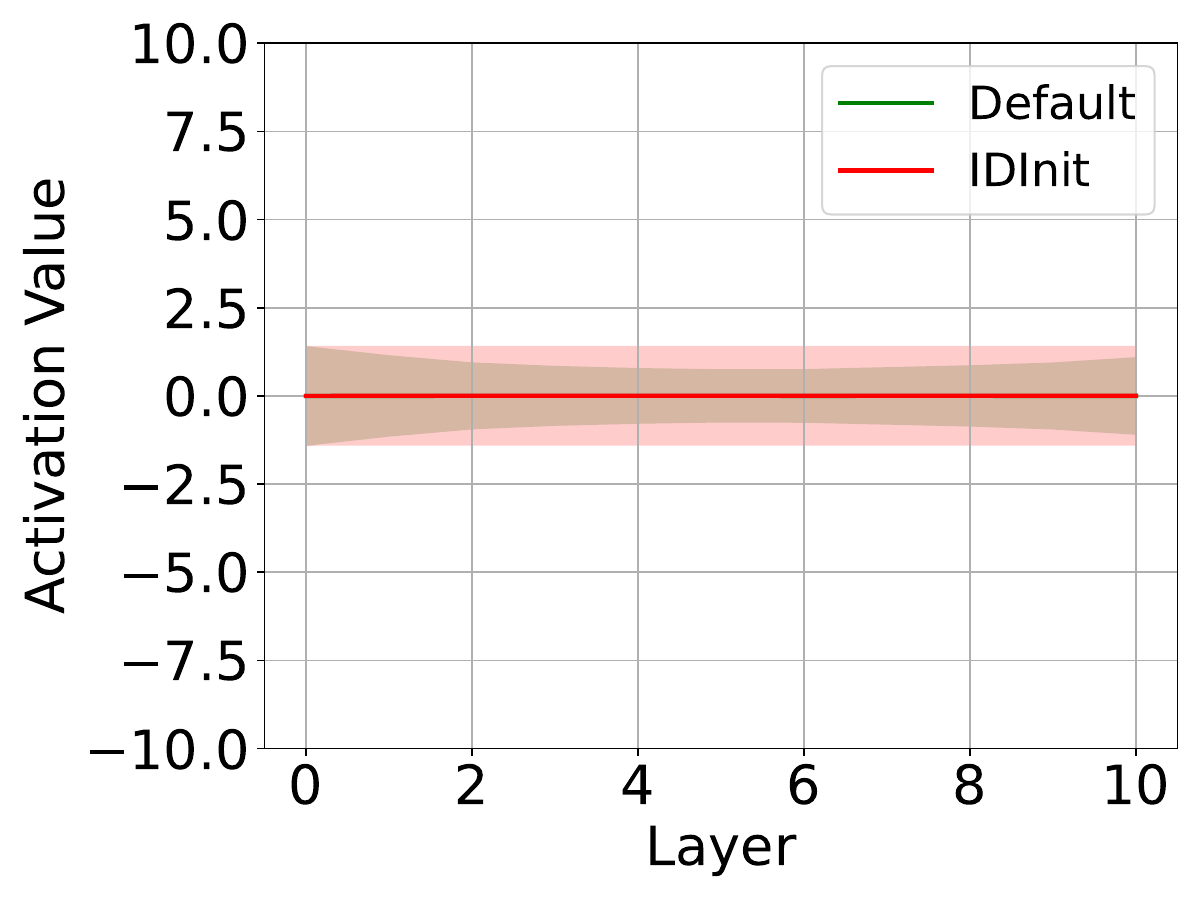}
	}
	\subfigure[ResFC-1.00]{
		\includegraphics[width=0.23\textwidth]{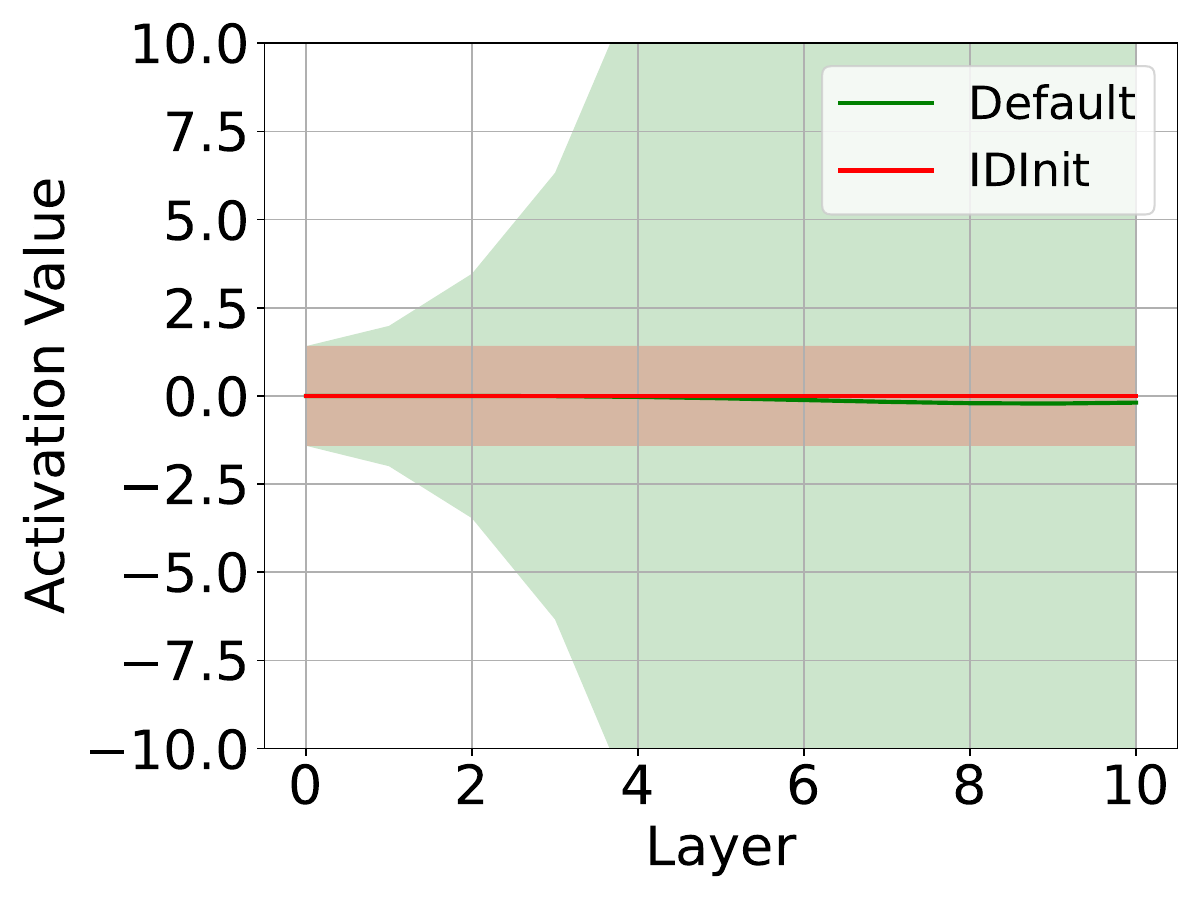}
	}
	\subfigure[Conv-1.00]{
		\includegraphics[width=0.23\textwidth]{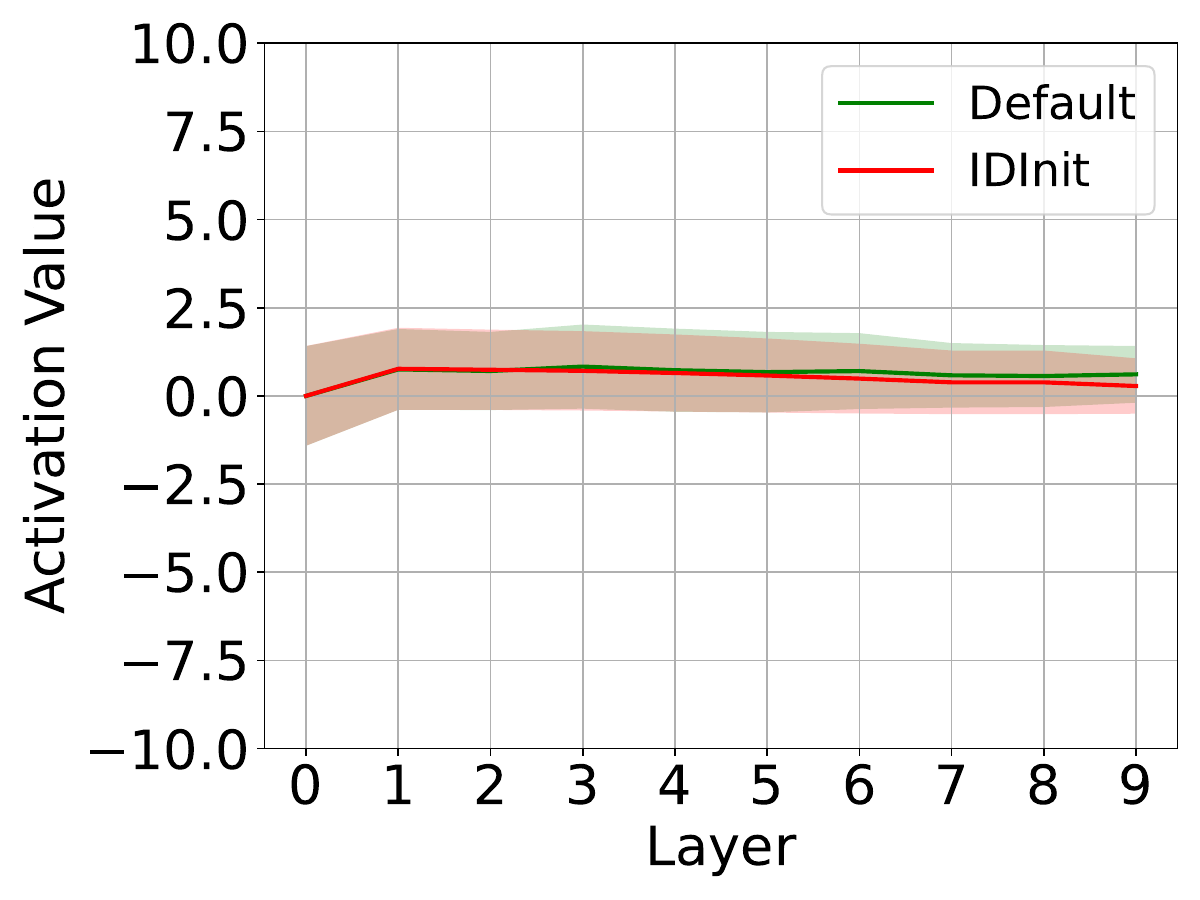}
	}
	\subfigure[ResConv-1.00]{
		\includegraphics[width=0.23\textwidth]{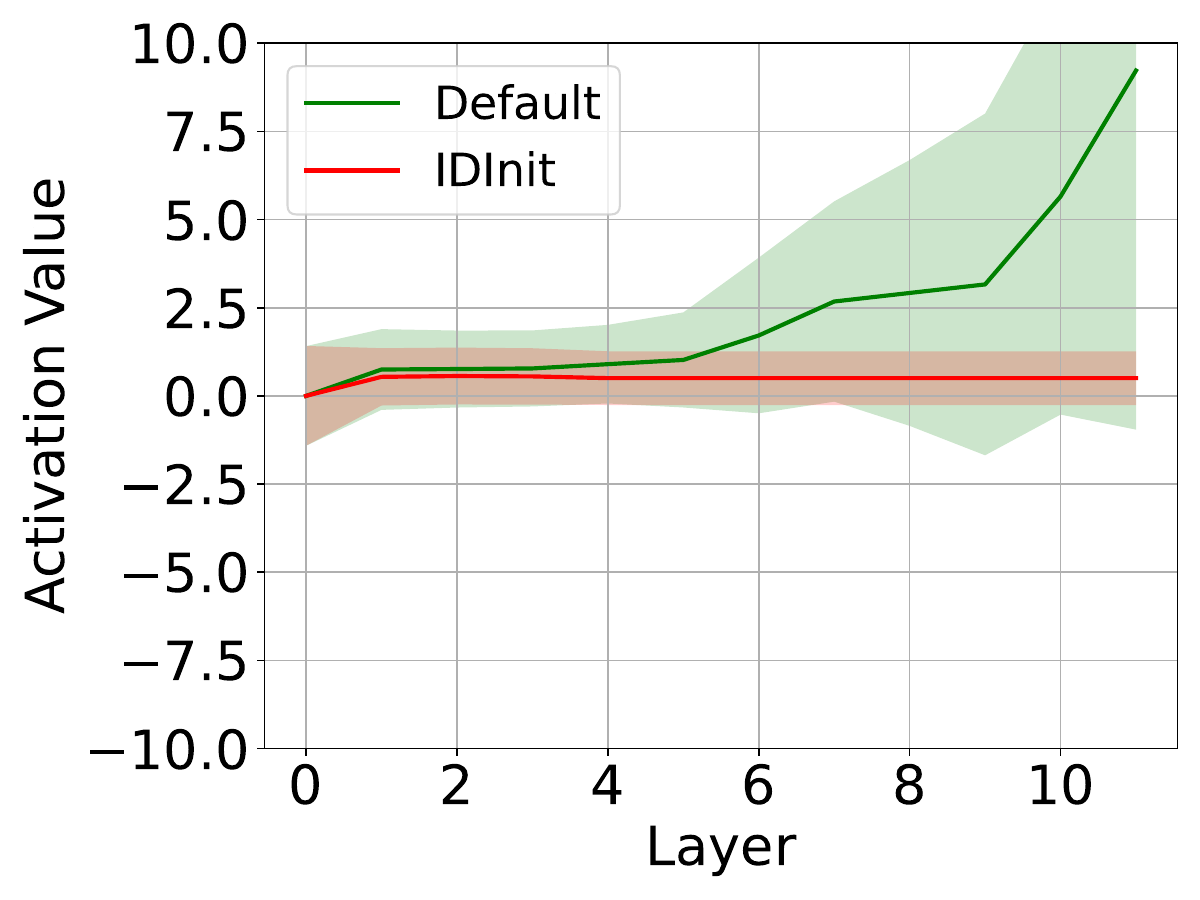}
	}
	\caption{Results of the analysis on variance propagation. The numerical value after the model name means the standard derivation of the noise. ``Default" means the default initialization of models, specifically, Xavier for FC and ResFC, and Kaiming for Conv and ResConv. The default methods can only work on non-residual networks FC and Conv, however, fail on residual networks ResFc and ResConv, for cause instability with giant standard derivation. By contrast, IDInit can consistently transit data-flow in an appropriate scale on all models and various noises, which shows sufficient robustness, and can provide models with stable and efficient training.}
	\label{fig:varpro}
\end{figure}

\subsection{Validation on Non-Residual Convolution}
\label{sec:nonresidual}

We use this experiment to show IDInit can achieve a good initial state for training on non-residual convolutional networks. In this experiment, we use AllConv~\citep{DBLP:journals/corr/SpringenbergDBR14} which consists of nine convolutional layers as the backbone network. We show the structure of AllConv in Table~\ref{tbl:allconv}. The dataset is Cifar10.  The optimizer is Stochastic Gradient Descent (SGD) with momentum 0.9, weight decay 5e-4, and learning rate 1e-1. The learning rate scheduler adopts a warm-up cosine reduction strategy. We run the model in 300 epochs on one Nvidia A100. We adopt Kaiming initialization and IDInit w/o $\operatorname{IDIC}_{\tau}$ initialization for comparison. Since there is no residual connection, we do not consider the $\operatorname{IDIZC}_{\varepsilon}$ function in this experiment. For each initialization, we have run them with 0, 10, 20, 30, 40, 50, and 60 warm-up epochs. The experiment is conducted on one Nvidia A100.

\begin{table}[h]
\caption{Architectures of the tensorial All-Conv networks. Window means the convolutional kernel window size. Channels indicate $\mathbf{c}_{in}$ and $\mathbf{c}_{out}$ of a standard convolutional kernel $\ca{C}\in \mathbb{R}^{\mathbf{c}_{in}\times \mathbf{c}_{out}\times k \times k}$. The avg pool denotes the average pooling operation.}
\label{tbl:allconv}
\begin{center}
\begin{tabular}{@{}ccc@{}}
\toprule
Layer & Window     & Channels                                                           \\ \midrule
conv1 & 3$\times$3 & 3$\times$ 96                                                       \\ \midrule
conv2 & 3$\times$3 & 96$\times$ 96                                                      \\ \midrule
conv3 & 3$\times$3 & 96$\times$ 96                                                      \\ \midrule
conv4 & 3$\times$3 & 96$\times$ 192                                                     \\ \midrule
conv5 & 3$\times$3 & 192$\times$ 192                                                    \\ \midrule
conv6 & 3$\times$3 & 192$\times$ 192                                                    \\ \midrule
conv7 & 3$\times$3 & 192$\times$ 192                                                    \\ \midrule
conv8 & 1$\times$1 & 192$\times$ 192                                                    \\ \midrule
conv9 & 1$\times$1 & \begin{tabular}[c]{@{}c@{}}192$\times$ 10 \\ avg pool\end{tabular} \\ \bottomrule
\end{tabular}
\end{center}

\end{table}

\begin{figure}[h]
	\centering
        \subfigure[Test Top-1]{
		\includegraphics[width=0.4\textwidth]{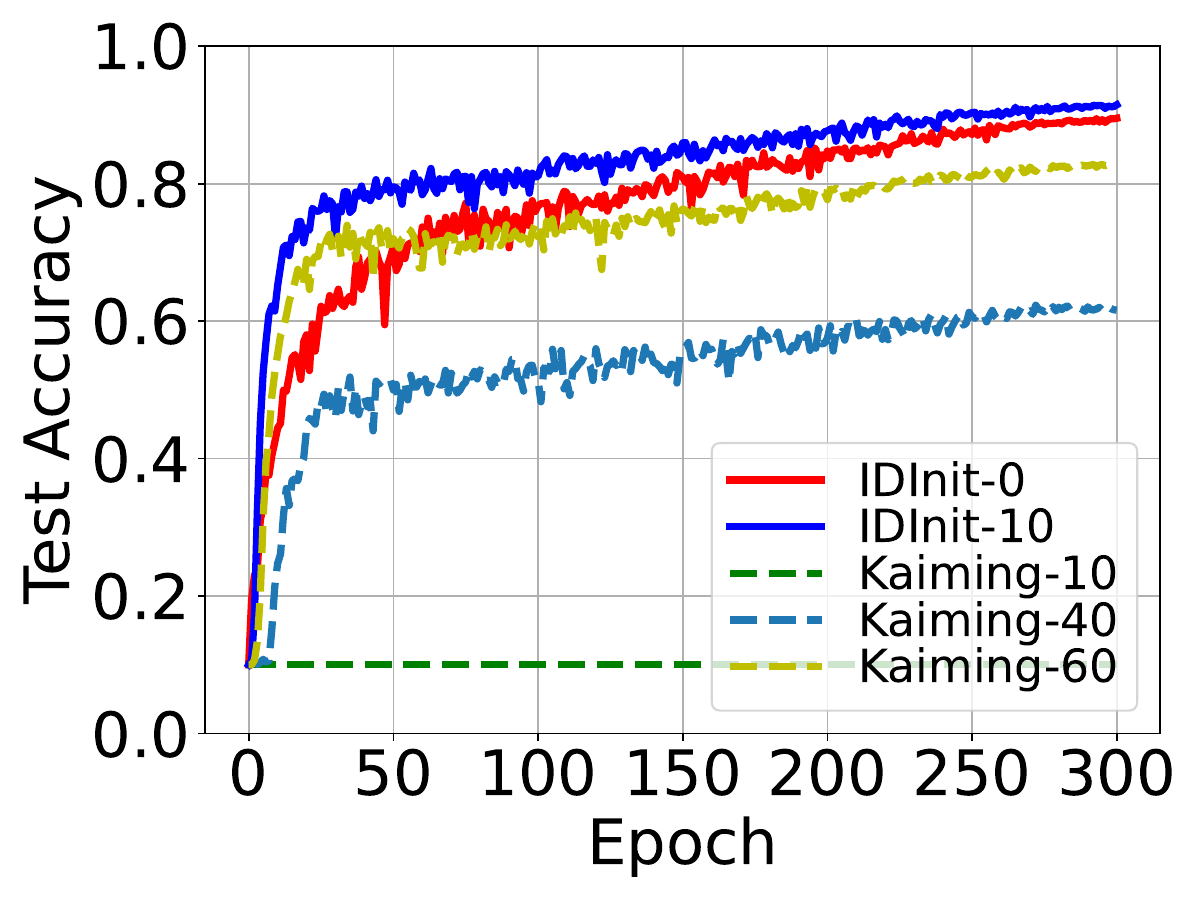}
		\label{fig:allconv_acc}
	}
	\subfigure[Test Best]{
		\includegraphics[width=0.4\textwidth]{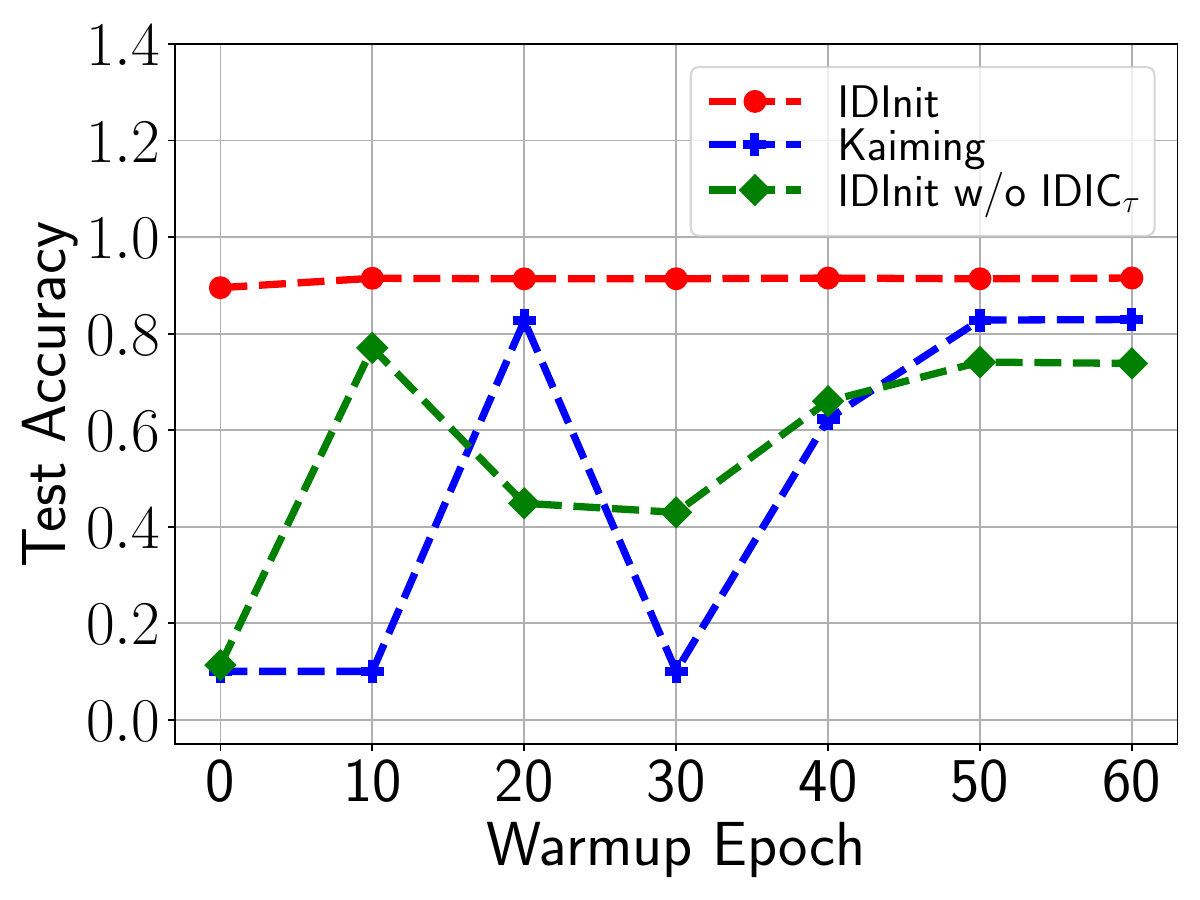}
		\label{fig:allconv_best}
	}
	\caption{Results of AllConv on Cifar10. The number behind the initialization denotes the warm-up epochs.}
    \label{fig:allconvcifar10}
\end{figure}

Results are shown in Figure~\ref{fig:allconvcifar10}, without a warm-up strategy which is a strong trick for training, both Kaiming and IDInit w/o $\operatorname{IDIC}_{\varepsilon}$ fail to train the model. By contrast, our initialization can train AllConv and maintain the highest performance in all situations, showing a strong effect on stability and performance. As IDInit w/o $\operatorname{IDIC}_{\varepsilon}$ performs poorly, we demonstrate the patch-maintain strategy mentioned in Sec.~\ref{sec:idiconv} can be good for increasing feature diversity. This experiment shows the identical method can be a feasible initialization for non-residual networks.

\subsection{Analysis on Variance Propagation}
\label{sec:varana}
Here we conduct an experiment on Cifar10 to demonstrate data-flow will keep stable. We use 4 types of networks: (1) FC: 10-layer fully-connected layers; (2) ResFC: 10 residual blocks (two fully-connected layers in a block); (3) Conv: 9-layer AllConv in Sec.~\ref{sec:nonresidual}; (4) ResConv: 10 residual blocks (two convolutional layers in a block). For (1) and (2) two fully-connected networks, we reshape Cifar10 data as $\mathbf{X} \in \mathbb{R}^{32\times 96}$ as input and does not use any activation function. For (1), hidden lengths are $\{200, 400, 600, 800, 1000, 1000, 800, 600, 400, 200\}$.  
For (2), hidden lengths are all set to 96. For (3) and (4) two convolution networks, we directly input images to them, and use ReLU as the activation function. For (3), we directly use AllConv as shown in Table~\ref{tbl:allconv}. For (4), we first use convolution to transfer an image to 16 channels, and then set the channels of all convolution within residual blocks to 16. For comparison, we use Xavier for (1) and (2), and Kaiming for (3) and (4) in terms of the activation function. We also employ noises with 0 mean, and $\{0.00, 0.01, 0.10, 1.00\}$ for comparing robustness. In the experiment, we run 500 rounds for each model. The experiment is conducted on one Nvidia A100.

Results are shown in Figure~\ref{fig:varpro}. The regular methods Xavier and Kaiming can only work on non-residual networks. On residual networks, they both cause giant standard derivation, leading to instability. By contrast, the proposed IDInit can consistently transit data-flow in an appropriate scale on all models and various noises, which shows sufficient robustness, and can provide models with stable and efficient training.

\subsection{Analysis on Weight Distribution}
\label{sec:weightdis}
In this experiment, we conduct an experiment on Cifar10 with ResNet-20 to show the weight distribution of IDInit. We use an SGD optimizer with a learning rate 0.2, and weight decay 5e-4. The batch size is 1024. Training epochs are 200. The learning rate is reduced with a cosine function. The experiment is conducted on one Nvidia A100.

\begin{figure}[h]
	\centering
        \subfigure[Kaiming-E4]{
		\includegraphics[width=0.23\textwidth]{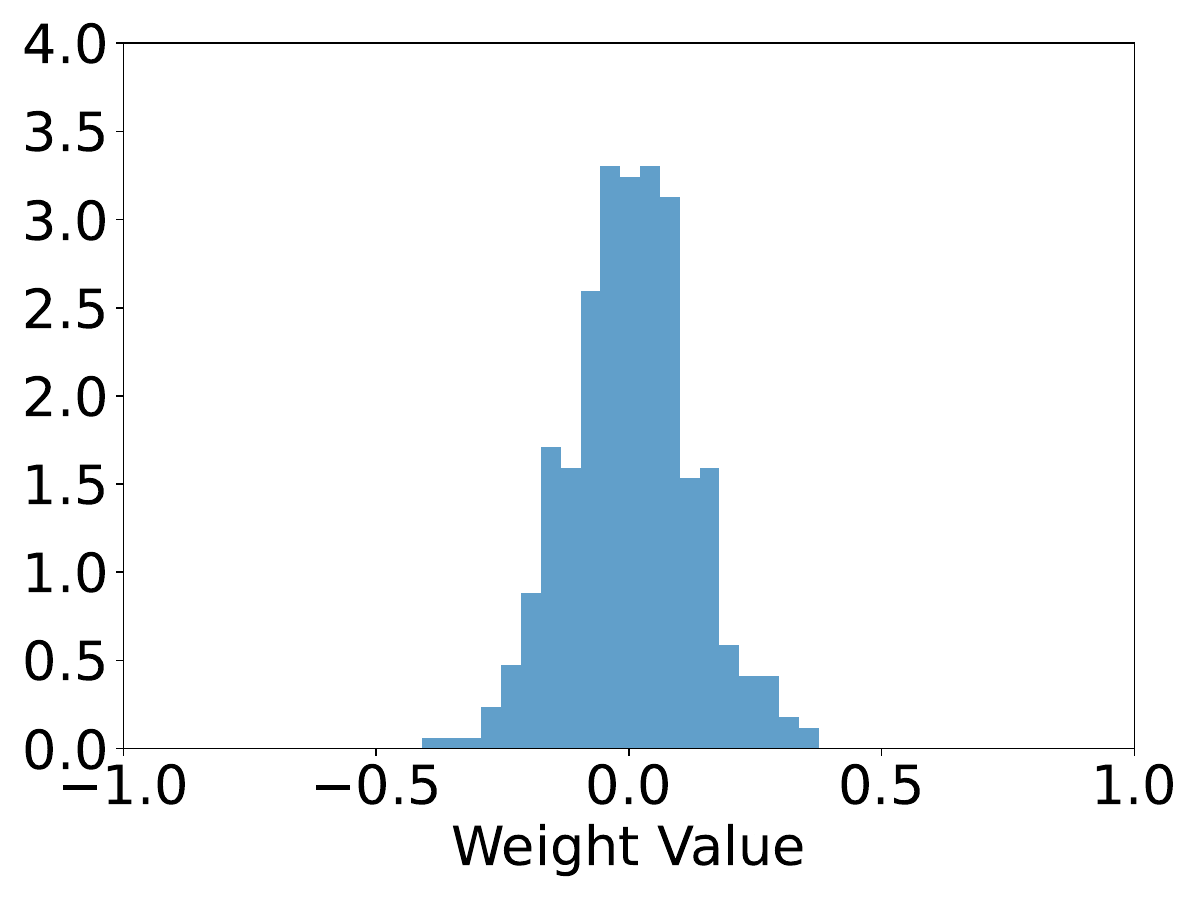}
	}
	\subfigure[Kaiming-E24]{
		\includegraphics[width=0.23\textwidth]{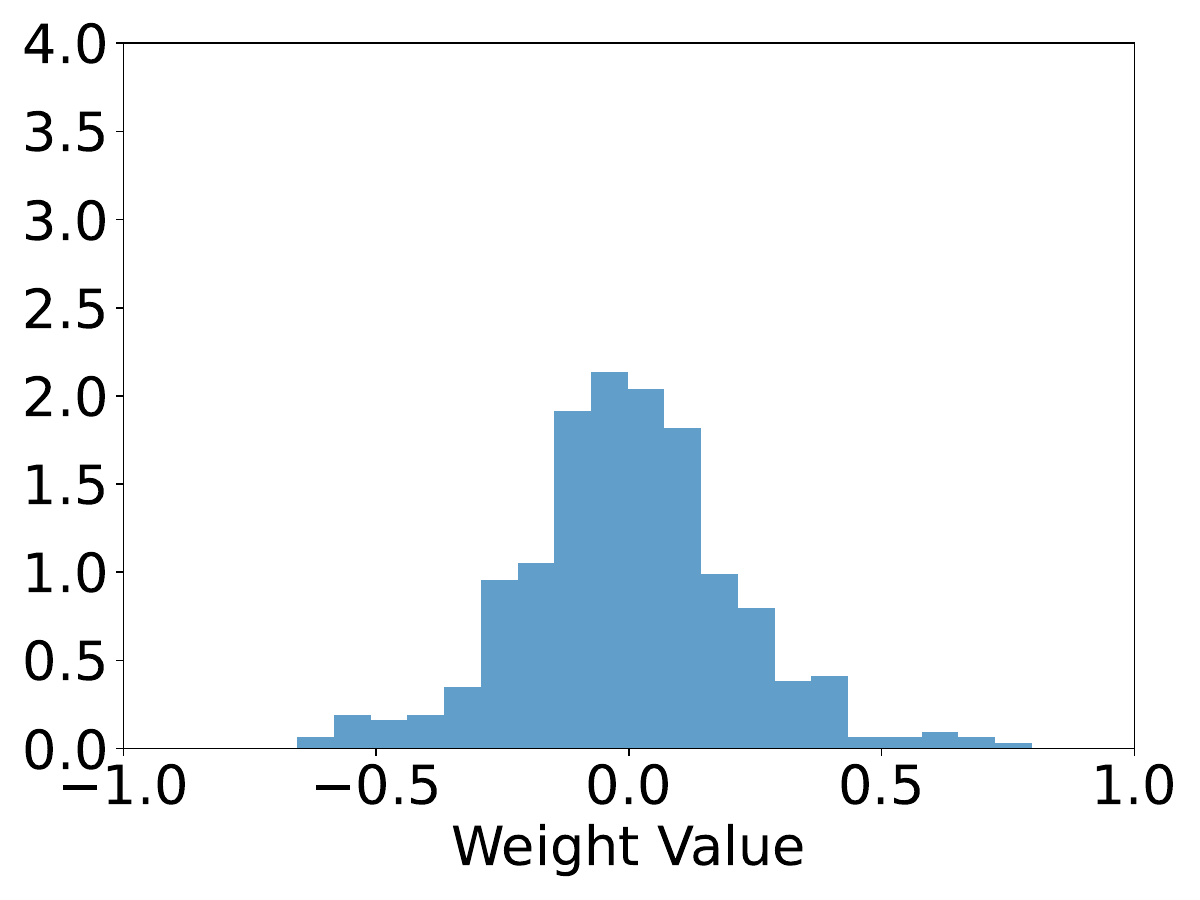}
	}
	\subfigure[Kaiming-E104]{
		\includegraphics[width=0.23\textwidth]{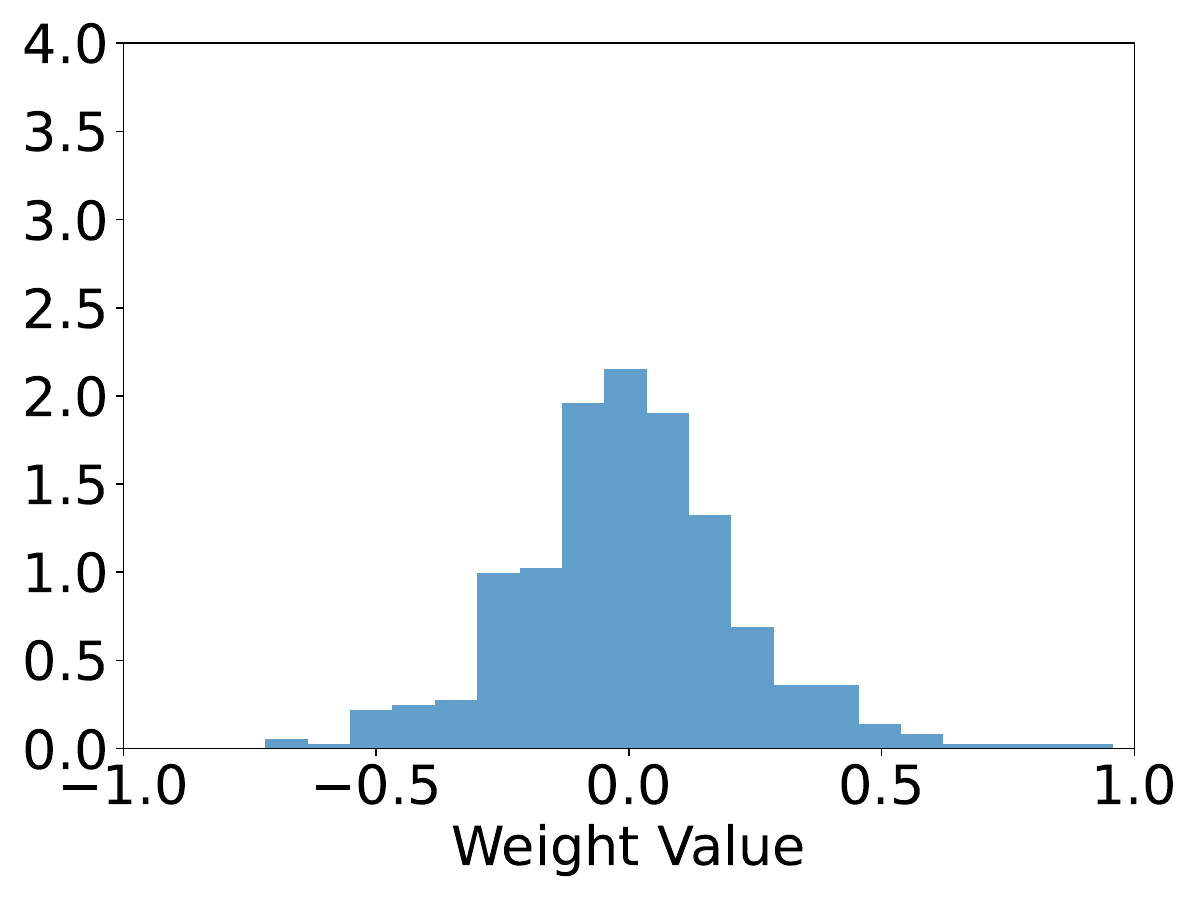}
	}
	\subfigure[Kaiming-E194]{
		\includegraphics[width=0.23\textwidth]{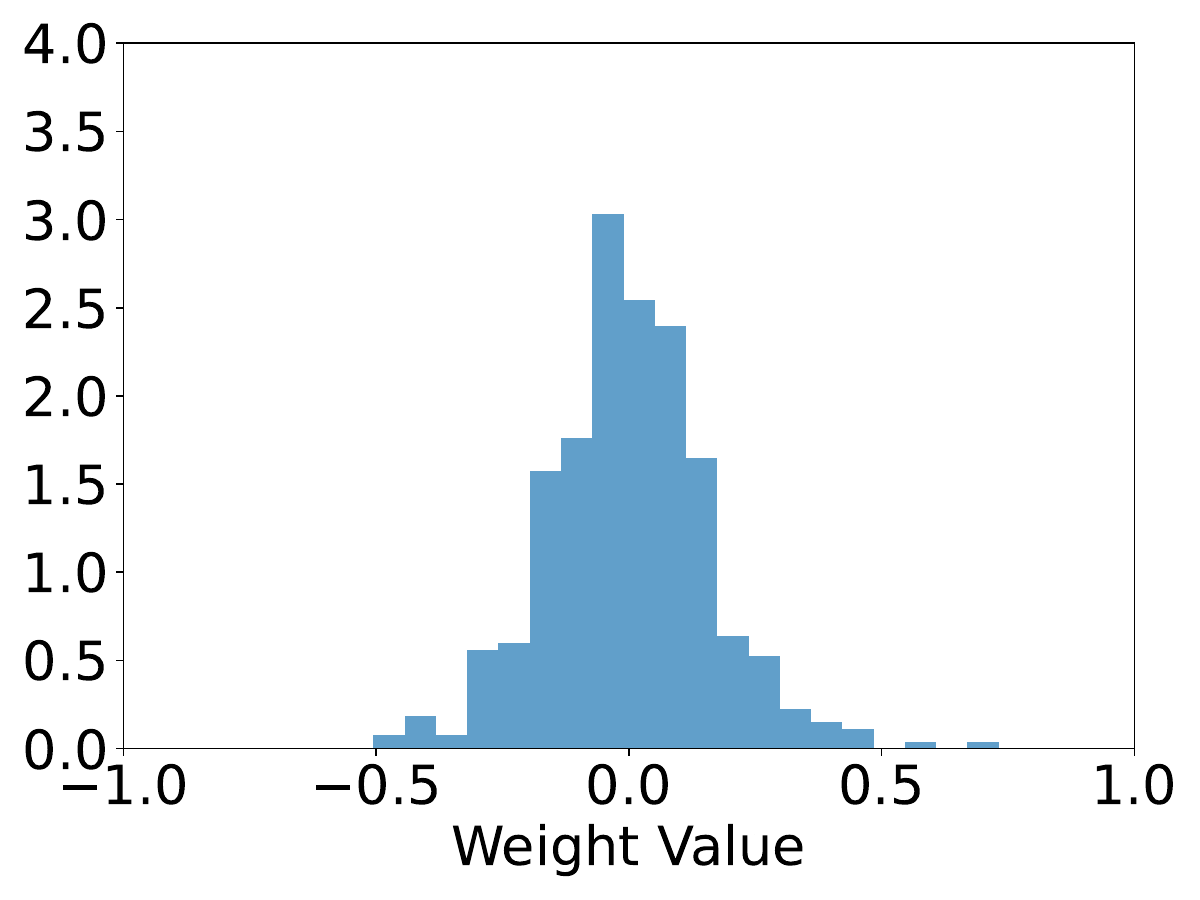}
	}
	
	\subfigure[IDInit-E4]{
		\includegraphics[width=0.23\textwidth]{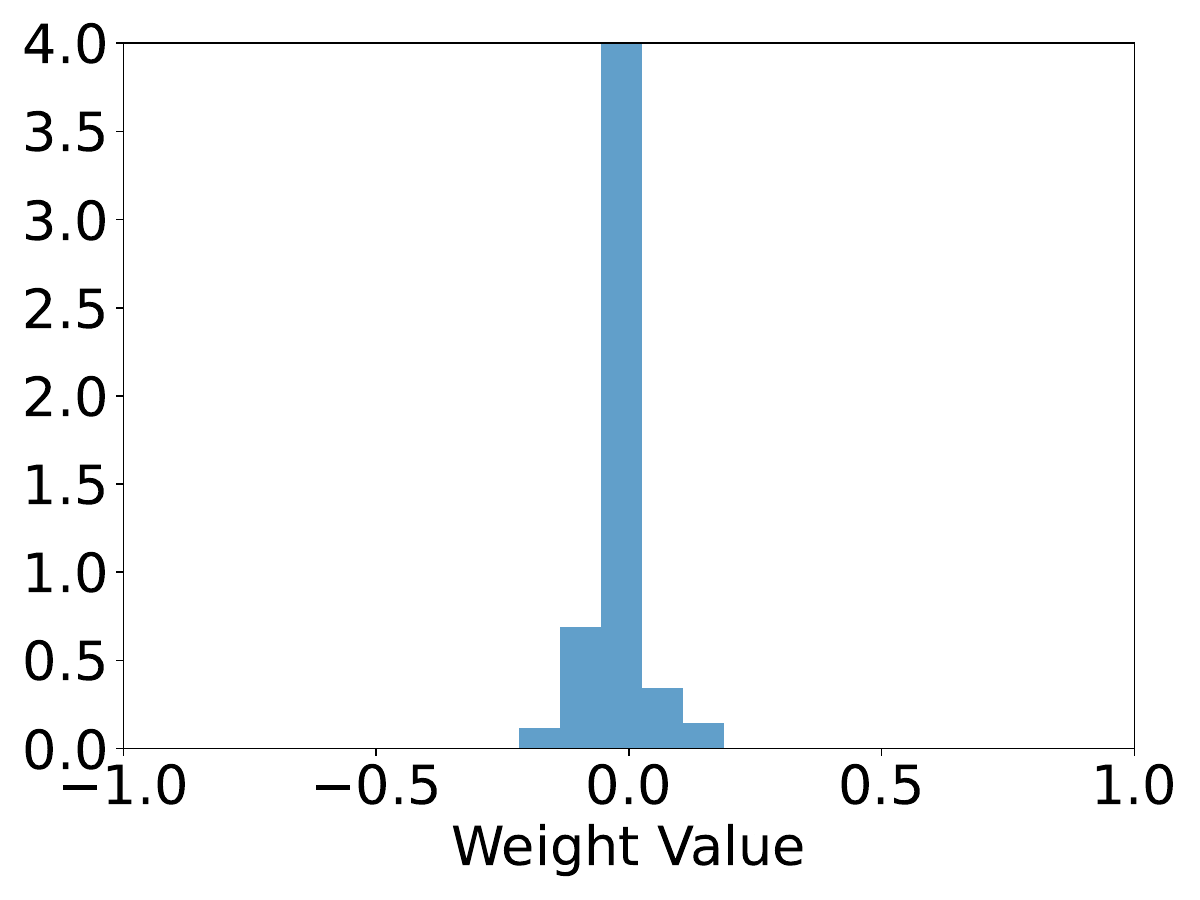}
	}
	\subfigure[IDInit-E24]{
		\includegraphics[width=0.23\textwidth]{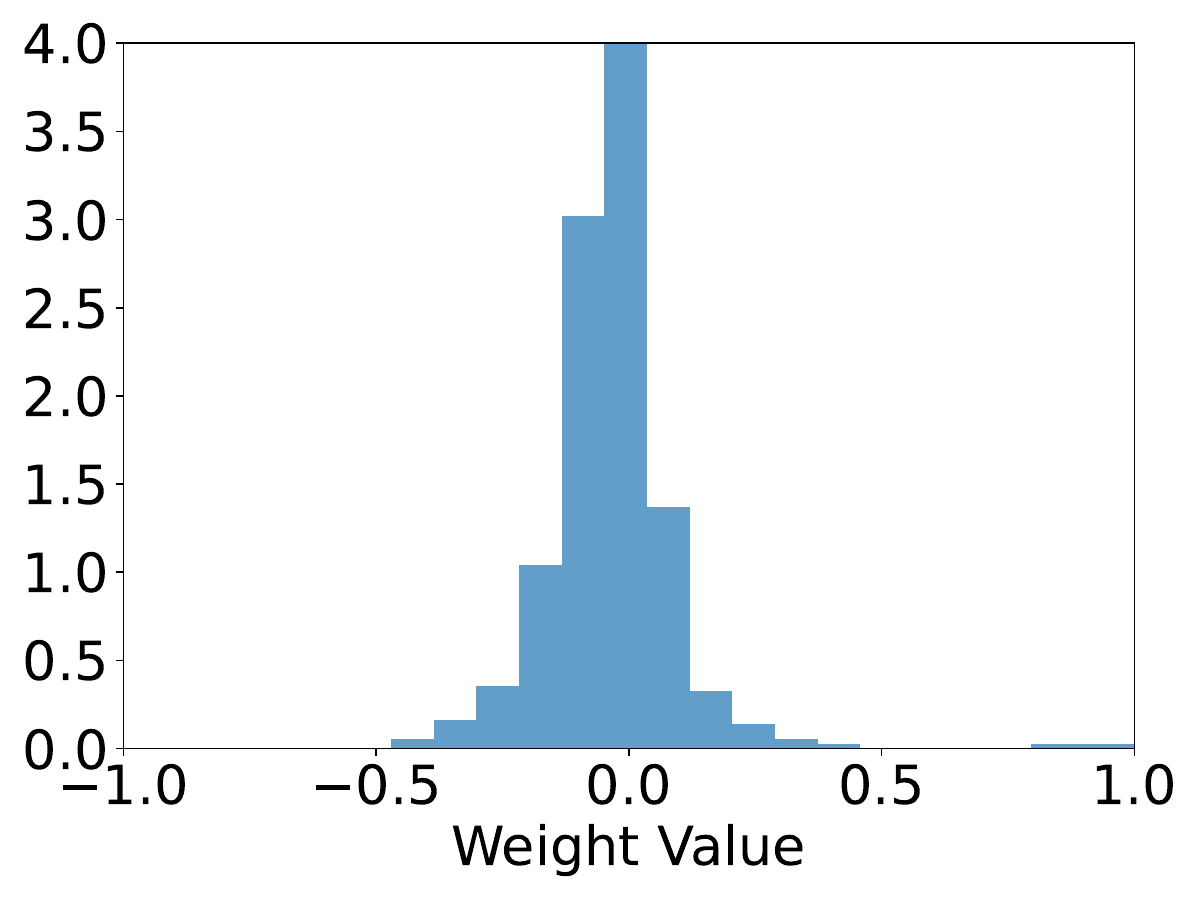}
	}
	\subfigure[IDInit-E104]{
		\includegraphics[width=0.23\textwidth]{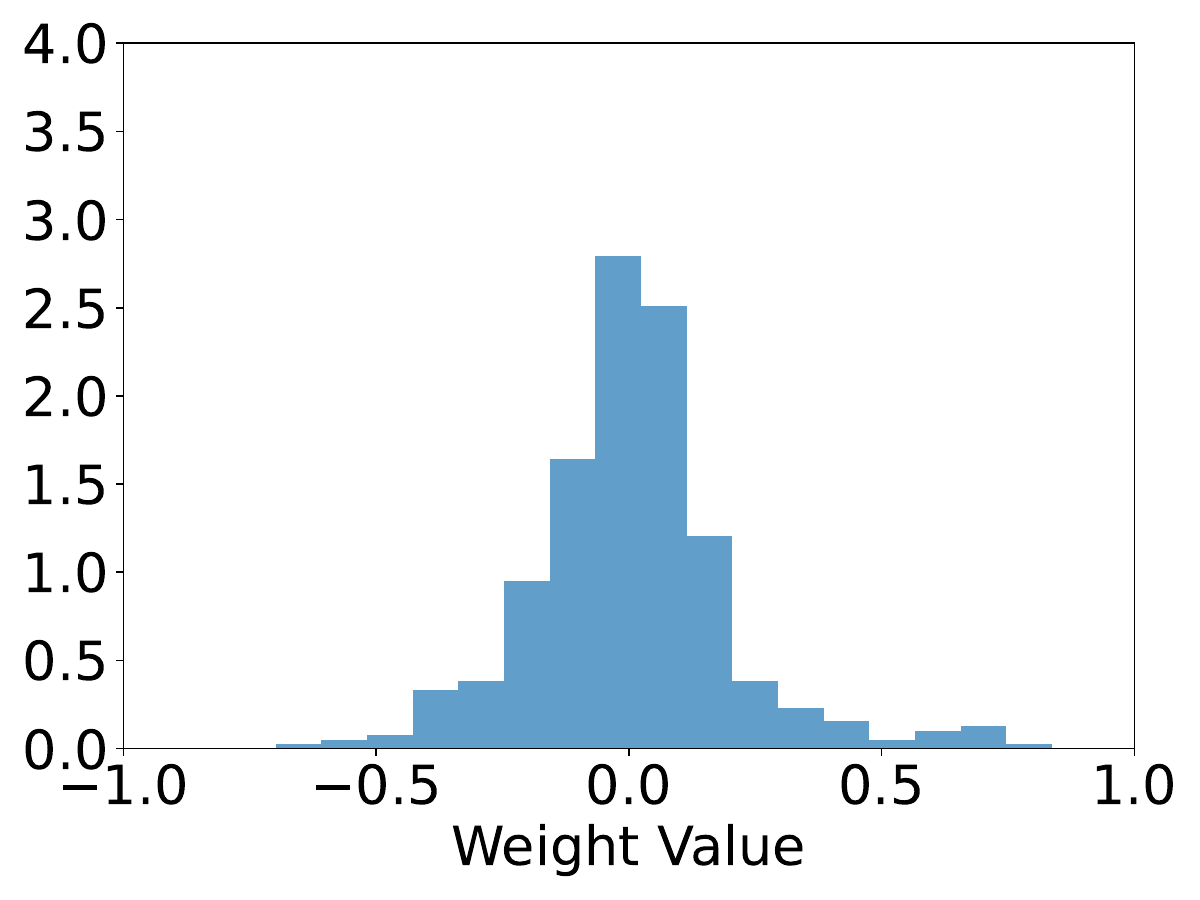}
	}
	\subfigure[IDInit-E194]{
		\includegraphics[width=0.23\textwidth]{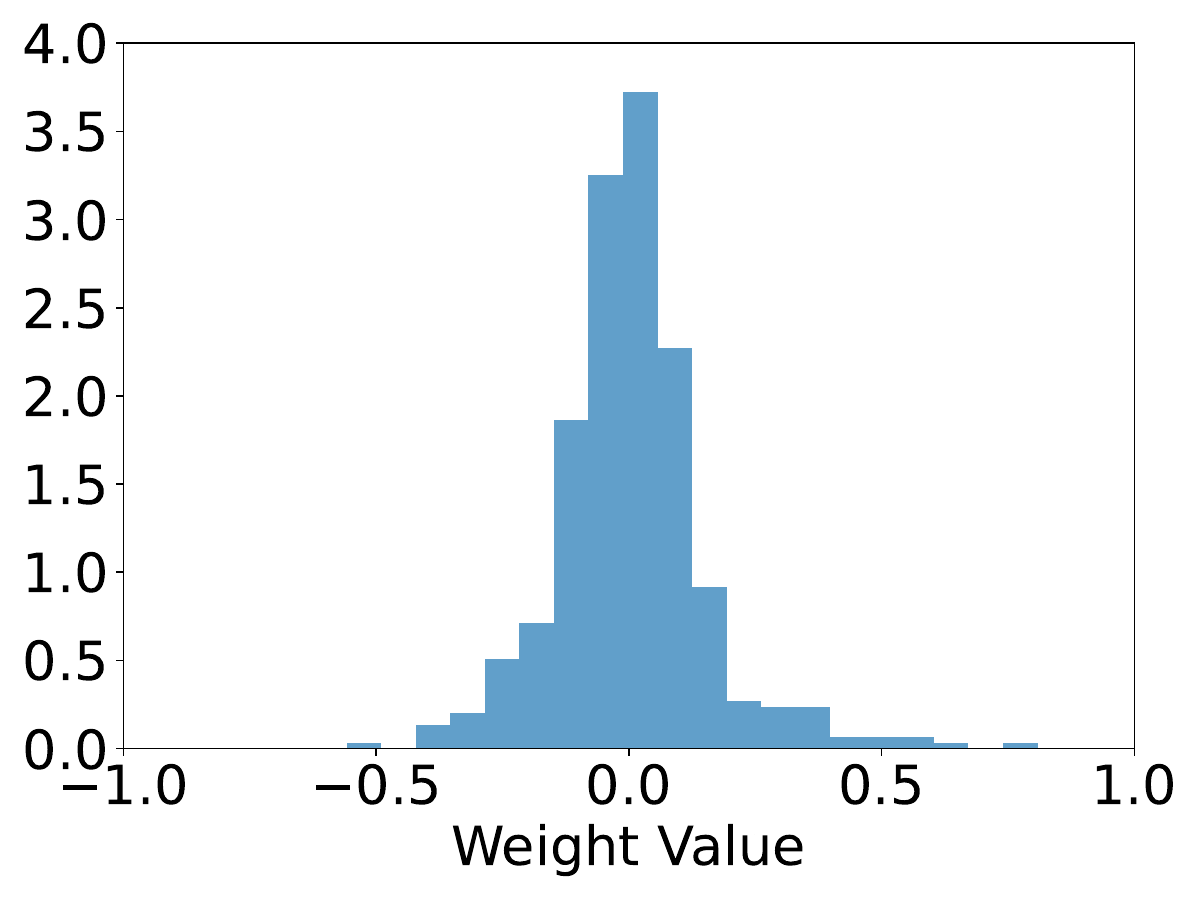}
	}

	\caption{Histograms of the first convolution weights in ResNet-20.  ``E'' means the epoch index. IDInit contains more zero values in each epoch compared with Kaiming initialization.}
	\label{fig:weightdis}
\end{figure}

The results are shown in Figure~\ref{fig:weightdis}, weights initialized with IDInit are almost full of zero at the beginning, while Kaiming uses a Gaussian distribution. At the end of the training, IDInit still contains more zero values than Kaiming, which is beneficial for memory occupation since a 0 value will not cost memory space.

\subsection{Analysis on input-output Jacobian}
\label{sec:jacana}
Here we conduct an experiment on Cifar10 with 64 blocks in Figure~\ref{fig:id-control} to demonstrate IDInit follows the dynamical isometry. We use the open-source code\footnote{\url{https://github.com/tbachlechner/ReZero-examples/blob/master/ReZero-Deep_Fast_NeuralNetwork.ipynb}}. We remove batch normalization for the more clear difference between IDInit and Kaiming. We use an Adagrad optimizer with a learning rate 0.01. The batch size is 100. The activation is ReLU. The experiment is conducted on one Nvidia A100.

\begin{figure}[h]
	\centering
        \subfigure[Default-E1]{
		\includegraphics[width=0.23\textwidth]{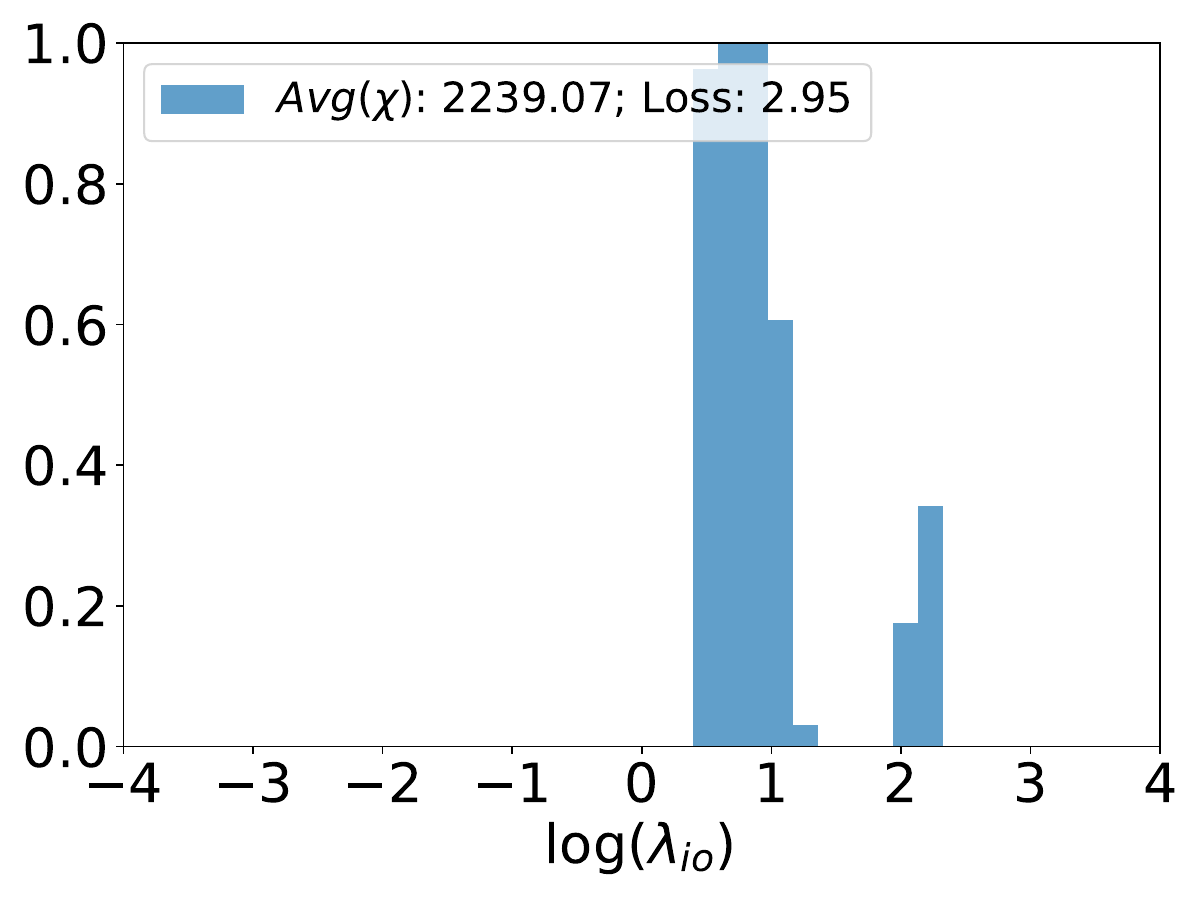}
	}
	\subfigure[Default-E2]{
		\includegraphics[width=0.23\textwidth]{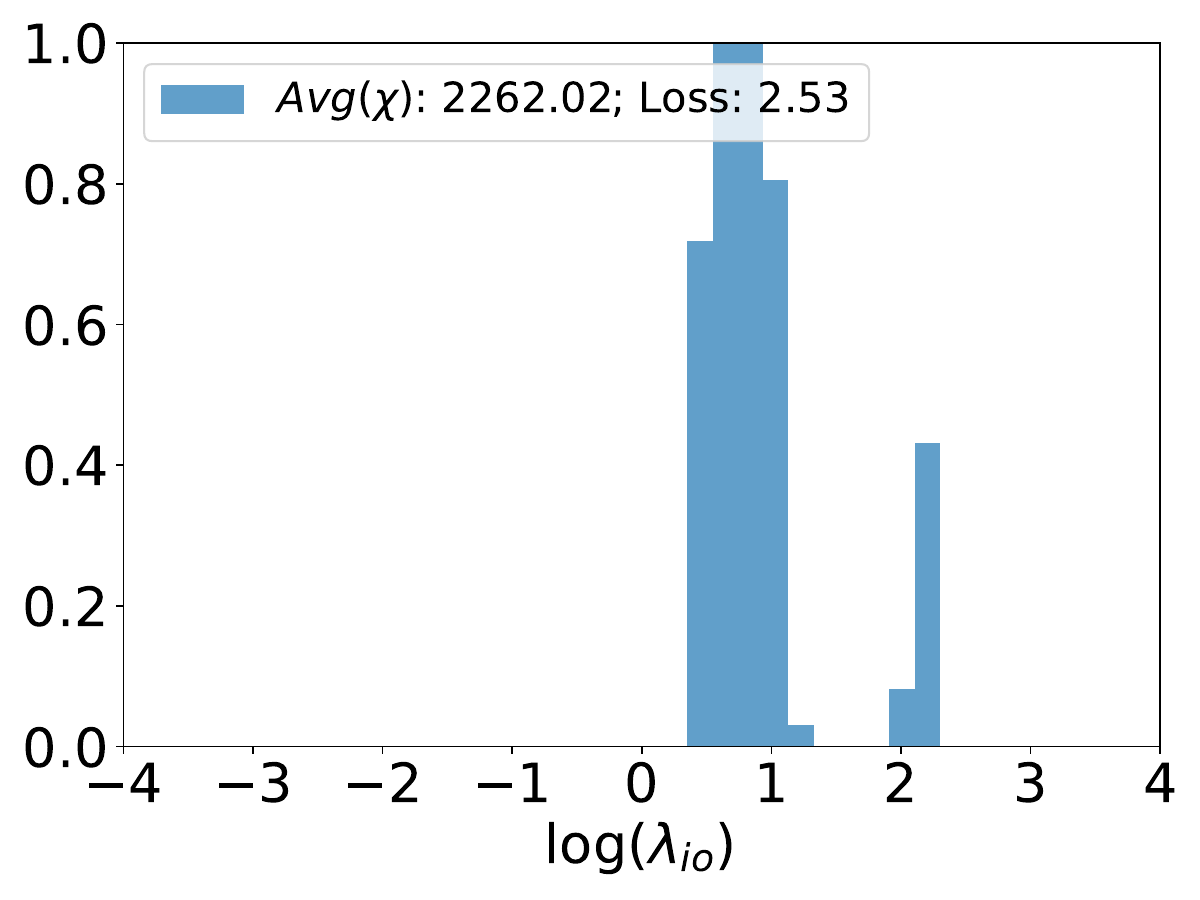}
	}
	\subfigure[Default-E3]{
		\includegraphics[width=0.23\textwidth]{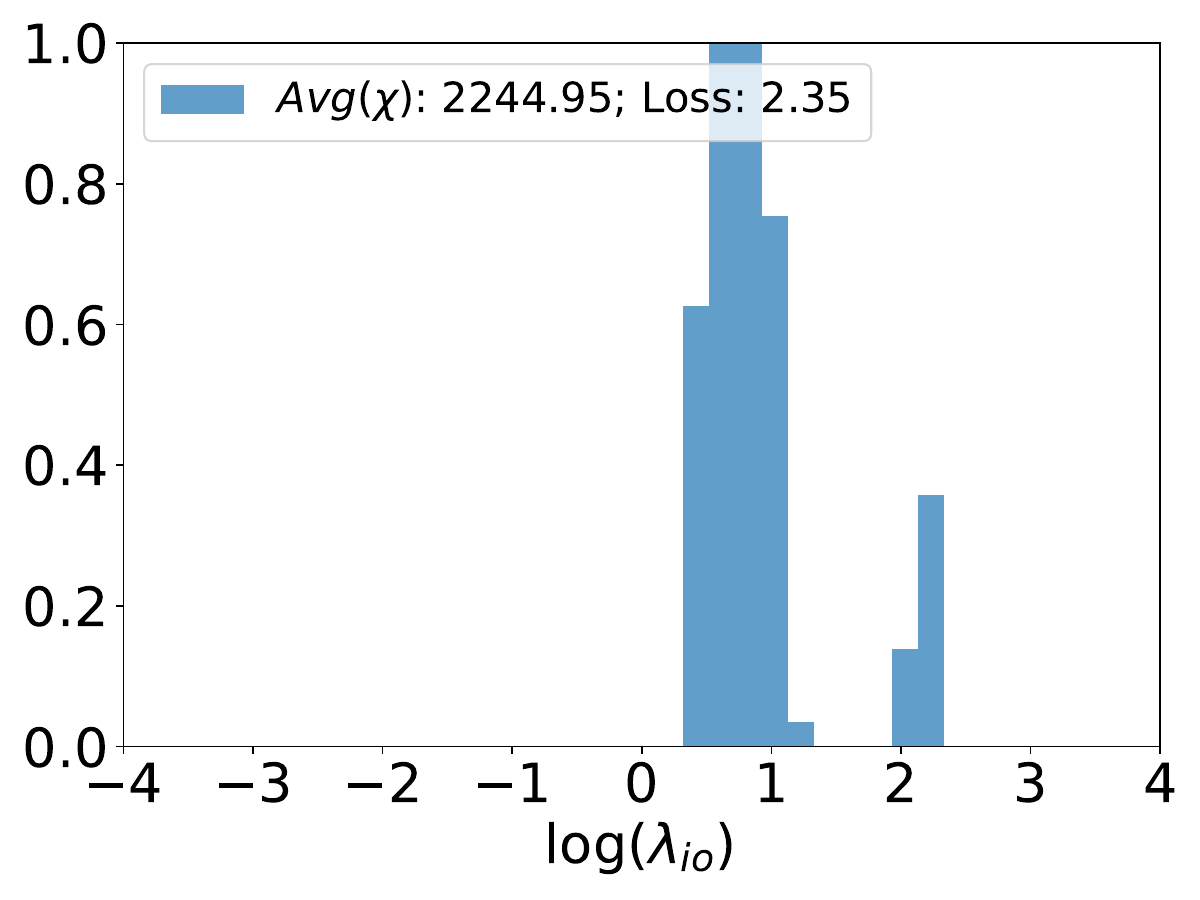}
	}
	\subfigure[Default-E4]{
		\includegraphics[width=0.23\textwidth]{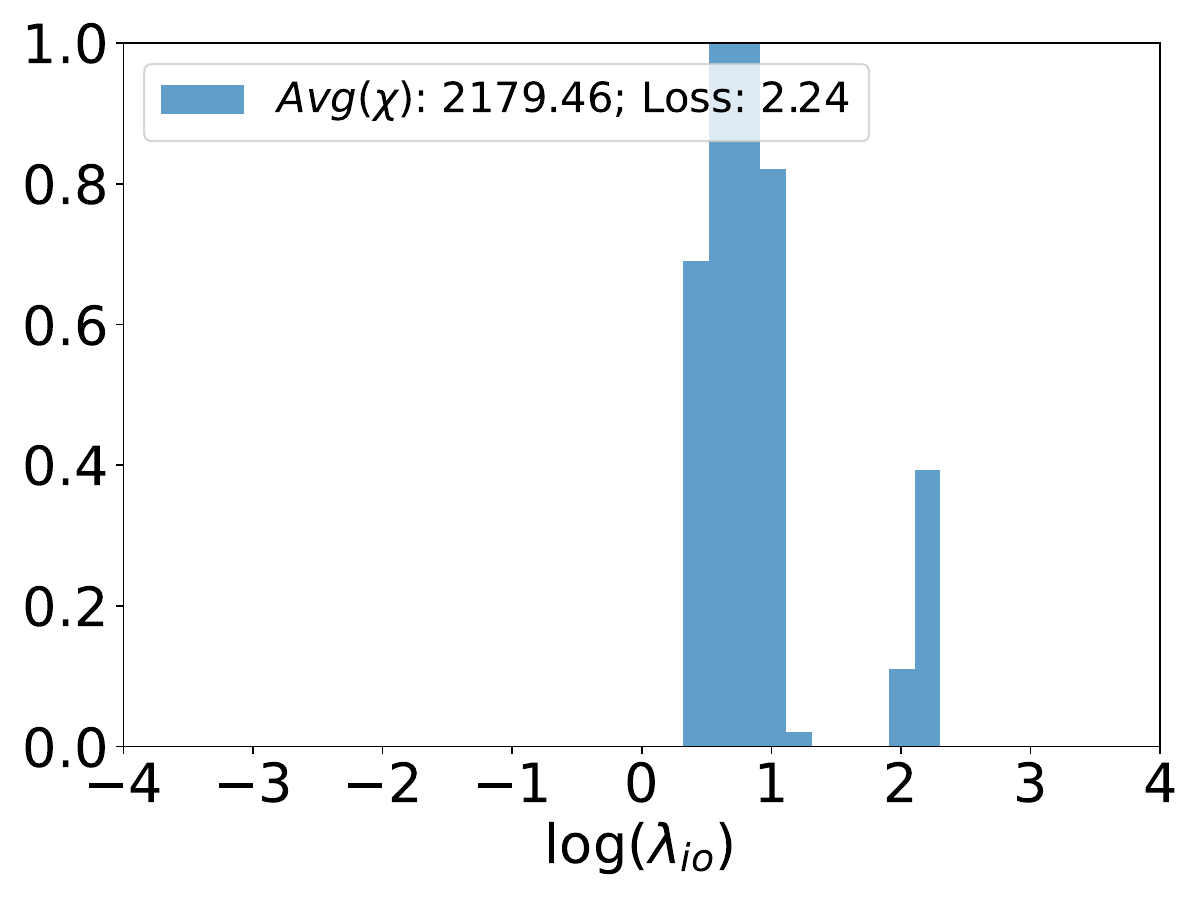}
	}
	
	\subfigure[IDInit-E1]{
		\includegraphics[width=0.23\textwidth]{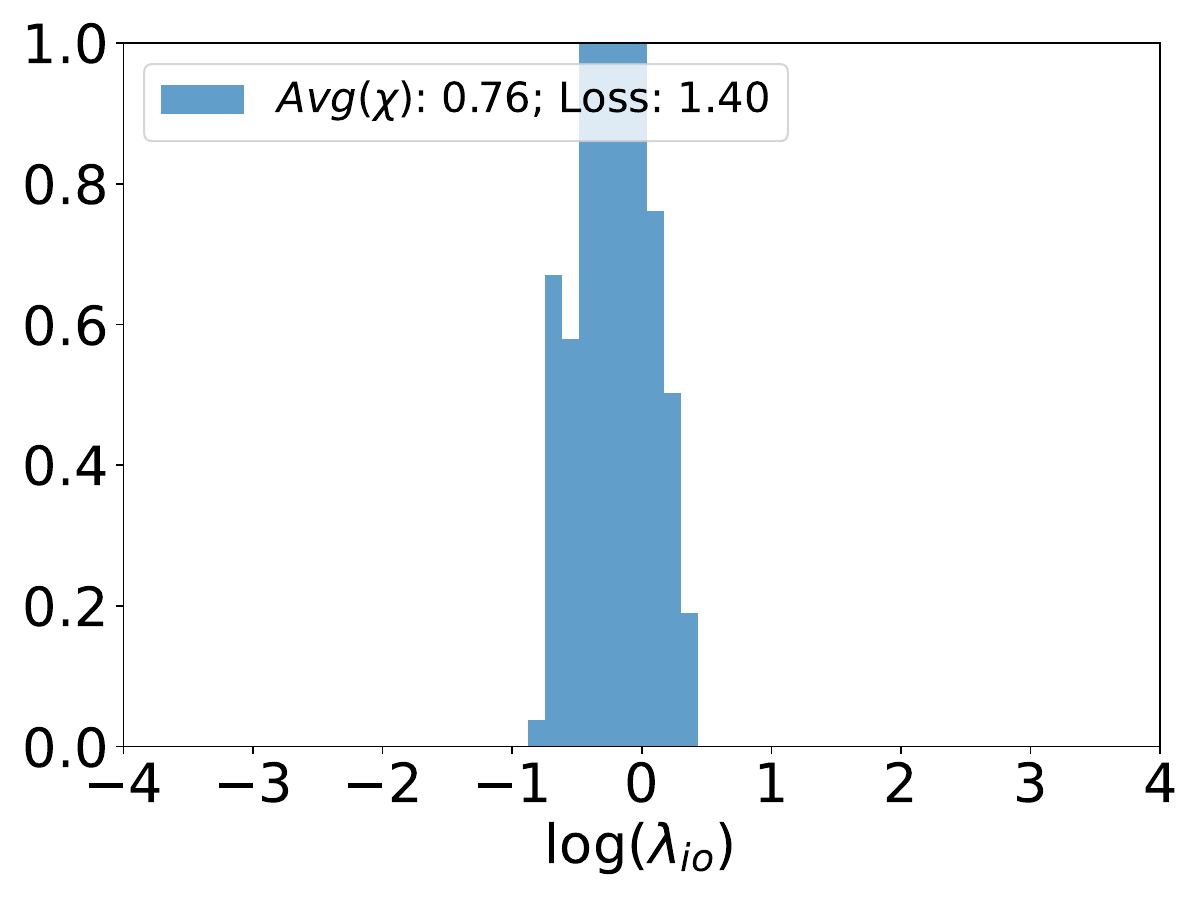}
	}
	\subfigure[IDInit-E2]{
		\includegraphics[width=0.23\textwidth]{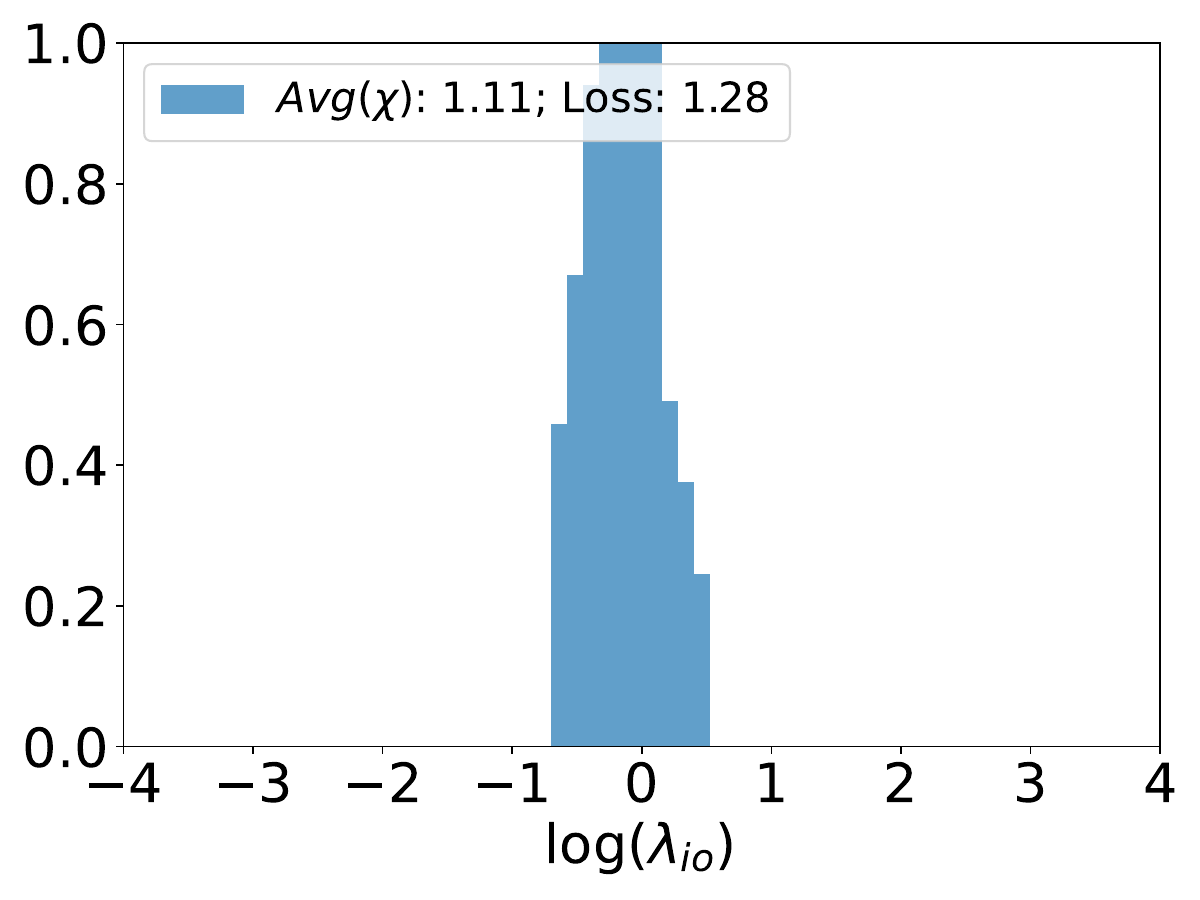}
	}
	\subfigure[IDInit-E3]{
		\includegraphics[width=0.23\textwidth]{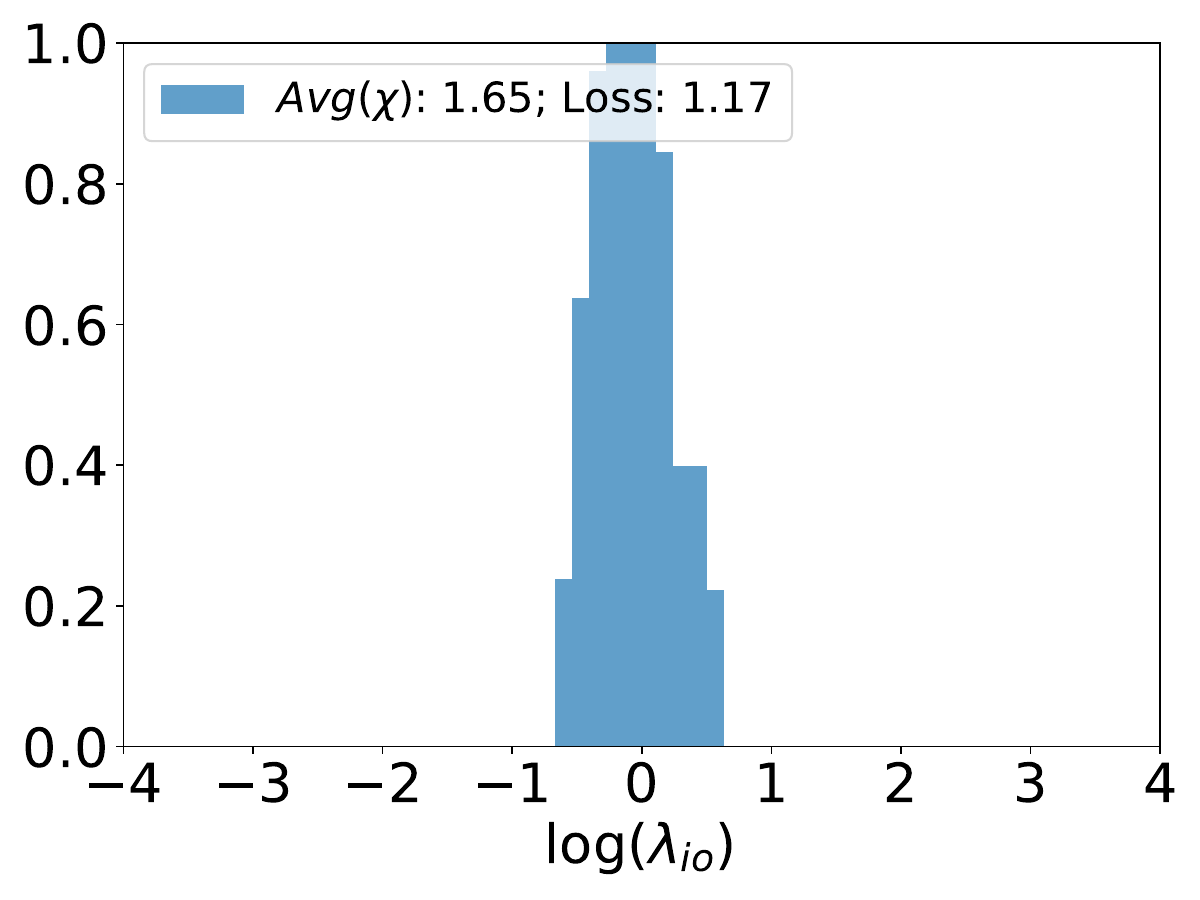}
	}
	\subfigure[IDInit-E4]{
		\includegraphics[width=0.23\textwidth]{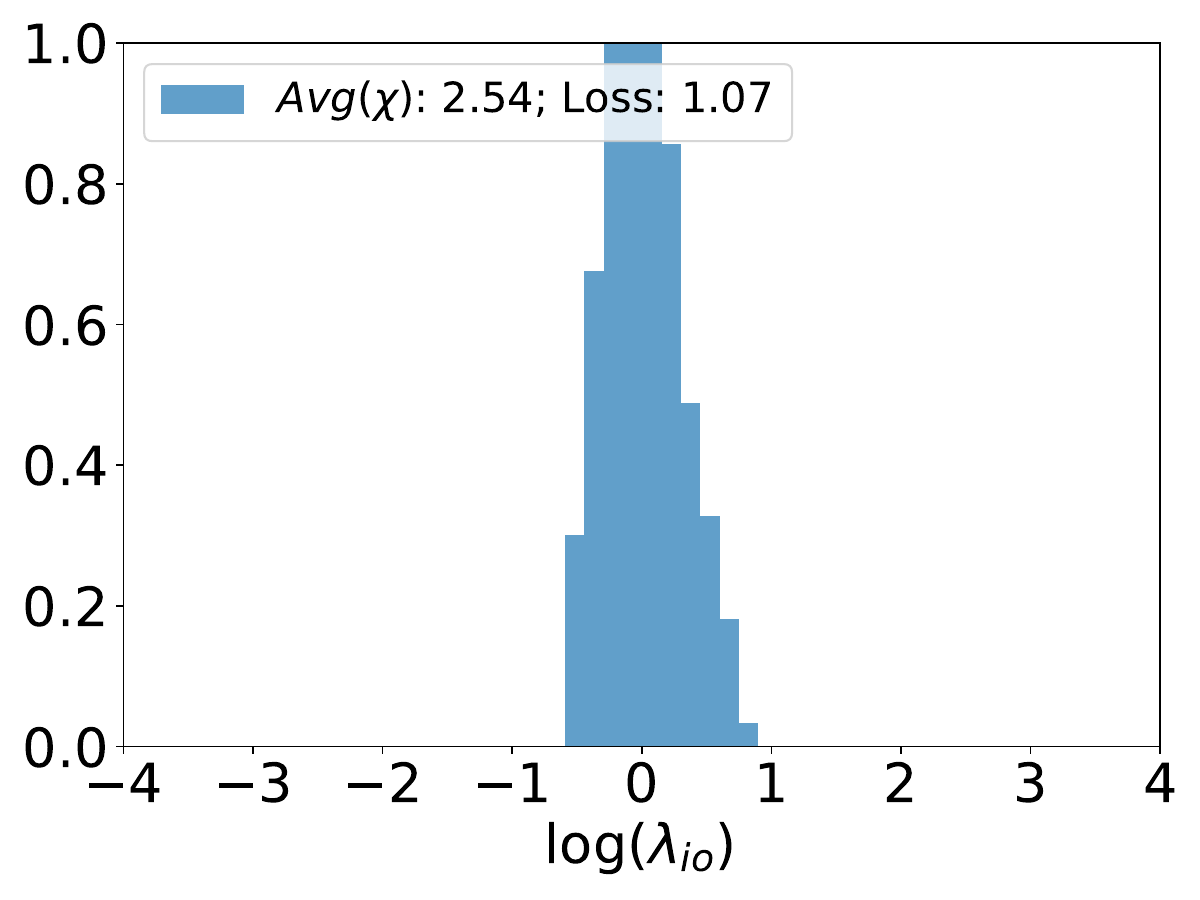}
	}

	\caption{Histograms of log singular values ($\log(\lambda_{io})$) for the input-output Jacobian. ``E'' means the epoch index. Compared with Default initialization, IDInit has a significantly smaller squared singular value $\chi$, which can achieve a faster reduction of the loss.}
	\label{fig:Jacbian}
\end{figure}

As shown in Figure~\ref{fig:Jacbian}, Default initialization cause a high squared singular value $\chi$, reaching more than 2000. Compared to Default, IDInit only derives $\chi$ around 1, indicating correspondence to the dynamical isometry. In addition, the loss of IDInit decreases faster than Default, which shows a good convergent ability.

\subsection{Failure of Long Residual Stem}
\label{sec:failuresubstem}

We conduct this experiment to show the failure case when the residual stem is long to show the importance of the stability of the residual stem. In this experiment, we conduct an experiment on Cifar10. We use a residual network named Res-112 as in Table~\ref{tbl:longstemarch}. We set 109 layers in the residual stem. Batch normalization is not applied for fairly validating the stability of initialization methods. We use an SGD optimizer with a learning rate 0.2, and weight decay 1e-8. The batch size is 768. Training epochs are 35. The learning rate is reduced with a cosine function. One Nvidia A100 is used.

\begin{figure}[h]
	\centering
        \subfigure[Test Accuracy]{
        \label{fig:long-acc}
		\includegraphics[width=0.4\textwidth]{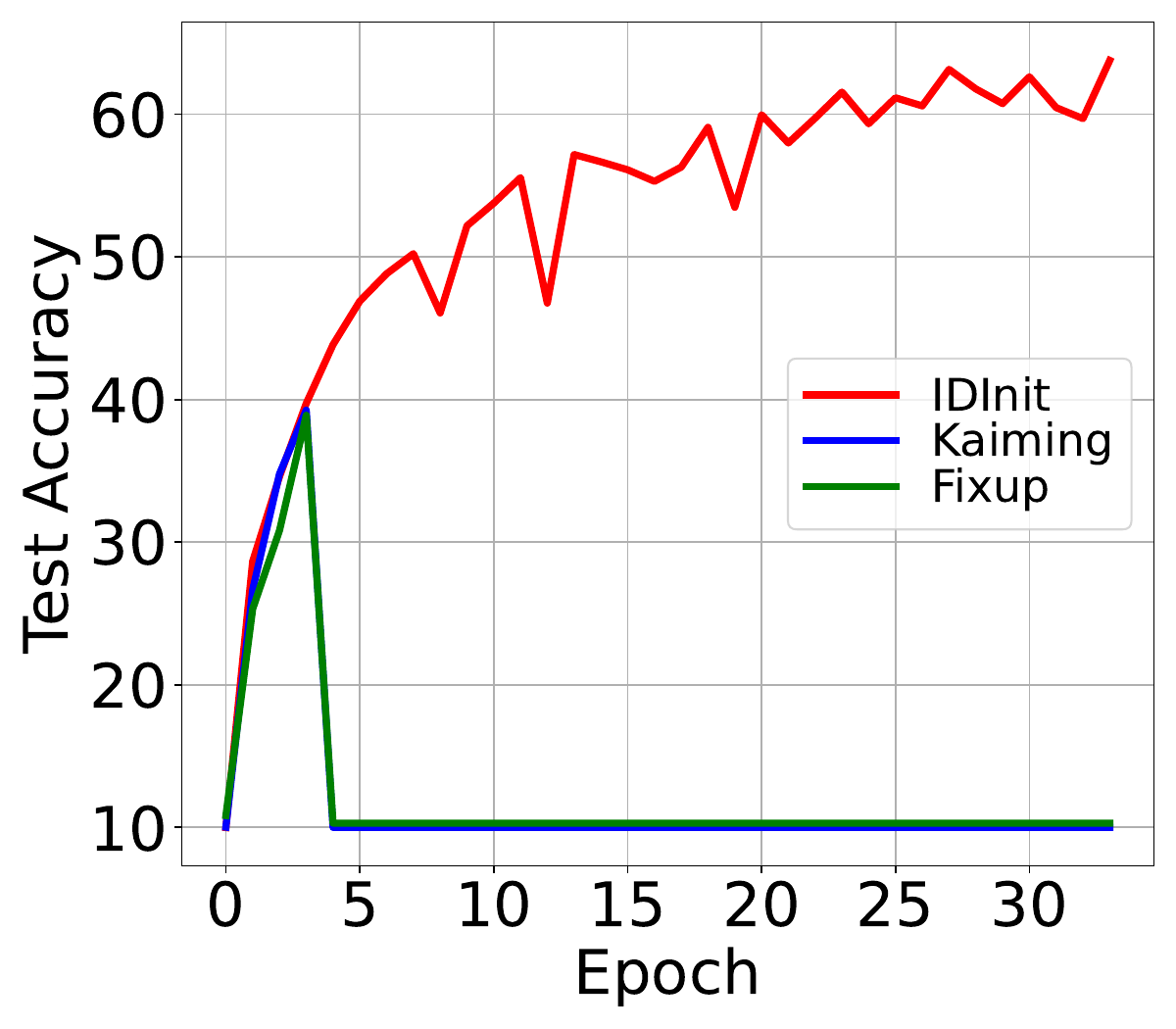}
	}~~~~
	\subfigure[Standard Derivation]{
 \label{fig:long-std}
		\includegraphics[width=0.4\textwidth]{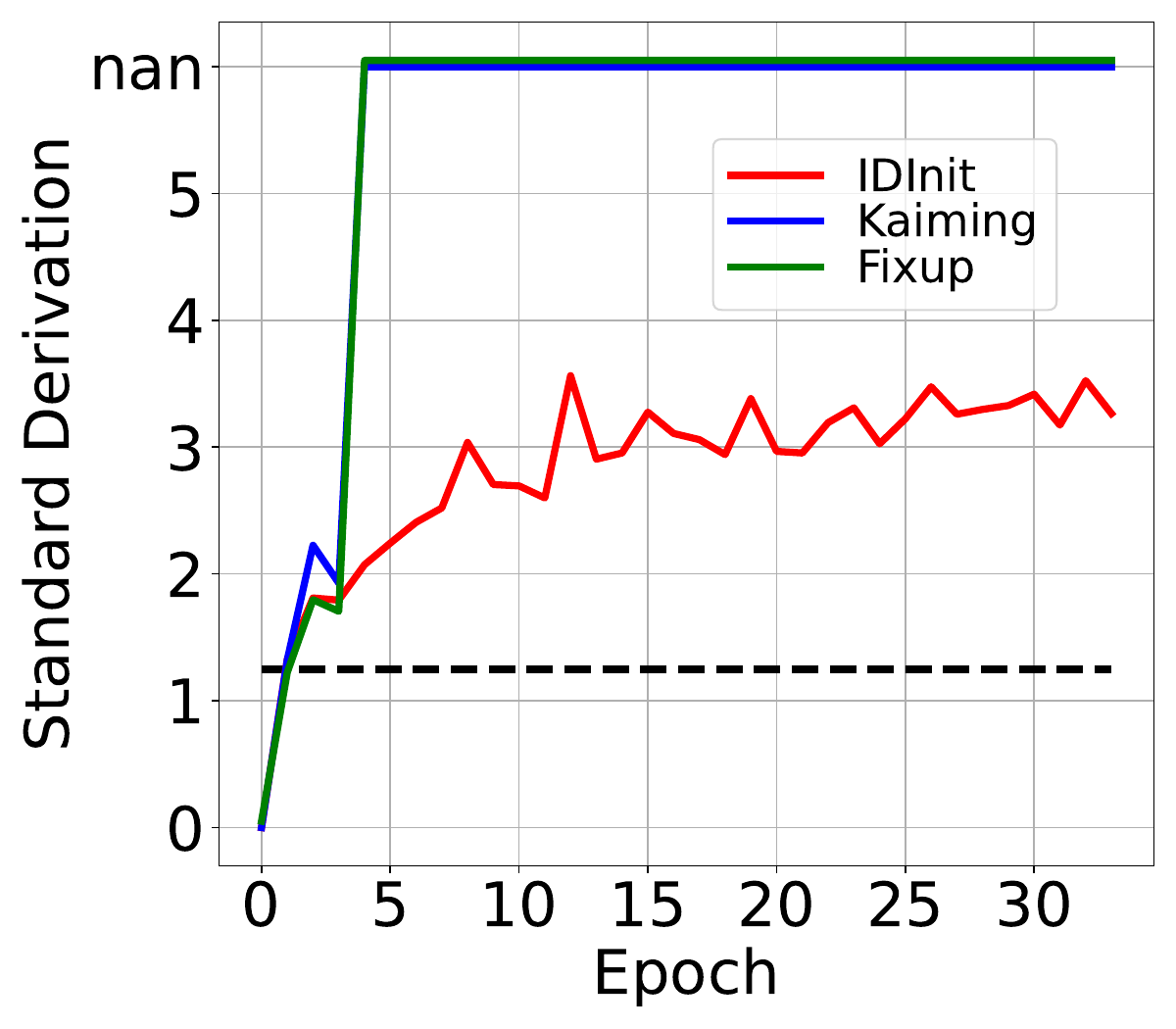}
	}
	\caption{Result of the experiment on the residual network with the long residual stem. Figure~\ref{fig:long-acc} shows the accuracy of different initialization. Figure~\ref{fig:long-std} shows the standard derivations of the outputs of networks with different initialization methods. The black dash line is the standard derivation of the network input.}
	\label{fig:failure-case}
\end{figure}

Results are shown in Figure~\ref{fig:failure-case}. When the network is trained for 4 epochs, both Kaiming and Fixup fail to train the network, since the standard derivations of their outputs explode. By contrast, IDInit successfully trains this network and the standard derivation of the output converges to a stable value. This experiment demonstrates the ability of IDInit to stabilize the residual stem, which can benefit the training of the whole network.

\begin{table}[h]

\renewcommand{\arraystretch}{1.1}
\caption{Architectures of Res-112. Window means the convolutional kernel window size. Channels indicate $\mathbf{c}_{in}$ and $\mathbf{c}_{out}$ of a standard convolutional kernel $\ca{C}\in \mathbb{R}^{\mathbf{c}_{in}\times \mathbf{c}_{out}\times k \times k}$. The avg pool denotes the average pooling operation. Linear means a linear layer.}
\label{tbl:longstemarch}
\begin{center}
\begin{tabular}{@{}ccc@{}}
\toprule
Layer                           & Window                      & Channels                                                        \\ \midrule
conv1                           & 3$\times$3                  & 3$\times$16                                                     \\ \midrule
\multirow{6}{*}{Residual Block} & 3$\times$3                  & {[}16$\times$16{]}$\times$18                                    \\ \cmidrule(l){2-3} 
                                & \multirow{2}{*}{3$\times$3} & 16$\times$32                                                   \\ \cmidrule(l){3-3} 
                                &                             & {[}32$\times$32{]}$\times$17                                    \\ \cmidrule(l){2-3} 
                                & \multirow{2}{*}{3$\times$3} & 32$\times$64                                                   \\ \cmidrule(l){3-3} 
                                &                             & {[}64$\times$64{]}$\times$17                                    \\ \cmidrule(l){2-3} 
                                & 3$\times$3                  & 64$\times$64                                                    \\ \midrule
conv2                           & 3$\times$3                  & \begin{tabular}[c]{@{}c@{}}64$\times$64\\ avg pool\end{tabular} \\ \midrule
Linear                          &                             & 64$\times$10                                                    \\ \bottomrule
\end{tabular}
\end{center}
\end{table}

\subsection{Experiment on GPT-Base-MOE}

\begin{figure}[h]
	\centering
    \includegraphics[width=0.6\textwidth]{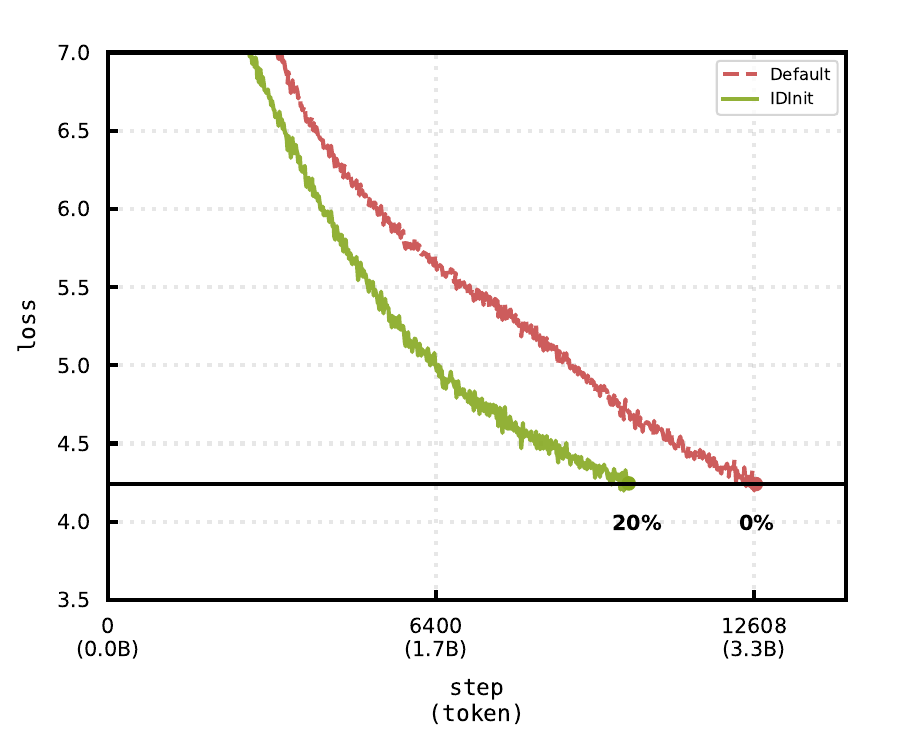}
	\caption{Pretraining on GPT-Base-MOE. IDIinit can achieve 20\% after than Default initialization.}
	\label{fig:gpt-moe}
\end{figure}

We conducted experiments on GPT-Base-MOE, modifying the GPT-Base with 8 experts. The training settings mainly follow \citet{DBLP:conf/nips/0005YYXSJL23}. The results, shown in Figure~\ref{fig:gpt-moe}, indicate that IDInit can achieve 20\% faster performance compared to the default random initialization, demonstrating the superior performance of IDInit.

\subsection{Experiment on DiT}

\begin{figure}[h]
	\centering
    \includegraphics[width=0.6\textwidth]{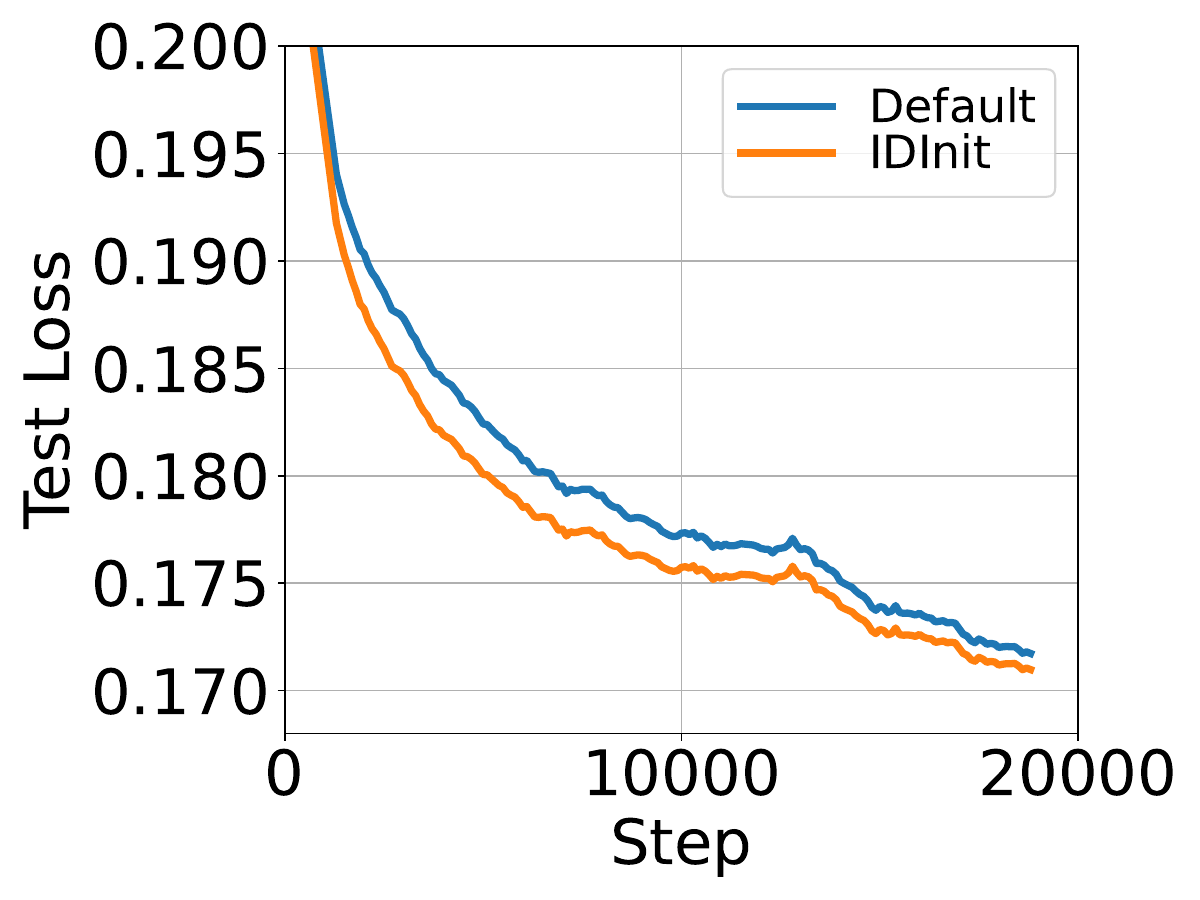}
	\caption{Training on DiT-S/4.}
	\label{fig:dit-s4}
\end{figure}

We train DiT-S/4 on ImageNet using the provided code\footnote{\url{https://github.com/facebookresearch/DiT}}. The experiment is conducted using the default training settings. As illustrated in Figure~\ref{fig:dit-s4}, IDInit consistently achieves faster convergence compared to the default initialization.

\section{Dynamical Isometry in IDInit}
\label{sec:toy}

\begin{figure}[h]
	\centering
        \subfigure[Non-Residual Plot.]{
		\includegraphics[width=0.4\textwidth]{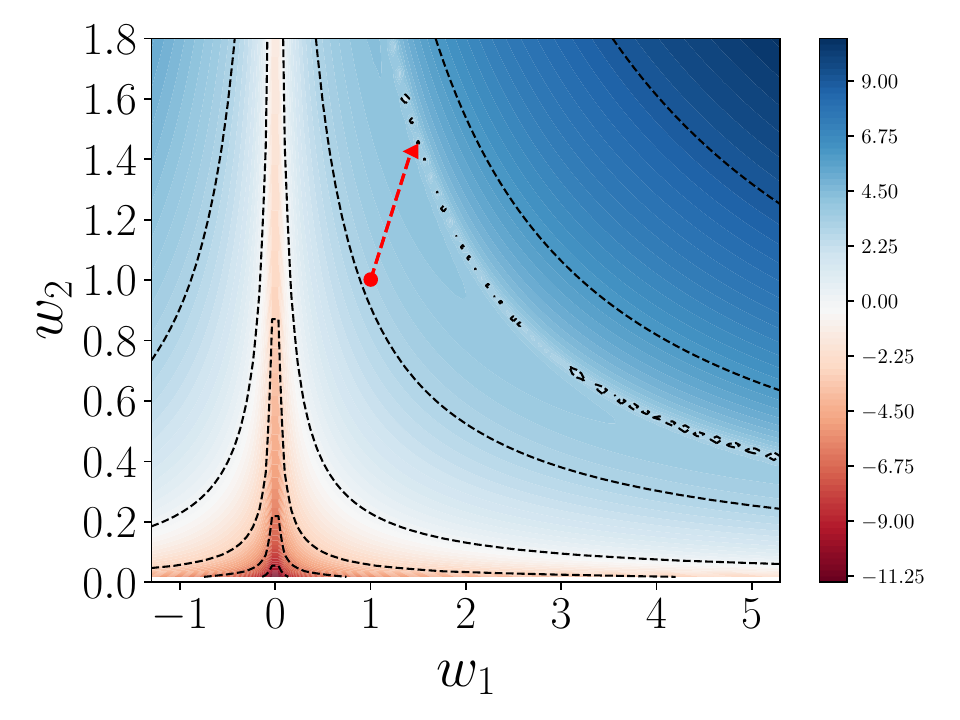}
		\label{fig:nonresidual-isometry}
	}
	\subfigure[Residual Plot.]{
		\includegraphics[width=0.4\textwidth]{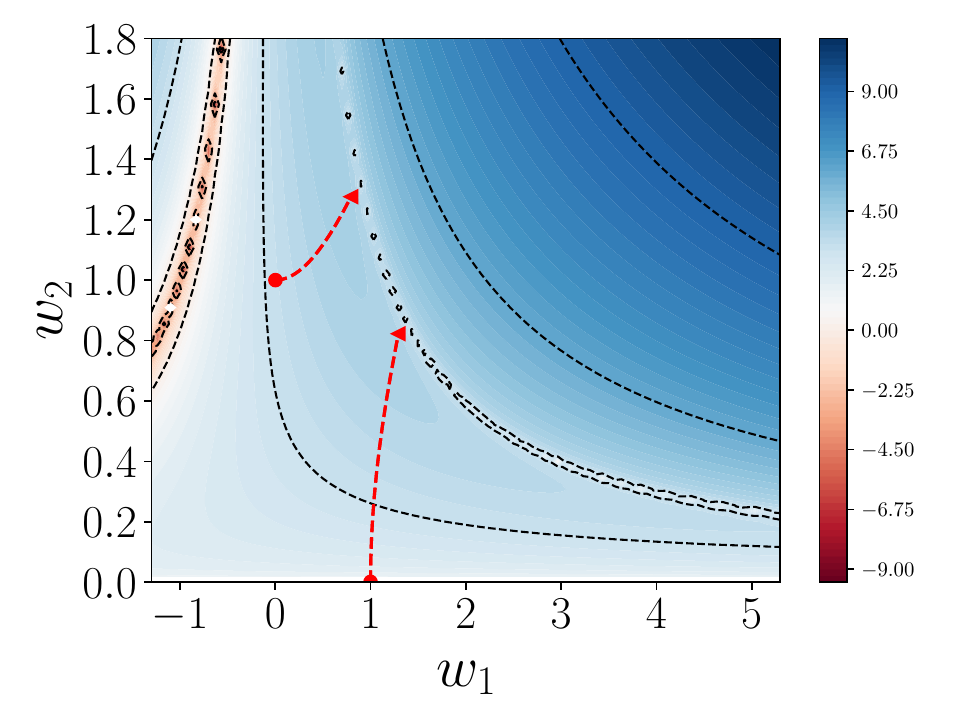}
		\label{fig:residual-isometry}
	}
	\caption{Contour plots of the log gradient norm $\log||\partial R||_2$ on non-residual and residual networks. $w^{(1)}$ and $w^{(2)}$ are both weights. The training process set as \citet{DBLP:conf/uai/BachlechnerMMCM21}, which is conducted on ground-truth $x^{(L)}=50\times x_0$ via gradient descent using a training set of $x_0 = \{1., 1.1, . . . , 1.8\}$. \subref{fig:nonresidual-isometry} shows $\{w^{(2)}=w^{(1)}=1\}$ can avoid poorly conditioned regions around 0, and converge to $w^{(1)}w^{(2)}=2.19$. \subref{fig:residual-isometry} cares about two initial position $\{w^{(1)}=0, w^{(2)}=1\}$ and $\{w^{(2)}=1, w^{(1)}=0\}$. The two points' trajectories do not also pass the poor regions around $w^{(1)}=-1, w^{(2)}=1$ and converge to the solution $w^{(1)}w^{(2)}=1.19$.}
    \label{fig:toy-isometry}
\end{figure}

Following \citet{DBLP:conf/uai/BachlechnerMMCM21}, we utilize a simple example of the mechanism that dynamical isometry helps IDInit to obtain a fast convergence. Considering a $L$-layer network with a simple special case of Eq. (\ref{eq:residual}):
\begin{align}
\label{eq:toylayer}
    x^{(L)} = (r + w^{(2)}w^{(1)})^Lx^{(0)},
\end{align}
where $w^{(1)}$ and $w^{(2)}$ denote the first weight and last weight in a residual stem respectively, and $x^{(\ast)}$ is the feature in layers. $r \in \{0, 1\}$ determines residual connection. Specifically, $r=0$ and $r=1$ represent non-residual and residual conditions respectively. The Jacobian of Eq.~(\ref{eq:toylayer}) is $J_{0L} = (r + w^{(2)}w^{(1)})^L$. Obviously, identity transition on both non-residual and residual settings, namely $\{r=0, w^{(2)}=w^{(1)}=1\}$ and $\{r=1, w^{(1)}=1, w^{(2)}=0\}$ respectively, will achieve $J_{0L}=1$, which conforms to the dynamical isometry mechanism that helps improving training ability~\citep{DBLP:conf/nips/PenningtonSG17}. Further, we delve into a gradient update analysis. Following gradient descent, $w_1$ can be updated with
\begin{align}
\label{eq:gdupdate}
    \Delta w^{(1)} = - \lambda Lw^{(2)}x^{(0)}(r+w^{(2)}w^{(1)})^{L-1}\partial_xR(x)|_{x=x^{(L)}},
\end{align}
where $R$ means the loss function, and $\lambda$ is a learning rate. As $w^{(1)}$ and $w^{(2)}$ are equivalent in Eq.~(\ref{eq:toylayer}), $w^{(2)}$ can be updated similar to Eq.~(\ref{eq:gdupdate}). When $w^{(1)}=1$, updates are required less than 1. Therefore, the learning rate is constrained to
\begin{align}
    \begin{cases}
  \lambda \propto L^{-1}, & \text{if non-residual}, \\
    \lambda \propto L^{-1}(1+w^{(2)})^{L-1}, & \text{if residual}.
\end{cases}
\end{align}
For the non-residual condition, the learning rate is polynomial to $L$, thereby insensitive to the depth. By contrast, in the residual block, $w^{(2)} >> 0$ will cause learning rate exponentially small and $w^{(2)} = -1$ also cause gradient diffusion. On this condition, setting $w^{(2)} = 0$ can be a good solution for avoiding large output and restricting gradients in a suitable norm. Besides, it is feasible to update $w^{(2)}$ with the first non-trial step
\begin{align}
    w^{(2)} = - \lambda Lw^{(1)}x^{(0)}\partial_xR(x)|_{x=x^{(L)}},
\end{align}
and will converge with a learning rate that is polynomial in the depth $L$ of the network. We plot the training dynamics in Figure~\ref{fig:toy-isometry}, and use this simple example to illustrate the mechanism of IDInit, which is always a well-conditioned position for training.

\end{document}